\pdfoutput=1
\documentclass[11pt]{article}
\usepackage{float}
\usepackage{colortbl}
\usepackage{xcolor}
\usepackage{pgf}
\usepackage{pgffor}
\usepackage{paralist}
\usepackage[most]{tcolorbox}

\definecolor{lightgreen}{RGB}{144,238,144}
\definecolor{lightgray2}{gray}{0.92}
\definecolor{lightgray1}{gray}{0.8}
\definecolor{resultgreen}{rgb}{0.0, 0.4, 0.0}   % Dark green
\definecolor{resultgreen}{rgb}{0.0, 0.4, 0.0}    % Slightly lighter for underline

\usepackage[final]{emnlp2023}
\usepackage{float}
\usepackage{times}
\usepackage{latexsym}
\usepackage{graphicx}
\usepackage{booktabs}
\usepackage{makecell}
\usepackage[utf8]{inputenc}
\usepackage{textgreek}
\usepackage[T1]{fontenc}
\DeclareUnicodeCharacter{202F}{\,}
\usepackage[utf8]{inputenc}

\usepackage{microtype}

\usepackage{inconsolata}

\usepackage{array}
\usepackage{multirow}

\newcommand{\x}{\phantom{$-$}}

\title{Cross-Cultural Transfer of Commonsense Reasoning in LLMs: \\Evidence from the Arab World}
\author{Saeed Almheiri\thanks{\hspace{0.2cm}Equal contribution} \quad Rania Hossam\footnotemark[1] \quad Mena Attia \quad Chenxi Wang   \\ 
\textbf{Preslav Nakov} \quad \textbf{Timothy Baldwin} \quad \textbf{Fajri Koto} \\
Mohamed bin Zayed University of Artificial Intelligence\\
\texttt{\small \{saeed.y, rania.elbadry, mena.attia, chenxi.wang\}@mbzuai.ac.ae}}

\begin{document}

\maketitle
\begin{abstract}

Large language models (LLMs) often reflect Western-centric biases, limiting their effectiveness in diverse cultural contexts. Although some work has explored cultural alignment, the potential for cross-cultural transfer, using alignment in one culture to improve performance in others, remains underexplored. This paper investigates cross-cultural transfer of commonsense reasoning in the Arab world, where linguistic and historical similarities coexist with local cultural differences. Using a culturally grounded commonsense reasoning dataset covering 13 Arab countries, we evaluate lightweight alignment methods such as in-context learning and demonstration-based reinforcement (DITTO), alongside baselines like supervised fine-tuning and direct preference optimization. Our results show that merely 12 culture-specific examples from one country can improve performance in others by 10\% on average, within multilingual models. In addition, we demonstrate that out-of-culture demonstrations from Indonesia and US contexts can match or surpass in-culture alignment for MCQ reasoning, highlighting cultural commonsense transferability beyond the Arab world. These findings demonstrate that efficient cross-cultural alignment is possible and offer a promising approach to adapt LLMs to low-resource cultural settings.

%Recent advances in large language models (LLMs) have highlighted the need for culturally aligned responses, particularly since current models often exhibit western-centric biases. In this paper, we investigate the cross-cultural effect, the use of knowledge from one defined culture to learn about another, through lightweight alignment approaches, namely in-context learning (ICL) and the novel approach of demonstration-based reinforcement learning (DITTO), to enhance LLM capability on Arab culture commonsense reasoning. Using a dataset covering 13 Arab countries, we demonstrate that a small number of culture-specific examples (precisely 12) from one country can improve accuracy in other culturally distinct countries by 15 to 20\% on average. 
% We further observe that while both alignment methods yield robust gains, ICL can outperform DITTO by up to 3\% on more fine-grained cultural topic alignment, such as food demonstrations. 
%Our analysis reveals varying correlations between geographic distance and cross-cultural transfer effect ranging between -0.8 and 0.65. Furthermore, targeted cultural alignment (for example, UAE-specific alignment) improves the linear separability of that country within the latent space of the model without affecting the overall performance gains on the other Arab cultures. These findings underscore the feasibility of cultural transferability through data-efficient alignment, offering a path toward mitigating Western-centric bias and improving fairness in low-source cultural domains.

\end{abstract}

\begin{figure}[h]
  \centering
  \includegraphics[width=\columnwidth]{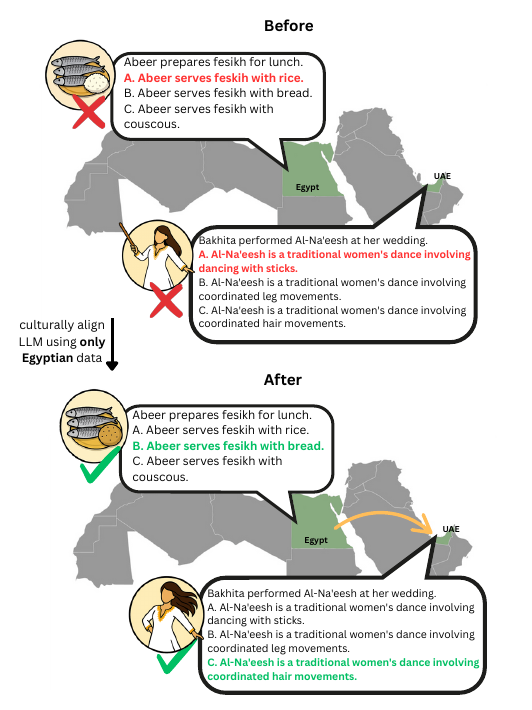}
  \caption{An illustration of cross-cultural transfer: if aligning an LLM with Egyptian data improves its performance on UAE culture, cultural knowledge has been transferred.}
  \label{fig:cultural-transfer}
\end{figure}

\begin{figure*}[t]
    \centering
    \includegraphics[width=1\linewidth]{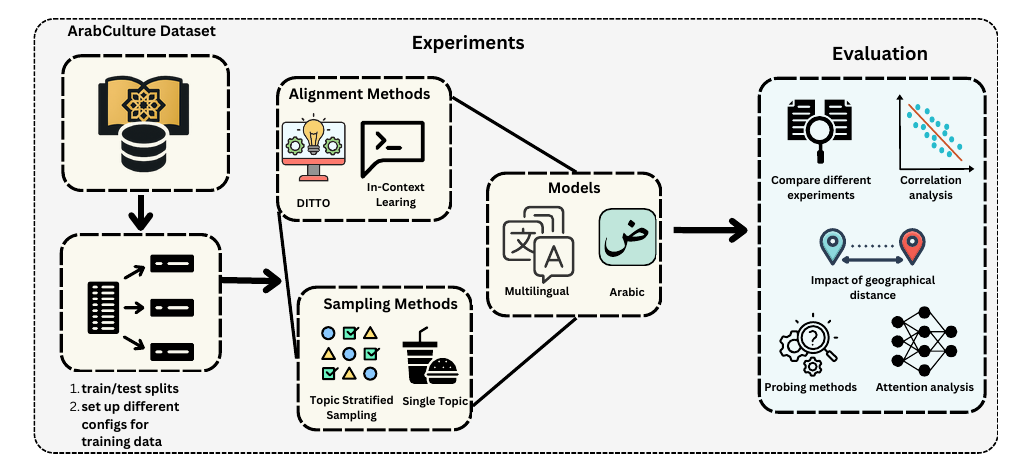} 
    \caption{This figure illustrates an overview of our alignment and evaluation pipeline. The ArabCulture dataset is split into train/test subsets, aligned via either In-Context Learning or DITTO on different models with different sampling methods, then evaluated and probed (stimulus, attention, correlation) to quantify cross-cultural transfer.}
    \label{fig:overview}
\end{figure*}

\section{Introduction}
Large language models (LLMs) are increasingly being deployed across diverse cultural contexts, yet they often reflect a Western-centric worldview, misaligning with local customs, values, and norms of non-Western cultures~\cite{naous2024havingbeerprayermeasuring, sadallah2025commonsensereasoningarabculture, wang-etal-2024-countries}. Prior studies have explored broad East--West cultural misalignments in LLMs \cite{naous2024havingbeerprayermeasuring}, but little is known about how these models handle intra-regional cultural variation, such as that found across the 22 Arab countries. For example, despite sharing linguistic ties, Emirati culture differs significantly from Egyptian or Syrian traditions in food, festivals, and gender roles. However, most Arabic LLMs are trained on translated English data or regionally-aggregated corpora~\cite{jais, sadallah2025commonsensereasoningarabculture}, potentially flattening these cultural distinctions.

A key challenge in aligning LLMs with country-specific cultural knowledge is the uneven availability of data. High-population countries such as Egypt provide vastly more online content than smaller ones like the UAE (114M vs.\ 1.3M population)~\cite{gmi2025uaepopulation, unfpa2025egypt}, resulting in cultural underrepresentation. This disparity motivates a central question: \textit{Can cultural knowledge from one country be transferred to benefit another with limited resources?}

In this paper, we investigate the feasibility of cross-cultural commonsense transfer within the Arab world (see Figure~\ref{fig:cultural-transfer}). We focus on this region because it combines a shared linguistic foundation with rich cultural diversity, and because most existing Arabic LLMs are trained on aggregated or translated data that risk obscuring local distinctions. Specifically, we evaluate whether aligning an LLM to the culture of one Arab country can enhance its performance on others through two lightweight alignment strategies: In-Context Learning (ICL) and Demonstration-based Iterative Task Tuning Optimization (DITTO)~\cite{shaikh2024showdonttellaligning} (see Figure~\ref{fig:overview}). While ICL is a strong few-shot baseline, DITTO offers a reinforcement learning alternative that requires only a handful of high-quality demonstrations, making it particularly suitable for low-resource cultural domains.

%In this paper, we investigate the feasibility of cross-cultural commonsense transfer within the Arab world. 
%Specifically, we ask: \textit{Can aligning an LLM to the culture of one Arab country improve performance on others?} 
%Specifically, we evaluate whether aligning an LLM to the culture of one Arab country can enhance its performance on others through two lightweight alignment strategies: In-Context Learning (ICL) and Demonstration-based Iterative Task Tuning Optimization (DITTO)~\cite{shaikh2024showdonttellaligning}. Although ICL is a strong few-shot baseline, DITTO offers a reinforcement learning alternative that requires only a handful of high-quality demonstrations, making it particularly suitable for low-resource cultural domains.

%We construct experiments over a 13-country, 3.2k-example ArabCulture dataset spanning diverse domains such as food, rituals, relationships, and social norms. Using only 12 cultural demonstrations per source country, we test transfer to unseen target cultures across four LLMs (Qwen2.5, Gemma-2, ALLaM, and SILMA)~\cite{qwen2.5,team2024gemma,bariallam,silma_01_2024}. We further probe whether cross-cultural improvement is predictable from geographic proximity, cosine similarity between data samples across countries, and whether alignment reshapes latent cultural representations in the model’s internal space.

We conduct experiments on the \texttt{ArabCulture} dataset \cite{sadallah2025commonsensereasoningarabculture}, covering 13 countries and 3.2k examples across domains such as food, rituals, relationships, and social norms. Using only 12 culture-specific demonstrations per source country, we evaluate transfer to unseen target cultures across four LLMs (Qwen2.5, Gemma-2, ALLaM, and SILMA)~\cite{qwen2.5,team2024gemma,bariallam,silma_01_2024}. Beyond performance, we examine whether cross-cultural gains can be predicted from geographic proximity or cross-country data similarity, and whether alignment reshapes latent cultural representations within the models.

\noindent Our contributions are:
\begin{compactitem}
    \item We pioneer the use of DITTO for cultural alignment, achieving up to 34\% accuracy gains in Arab commonsense reasoning MCQ with only 12 demonstrations per country.
    \item We show that cross-cultural transfer is feasible: cultural knowledge from high-resource countries improves LLM performance on culturally distinct, low-resource ones.
    \item We perform probing and correlation analyses to explain improvements with factors such as geographic proximity and cultural similarity modeling, and that targeted alignment enhances the linear separability of specific cultures in the model's latent space.
\end{compactitem}

%Our findings offer a compelling path toward culturally adaptive NLP systems using minimal, targeted supervision, a crucial step for inclusive and globally applicable AI.

\section{Related Work}
\subsection{Cultural Reasoning}
While language models encode cultural knowledge, they often overrepresent high-resource languages and cultures \cite{shen-etal-2024-understanding, wang-etal-2024-countries, naous2024havingbeerprayermeasuring}. To evaluate such biases, several benchmarks have been introduced, including cultural reasoning tasks for Indonesian \cite{koto2024} and Arabic \cite{sadallah2025commonsensereasoningarabculture, huang-etal-2024-acegpt, mousi-etal-2025-aradice}. These studies show that LLMs struggle with cultural reasoning compared to general commonsense reasoning in English \cite{roemmele2011choice}. For Arabic, available resources include \texttt{ACVA} \cite{huang-etal-2024-acegpt}, which provides general true/false statements about Arab culture as a whole, and \texttt{AraDiCE} \cite{mousi-etal-2025-aradice}, which covers cultural questions from six Arab countries. In this work, we rely on \texttt{ArabCulture} \cite{sadallah2025commonsensereasoningarabculture}, as it offers the most comprehensive dataset in terms of scale and country-level coverage.  

%Disparities in cultural knowledge persist, often favoring dominant cultures \cite{shen-etal-2024-understanding, wang-etal-2024-countries, naous2024havingbeerprayermeasuring}. Recent work shows that including geographical context in prompts boosts model performance on low-resource cultural reasoning tasks \cite{koto2024}. Likewise, culturally aware data collection, targeted model adaptation, and robust evaluation frameworks are crucial in addressing linguistic diversity and cultural biases \cite{hershcovich-etal-2022-challenges}. Several Arabic cultural datasets and benchmarks have been developed, including the ACVA Arabic Culture benchmark, which includes general true/false statements about Arab culture as a whole \cite{huang-etal-2024-acegpt}, and the AraDiCE-Culture benchmark, which includes cultural questions from only six Arab countries \cite{mousi-etal-2025-aradice}.

%The assumption that underlies most of the work related to Arabic cultural alignment of LLMs is that Arabs share the same culture, either entirely or regionally, raising the question of whether Arab culture is a homogeneous culture or a diverse set of cultures \cite{keleg-2025-llm}. We further investigate this assumption by experimenting with country-level cultural alignment using the Arabic Culture dataset \cite{sadallah2025commonsensereasoningarabculture}, which consists of cultural data from 13 Arab countries, and conducting a fine-grained comprehensive analysis.

Much of the prior work on Arabic cultural alignment has focused on evaluation benchmarks or on large-scale pretraining approaches such as instruction fine-tuning \cite{jais,bariallam}. However, none of these studies examine whether adapting an LLM to one culture can improve—or potentially degrade—its performance on another. This question is particularly important in the Arab world, where countries share linguistic and historical ties but maintain distinct cultural practices. To explore this, we leverage the \texttt{ArabCulture} dataset \cite{sadallah2025commonsensereasoningarabculture}, which provides fine-grained, country-level cultural knowledge across 13 Arab countries, enabling us to systematically study cross-cultural transfer within the region.

\subsection{Cultural Alignment Approaches}

Recent work has explored improving the cultural awareness of language models through alignment techniques such as fine-tuning \cite{li2024culturellmincorporatingculturaldifferences} and in-context learning via few-shot prompting \cite{wang-etal-2024-countries, alkhamissi-etal-2024-investigating}. Fine-tuning can effectively adapt models to cultural data, but it often requires substantial resources and risks catastrophic forgetting of prior knowledge \cite{choenni-etal-2024-echoes, alkhamissi-etal-2024-investigating}. Reinforcement learning provides an alternative by leveraging feedback from a reward model to guide LLMs with only a small set of demonstrations. Recent preference-alignment methods such as Direct Preference Optimization (DPO: \citet{rafailov2024direct}) and its extension DITTO \cite{shaikh2024showdonttellaligning} demonstrate that iterative feedback can align models efficiently without large-scale data. Although DITTO was originally developed for stylistic adaptation, we adopt it here for the novel task of cultural alignment. Building on these approaches, we study how LLMs can be aligned with regional cultural nuances while maintaining broader commonsense reasoning, with a particular focus on the Arab world. Specifically, we examine whether adapting to the cultural knowledge of one country improves or harms performance in others, and how factors such as geographical distance or cultural similarity influence cross-cultural generalization.

% \begin{figure}[!ht]
%   \centering
%   \includegraphics[width=0.5\textwidth]{analysis_graphs/data.png}
%   \caption{Countries Ranked by Effectiveness as Cultural Teachers. The chart displays which countries' cultural data provides the most effective training examples for cross-cultural transfer.}
%   \label{fig:countries}
% \end{figure}

\begin{figure}[!t]
  \centering
  \includegraphics[width=\linewidth]{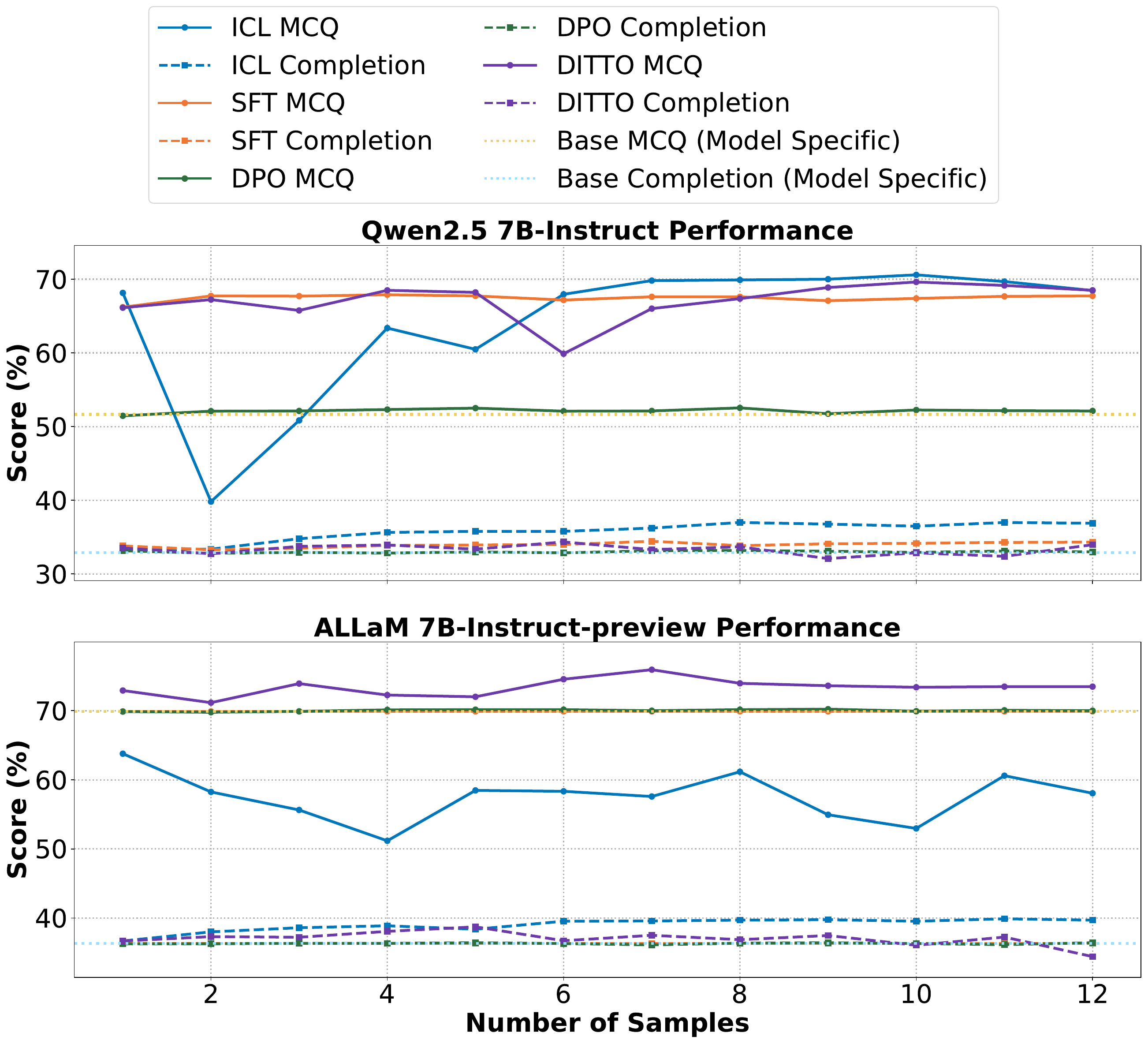}
  \caption{Sample efficiency of different alignment methods for cultural alignment, evaluated on multiple-choice questions (MCQ) and completions using cultural demonstrations from the UAE.}
  \label{fig:sample_efficancy}
\end{figure}

\section{Methodology}
%An overview of our methodology is outlined in Figure \ref{fig:overview}. The dataset is split into train/test sets, aligned using ICL or DITTO across models and sampling strategies, then evaluated and analyzed to measure cross-cultural transfer.
We describe the data, alignment procedures (ICL and DITTO), sampling strategies, and the evaluation protocol for measuring cross-cultural transfer (Figure \ref{fig:overview}). We emphasize ICL and DITTO because pilot experiments (Figure~\ref{fig:sample_efficancy}) showed they surpassed instruction fine-tuning and vanilla DPO on our tasks, warranting deeper exploration.

\subsection{Arabic Culture Dataset}
We use the \texttt{ArabCulture} dataset \cite{sadallah2025commonsensereasoningarabculture}, which consists of approximately 3,200 handcrafted cultural statements and the corresponding multiple choice options (one correct, two incorrect). The dataset spans 12 topics and 40+ subtopics from 13 countries grouped into 4 regions of the Arab world (North Africa, Gulf region, Nile Valley, and Levant). Each country subset consists of roughly 250 pairs of statements and choices. For each country, we split these 10\% for training/alignment examples and 90\% held-out for evaluation. %This ensures that our evaluation always assesses the model on unseen cultural statements.

\definecolor{resultgreen}{rgb}{0.0, 0.4, 0.0}  % Darker green for top values

% Table comparing all models with updated color formatting
\begin{table*}[t]
\centering
\Large
\scriptsize
\setlength{\tabcolsep}{2.2pt}
\label{tab:full_model_scores}
\scalebox{1.04}{

\begin{tabular}{l|rrrr|rrrr|rrrr|rrrr|r}
\toprule
\multirow{3}{*}{\textbf{Country}} 
& \multicolumn{4}{c|}{\rotatebox{0}{\includegraphics[height=2.5ex]{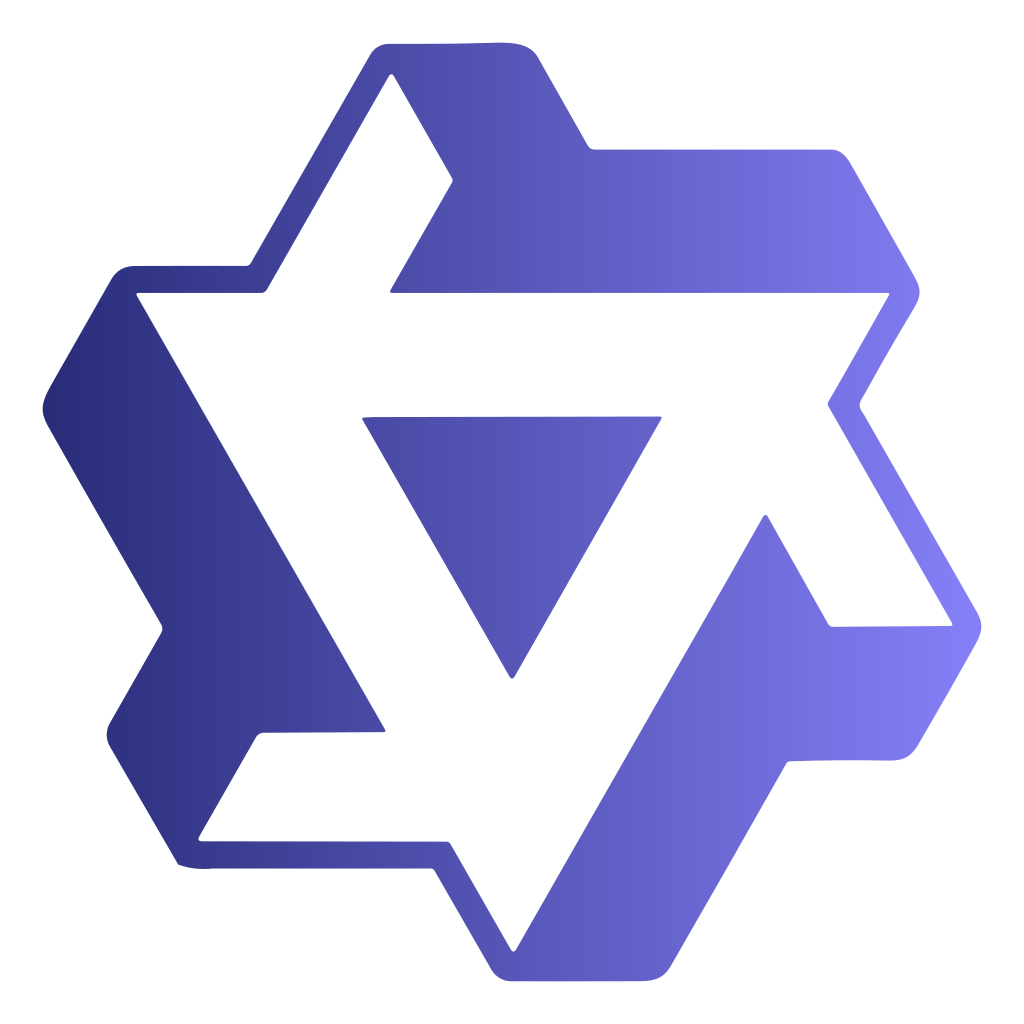}} Qwen2.5 7B-Inst} 
& \multicolumn{4}{c|}{\rotatebox{0}{\includegraphics[height=2.5ex]{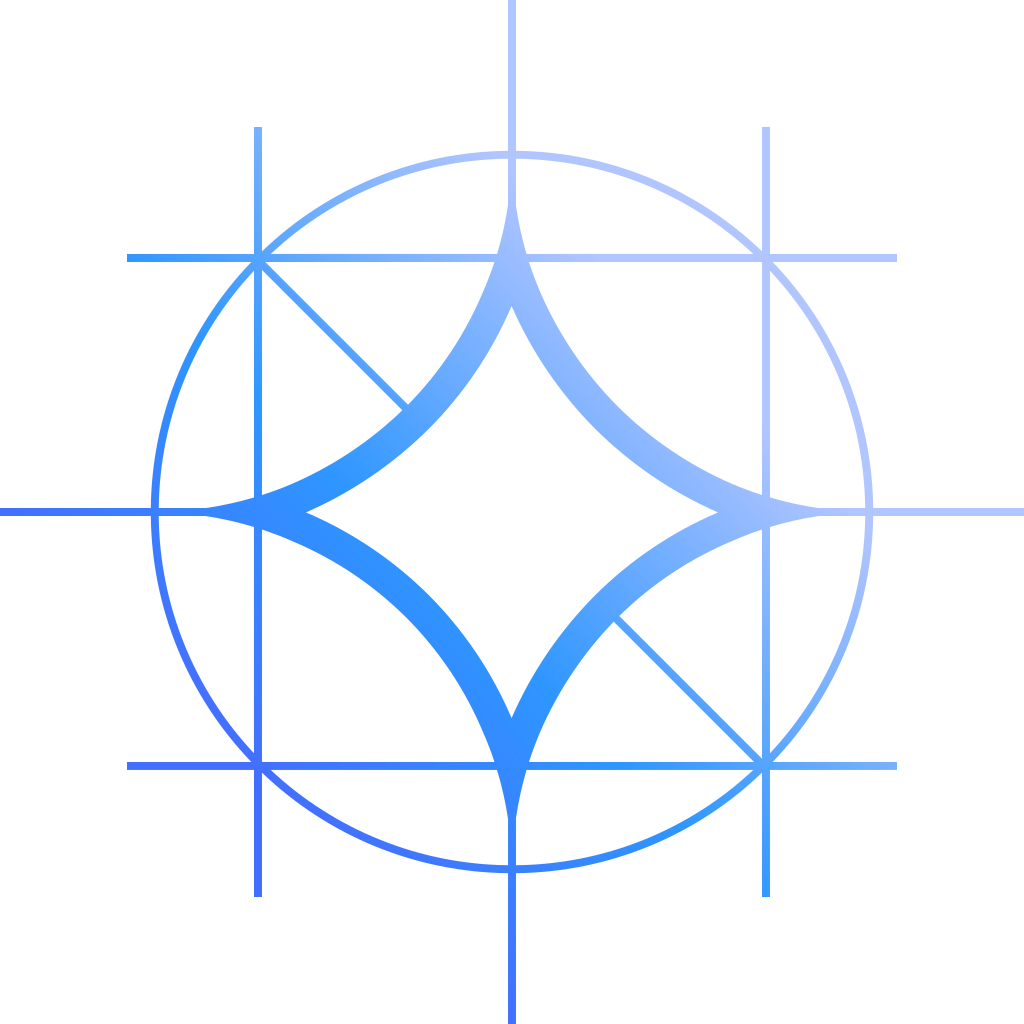}} Gemma-2 9B-it} 
& \multicolumn{4}{c|}{\rotatebox{0}{\includegraphics[height=2.5ex]{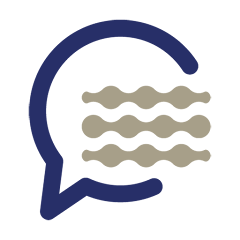}} ALLaM 7B-Inst} 
& \multicolumn{4}{c|}{\rotatebox{0}{\includegraphics[height=2.5ex]{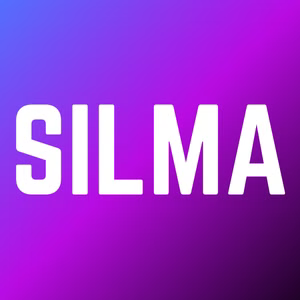}} SILMA 9B-Inst} 
& \multirow{3}{*}{\textbf{Avg.}} \\
\cmidrule{2-17}
& \multicolumn{2}{c}{\textbf{DITTO}} & \multicolumn{2}{c|}{\textbf{ICL}} 
& \multicolumn{2}{c}{\textbf{DITTO}} & \multicolumn{2}{c|}{\textbf{ICL}} 
& \multicolumn{2}{c}{\textbf{DITTO}} & \multicolumn{2}{c|}{\textbf{ICL}} 
& \multicolumn{2}{c}{\textbf{DITTO}} & \multicolumn{2}{c|}{\textbf{ICL}} 
& \\
\cmidrule{2-17}
& \textbf{Comp.} & \textbf{MCQ} & \textbf{Comp.} & \textbf{MCQ}
& \textbf{Comp.} & \textbf{MCQ} & \textbf{Comp.} & \textbf{MCQ}
& \textbf{Comp.} & \textbf{MCQ} & \textbf{Comp.} & \textbf{MCQ}
& \textbf{Comp.} & \textbf{MCQ} & \textbf{Comp.} & \textbf{MCQ}
& \\
\midrule
Algeria   & 0.19  & 16.74  & 2.80  & 18.50  & $-$0.22 & 25.26  & \textcolor{resultgreen}{3.86}  & 4.52  & $-$0.31  & 1.57  & 3.99  & $-$10.49 & $-$0.88 & 2.14 & 1.82 & 0.79  & 4.39 \\
Egypt     & $-$0.09 & 17.56  & 2.39  & \textcolor{resultgreen}{19.67}  & $-$1.61 & 28.34  & 1.72  & 0.66  & $-$2.77  & $-$1.10 & 3.61  & $-$25.32 & $-$0.97 & $-$1.22 & 0.03 & $-$0.66 & 2.52 \\
Jordan    & $-$0.31 & 18.91  & 2.77  & 17.09  & 1.47  & \textcolor{resultgreen}{33.93}  & \textcolor{resultgreen}{4.80}  & 11.21 & $-$4.84  & $-$12.82 & 3.86  & $-$4.84  & 1.13  & 1.57 & \textcolor{resultgreen}{2.07} & 3.11  & \textcolor{resultgreen}{4.94} \\
KSA       & $-$0.06 & 17.84  & 3.24  & \textcolor{resultgreen}{19.92}  & 3.30  & 27.46  & 3.42  & 6.00  & 0.78   & $-$1.91 & 3.14  & $-$16.27 & 0.22  & 0.66 & 1.26 & 1.39  & 4.40 \\
Lebanon   & 0.91  & 18.38  & \textcolor{resultgreen}{3.52}  & 18.66  & 1.28  & 7.19   & 3.52  & $-$0.03 & 0.34   & \textcolor{resultgreen}{3.71}  & 3.99  & $-$14.92 & 0.88  & $-$3.01 & 0.63 & $-$1.54 & 2.72 \\
Libya     & $-$0.12 & 15.11  & 3.27  & 16.71  & 0.37  & \textcolor{resultgreen}{33.55}  & 2.38  & $-$0.06 & $-$2.07  & $-$0.28 & 2.64  & $-$13.22 & 0.82  & $-$0.34 & 1.35 & $-$6.97 & 3.32 \\
Morocco   & 1.64  & 17.25  & 3.33  & 18.79  & $-$0.41 & 13.70  & 3.77  & 6.91  & $-$0.09  & 3.14  & 3.93  & $-$11.72 & 1.13  & 2.71 & 1.89 & \textcolor{resultgreen}{3.49}  & 4.34 \\
Palestine & $-$0.03 & 17.97  & 1.45  & 18.03  & 0.31  & 24.94  & 3.42  & 0.47  & $-$3.33  & 0.82  & 2.80  & $-$19.76 & 0.38  & 2.05 & 1.54 & 0.00  & 3.19 \\
Sudan     & 1.07  & 18.98  & 3.21  & 16.15  & 1.44  & 15.11  & 3.26  & 14.32 & $-$1.67  & 1.73  & 3.20  & $-$22.87 & 1.70  & 2.74 & 1.85 & 1.10  & 3.83 \\
Syria     & 0.98  & 17.00  & 3.30  & 19.04  & 0.15  & 31.82  & 2.60  & 0.44  & $-$1.45  & 2.55  & \textcolor{resultgreen}{4.21}  & $-$5.90  & $-$0.28 & 1.95 & 0.57 & \textcolor{resultgreen}{3.40}  & \textcolor{resultgreen}{5.02} \\
Tunisia   & $-$0.81 & 17.18  & 2.01  & 18.13  & 1.35  & 20.58  & 2.29  & 0.60  & 0.22   & 1.35  & \textcolor{resultgreen}{4.33}  & $-$18.60 & 0.28  & 2.33 & 0.38 & 0.35  & 3.25 \\
UAE       & 1.07  & 16.84  & \textcolor{resultgreen}{3.99}  & 16.81  & 2.38  & 28.15  & 3.55  & 1.57  & $-$2.07  & \textcolor{resultgreen}{3.58}  & 3.36  & $-$11.84 & 1.70  & 2.20 & \textcolor{resultgreen}{2.04} & 2.27  & 4.73 \\
Yemen     & $-$0.91 & 18.57  & 2.14  & 12.00  & $-$0.35 & 5.72   & 2.67  & 0.22  & 0.53   & $-$0.50 & 2.86  & $-$17.65 & $-$0.12 & 0.44 & 0.63 & $-$0.75 & 1.59 \\
\midrule
\rowcolor{gray!10}
\textbf{Avg.} 
          & 0.27 & 17.56 & 2.88 & \textcolor{resultgreen}{17.65} 
          & 0.73 & \textcolor{resultgreen}{22.75} & \textcolor{resultgreen}{3.17} & 3.60 
          & $-$1.29 & 0.14  & \textcolor{resultgreen}{3.53} & $-$14.88 
          & 0.46 & 1.09  & 1.24 & 0.46 & \\

\bottomrule
\multicolumn{18}{l}{\vspace{0.05cm} Accuracy Baselines (Comp.\%/MCQ\%): Qwen2.5 (32.89/51.65), Gemma-2 (32.52/34.56), ALLaM (36.35/69.9), SILMA (32.39/70.81)}
\end{tabular}
}
\caption{Overall accuracy improvements for Arab cultural commonsense reasoning when training on country-specific knowledge across different models with topic-based sampling. Results show performance on Completion and MCQ tasks using DITTO and ICL methods. Green-colored values represent the top two improvements in each model across MCQ and Completion.}
\label{table:main}
\end{table*}

\subsection{Alignment Methods}
%We adopt two main alignment approaches for LLMs to align on country specific cultural examples: in-context learning (ICL) and DITTO, a recently proposed lightweight method that extends Direct Preference Optimization (DPO) \citep{rafailov2024direct} and iteratively aligns model outputs to a small set of user-provided demonstrations \citep{shaikh2024showdonttellaligning}.  DITTO treats high-quality user-provided demonstrations as strictly preferred over intermediate model outputs, guiding the model toward better alignment through iterative preference-based updates. DITTO offers a data-efficient alternative to large-scale supervised fine-tuning or full-scale reinforcement learning from human feedback (RLHF) \citep{bai2022traininghelpfulharmlessassistant}, enabling precise cultural alignment from a small number of carefully selected examples as highlighted in Figure~\ref{fig:sample_efficancy}, resulting in high improvement in overall performance in Arab cultures. In our preliminary experiments, we compared DITTO and ICL with other relevant alignment techniques (SFT, DPO), as illustrated in Figure~\ref{fig:sample_efficancy}. Our results revealed distinct performance patterns, with DITTO excelling in MCQ tasks and ICL demonstrating stronger gains in completion tasks across Qwen-2.5-7B and ALLaM-7B, all while maintaining strong sample efficiency. Based on these findings, we selected DITTO and ICL for alignment on cultural commonsense reasoning.

We use two alignment methods for country-specific cultural examples: in-context learning (ICL) and DITTO, a lightweight DPO variant that iteratively prefers curated demonstrations \citep{rafailov2024direct,shaikh2024showdonttellaligning}. DITTO is data-efficient relative to SFT and full RLHF \citep{bai2022traininghelpfulharmlessassistant}, enabling alignment from few examples (Figure~\ref{fig:sample_efficancy}). In preliminary comparisons with SFT and vanilla DPO, DITTO led on MCQ, while ICL was stronger on completion tasks across Qwen-2.5-7B and ALLaM-7B (Figure~\ref{fig:sample_efficancy}). We therefore study ICL and DITTO in depth for cultural commonsense reasoning.

We evaluate four instruction-tuned baselines: two multilingual models (Qwen-2.5-7B-Instruct \cite{qwen2.5}, Gemma-2-9B-It \cite{team2024gemma}), and two Arabic-centric models (ALLaM-7B-Instruct-preview \cite{bariallam} and SILMA-9B-Instruct \cite{silma_01_2024}), enabling direct comparison between general multilingual pretraining and Arabic-oriented models.

%For our experiments, we use 2 multilingual models (Qwen-2.5-7B-Instruct \cite{qwen2.5} and Gemma-2-9B-It \cite{team2024gemma}), and 2 Arabic-centric models (ALLaM-7B-Instruct-preview \cite{bariallam} and SILMA-9B-Instruct \cite{silma_01_2024}) as baseline LLMs. This selection enables us to capture a broad view of cross-lingual and cross-cultural alignment effects, directly comparing models with general multilingual pretraining against those explicitly optimized for Arabic language and cultural understanding.

\subsection{Demonstration Sampling}

We adopt two in-context selection schemes for both ICL and DITTO: topic-stratified ($k=12$ per country, one per main topic) and food-focused ($k=12$ from the food topic spanning its subtopics). In both settings, the model conditions on the demonstrations and selects the culturally appropriate completion for a held-out statement–choice pair. Demonstrations are curated for coverage and cultural relevance.
%We employ two complementary in-context sampling strategies for both ICL and DITTO: topic-based sampling and food-based sampling. For topic-based sampling, we select 12 demonstration examples from the training subset of a specific country, ensuring one example per main topic to capture a broad thematic spectrum. In contrast, food-based sampling draws all 12 demonstrations exclusively from the "food" topic, covering a range of subtopics within this domain. In both setups, the model is prompted with these demonstrations and tasked with selecting the most culturally appropriate completion for an unseen statement–choice pair. Demonstration examples are curated to represent diverse topics, promoting comprehensive coverage of region-specific cultural knowledge and reasoning patterns.

\subsection{Evaluation}

% We evaluate models on the ArabCulture benchmark using the lmeval [cite] framework, which computes accuracy based on log-likelihood in both multiple-choice (MCQ) and completion settings. For MCQ, the model selects the option with the highest log-likelihood given the prompt, and accuracy reflects agreement with the gold answer. For completion, accuracy is based on the log-likelihood assigned to the gold continuation, measuring the model’s ability to prefer culturally appropriate completions. This unified probabilistic metric captures both discrete selection and generative performance.

%We evaluate the effects of cultural alignment by calculating the country-level accuracy improvements of the culturally aligned models over the baseline models for both multiple-choice (MCQ) and completion settings. We use the \texttt{lm-eval} framework \cite{eval-harness}, which computes accuracy based on log-likelihood. For completion, the log-likelihood assigned to the gold continuation is used to calculate accuracy. To analyze the impact of geographical distance on cultural alignment, we calculate the Pearson correlation between distances and accuracy improvements. We also model cultural similarity using cosine similarity scores and calculate the Pearson correlation between cosine similarity scores and accuracy improvements. In addition to country-level analysis, we also analyze performance by topic and report our findings.

We quantify cultural alignment as the country-level accuracy gain (aligned $-$ baseline) for both MCQ and completion using \texttt{lm-eval} framework \cite{eval-harness}. For completion, accuracy is computed from the log-likelihood of the gold continuation. We then test whether gains track geography and culture by computing Pearson correlations between gains and (i) geographical distance, and (ii) cosine-based cultural similarity. Finally, we report topic-level breakouts.

\section{Results}
Our experiments reveal key findings across the four language models, as evident from Table~\ref{table:main} and Table~\ref{tab:gemma_silma_results}, highlighting accuracy improvements when training on data from one country and evaluating across others. Each cell shows absolute percentage-point gains relative to the respective baseline models.

% \begin{table}[ht]
% \centering
% \small
% \resizebox{\columnwidth}{!}{%
% \begin{tabular}{l|rr|rr}
% \hline
% \textbf{Model} & \multicolumn{2}{c|}{\textbf{DITTO}} & \multicolumn{2}{c}{\textbf{ICL}} \\
% \cline{2-5}
%                & \textbf{Completion} & \textbf{MCQ} & \textbf{Completion} & \textbf{MCQ} \\
% \hline
% Qwen   & 0.272 & 17.564  & 2.878 & 17.654 \\
% Gemma  & 0.728 & 22.750 & 3.174 &  3.602 \\
% ALLaM  & 3.532 &  0.142 & 3.532 & -14.877 \\
% SILMA  & 0.461 &  1.094 & 1.235 &  0.460 \\
% \hline
% \end{tabular}
% }
% \caption{Overall Evaluation Accuracy Improvements over Model Baselines for DITTO and ICL models on Completion and MCQ tasks.}
% \label{tab:ditto_icl_scores}
% \end{table}

\begin{table}[!h]
\centering
\resizebox{\linewidth}{!}{%
\Large
\scriptsize
\renewcommand{\arraystretch}{1.4}  % increase vertical space
\setlength{\tabcolsep}{1.2pt}
\begin{tabular}{l|rrrr|rrrr|r}
\toprule
\multirow{3}{*}{\hspace{3pt}\textbf{Country}\hspace{4pt}} 
& \multicolumn{4}{c|}{\rotatebox{0}{\includegraphics[height=2.5ex]{images/gemma-icon.png}} \textbf{Gemma-2 9B-it}} 
& \multicolumn{4}{c|}{\rotatebox{0}{\includegraphics[height=2.5ex]{images/silma-icon.png}} \textbf{SILMA 9B-Inst}} 
& \multirow{3}{*}{\hspace{4pt}\textbf{Avg.}\hspace{1pt}} \\
\cmidrule{2-9}
& \multicolumn{2}{c}{\textbf{DITTO}} & \multicolumn{2}{c|}{\textbf{ICL}} 
& \multicolumn{2}{c}{\textbf{DITTO}} & \multicolumn{2}{c|}{\textbf{ICL}} 
& \\
\cmidrule{2-9}
& \textbf{Comp.} & \textbf{MCQ} & \textbf{Comp.} & \textbf{MCQ}
& \textbf{Comp.} & \textbf{MCQ} & \textbf{Comp.} & \textbf{MCQ}
& \\
\midrule
Algeria   & 0.44 & 28.71 & \textbf{\textcolor{resultgreen}{3.92}} & 0.00 & $-$0.03 & 1.86 & \textbf{\textcolor{resultgreen}{1.54}} & $-$2.73 & 4.21 \\
Egypt     & $-$0.66 & 37.51 & 2.35 & 5.50 & $-$0.03 & 2.27 & 0.00 & 2.49 & 6.18 \\
Jordan    & 1.10 & 22.78 & 3.58 & 11.37 & 0.63 & 1.98 & 1.19 & 2.20 & 5.60 \\
KSA       & $-$1.26 & 30.88 & 1.54 & 4.46 & $-$0.19 & 0.98 & 0.69 & 0.26 & 4.67 \\
Lebanon   & 0.97 & \underline{\textcolor{black}{32.01}} & \underline{\textcolor{black}{3.52}} & 3.27 & 1.13 & 2.61 & \underline{\textcolor{black}{1.32}} & 3.81 & 6.08 \\
Libya     & $-$0.85 & 20.77 & 2.23 & 2.45 & 0.10 & 2.36 & 1.19 & 1.79 & 3.76 \\
Morocco   & 1.19 & 20.45 & 1.76 & 3.45 & $-$1.10 & \textbf{\textcolor{resultgreen}{3.46}} & 0.69 & 1.89 & 3.97 \\
Palestine & 0.84 & 28.53 & 2.92 & 0.25 & $-$0.34 & 2.55 & 1.13 & 1.64 & 4.69 \\
Sudan     & 0.75 & 31.45 & 3.14 & 8.36 & $-$0.28 & 1.73 & 1.19 & 3.08 & 6.18 \\
Syria     & 0.00 & \textbf{\textcolor{resultgreen}{37.38}} & 2.82 & 8.20 & 0.72 & 0.69 & 0.85 & 2.61 & \underline{\textcolor{black}{6.66}} \\
Tunisia   & $-$1.42 & 21.58 & 2.86 & 1.16 & 0.22 & 1.54 & 0.66 & 2.61 & 3.65 \\
UAE       & 0.50 & 29.06 & 2.42 & 17.97 & $-$0.34 & \underline{\textcolor{black}{3.24}} & 0.10 & 2.45 & \textbf{\textcolor{resultgreen}{6.93}} \\
Yemen     & $-$0.91 & 18.66 & 2.26 & 4.59 & $-$1.10 & $-$0.18 & 0.28 & 1.86 & 3.18 \\
\rowcolor{gray!15}
\specialrule{.1em}{.05em}{.05em}  % Thick line before the Avg row
\rowcolor{gray!10}
\textbf{Avg.} & 0.05 & \textbf{\textcolor{resultgreen}{27.67}} & \textbf{\textcolor{resultgreen}{2.72}} & \underline{\textcolor{black}{5.46}} & $-$0.05 & 1.93 & \underline{\textcolor{black}{0.83}} & 1.84 & ~ \\
\specialrule{.1em}{.05em}{.05em}  % Thick line before the Avg row
\multicolumn{10}{p{1\linewidth}}{\small \vspace{0.005cm} Baselines: Gemma-2 (32.52/34.56), SILMA (32.39/70.81)}
\end{tabular}
}
\caption{Overall accuracy improvements in Arab cultural commonsense reasoning when training on food-based country-specific knowledge across different models. Results show performance on Completion and MCQ tasks using DITTO and ICL methods. \textbf{Bold} and \textcolor{resultgreen}{green} cells indicate top two MCQ and Completion values for each model. \underline{Underlined} marks second best.}
\label{tab:gemma_silma_results}
\end{table}

\paragraph{Strong Cross-Cultural Transfer.} 
% Training on data from a single country culture exhibits a substantial improvement in other Arab cultures, that is, cross-cultural, averaging 18\% improvement in all instances. For example, training on cultural demonstrations from Morocco or the UAE yields 14\%-19\% gains across countries, alignment methods, and samples resulting in a similar cross-cultural effect given the notable geographical distance between them. Notably, Yemen emerges as the strongest source of cultural knowledge, achieving an average improvement of 19\%, followed by Syria at 18.7\%. On the other end of the stick, Palestine, Morocco, Libya, and the UAE are susceptible to yield the highest improvements of 20 to 33\% through cross-cultural transfer. These results suggest that the knowledge learned from localized demonstrations is effectively generalized to broader cultural contexts.
Training on small demonstration sets from a single Arab country consistently improves model performance on other Arab countries, that is, cross-cultural, averaging 2--5\% gains in MCQ and completion tasks across models and methods. Interestingly, Syria as a source country (``teacher'') results in the highest average improvement (5.02\%) across all models and methods, followed by Jordan (4.94\%) and the UAE (4.73\%). Furthermore, Jordan-trained Gemma-2 exhibits strong cross-cultural improvements, yielding a 4.8\% completion gain with ICL and a 33.9\% MCQ gain with DITTO. This cultural transfer occurs despite the geographical and cultural differences between countries, suggesting that cultural knowledge effectively transfers across the Arab region regardless of model architecture. Consistent cross-cultural improvements suggest that these models develop broader Arab cultural understanding rather than just memorizing country-specific features.

\paragraph{Multilingual vs.\ Arabic-centric.} Table~\ref{table:main} shows distinct patterns between multilingual models (Qwen-2.5-7B-Instruct~\cite{qwen2.5}, Gemma-2-9B-It~\cite{team2024gemma}) and Arabic-centric models (ALLaM-7B-Instruct~\cite{bariallam}, SILMA-9B-Instruct~\cite{silma_01_2024}). Given their lower baselines, multilingual models—especially Gemma-2—yield the largest MCQ gains (Gemma-2: baseline 34.56; +22.75 pp with DITTO), surpassing Qwen-2.5 (baseline 51.65; +17.56 pp) and both Arabic-centric models. By contrast, ALLaM shows the strongest improvement on completion (+3.53 pp with ICL) despite its higher baseline (36.35). These results suggest that multilingual models adapt more on culturally grounded MCQ with demonstration-based alignment, whereas Arabic-centric models obtain larger generative gains. Notably, Jordan’s data produces exceptional MCQ gains for Gemma-2 (+33.93 pp), while Syria yields the highest cross-model average improvement (5.02 pp). The pattern persists under food-based sampling (Table~\ref{tab:gemma_silma_results}), though completion gaps narrow; SILMA shows more balanced cross-task gains, indicating that Arabic-centric models benefit from fine-grained domain knowledge.

\begin{figure*}[t]
  \centering
  \includegraphics[width=1\linewidth]{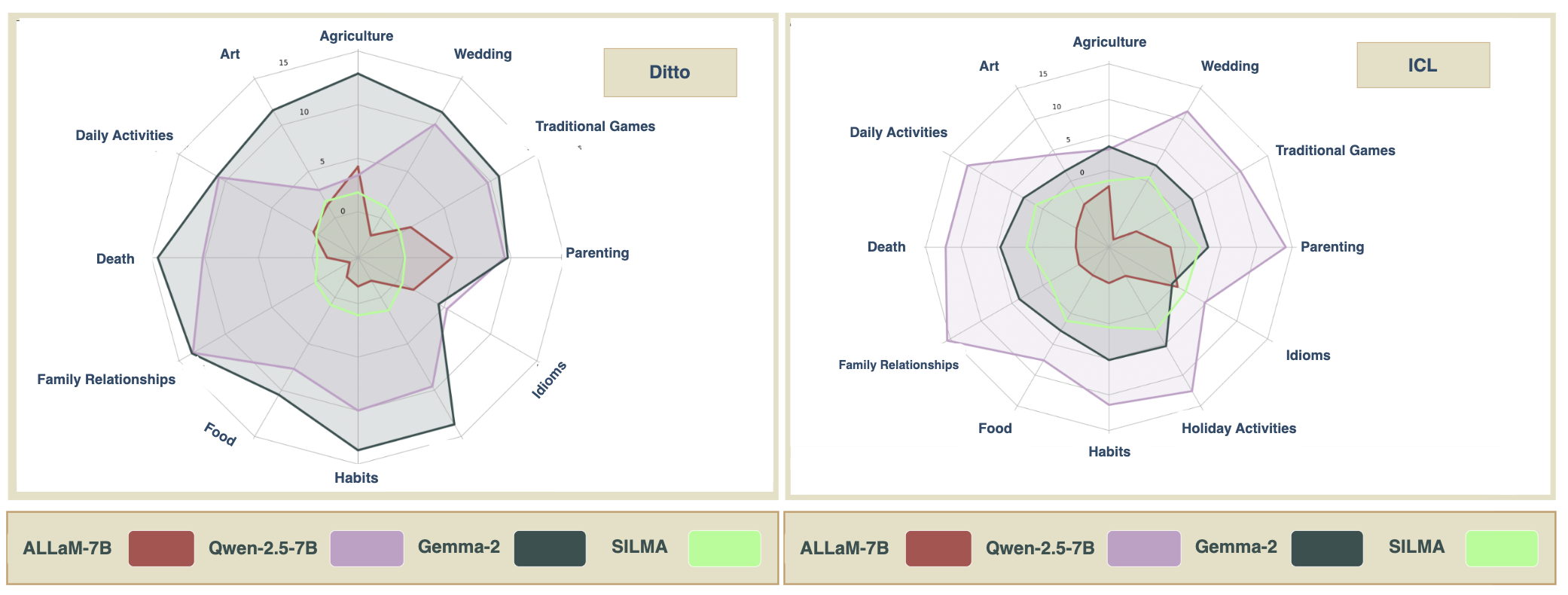}
  \caption{Radar charts comparing topic-level improvements for ICL (left) and DITTO (right) methods across 12 cultural domains. Values represent average improvement in percentage points, with DITTO showing superior performance in most topics, particularly structured social domains like Family Relationships and Agriculture.}
  \label{fig:topics}
\end{figure*}

\paragraph{Performance Comparison of DITTO and ICL.} ICL yields small but consistent gains with few negative transfers, whereas DITTO reaches higher ceilings—especially on MCQ---at the cost of greater variance. On MCQ, DITTO is strongest with multilingual models (e.g., Gemma-2: $+$22.75 pp overall; $+$33.93 pp with Jordan), but occasional negative transfers appear in Arabic-centric settings (e.g., SILMA-Lebanon: $-$3.01 pp), and ICL can also hurt in some cases (ALLaM MCQ: $-$14.88 pp). For completion, ICL consistently outperforms DITTO across models (e.g., ALLaM: $+$3.53 pp with ICL vs.\ $-$1.29 pp with DITTO). Overall, DITTO is preferable for MCQ on multilingual models, while ICL is the safer choice for completion; both are sensitive to small demonstration sets.

This asymmetry suggests that DITTO’s iterative preference updates better suit discriminative MCQ settings in multilingual models, whereas in-context demonstrations more effectively enhance generative completion, particularly for Arabic-centric models. The gap narrows under food-based sampling (Table~\ref{tab:gemma_silma_results}), indicating that domain-specific examples reduce method-dependent variance. Ablations (Appendix~\ref{sec:appendix_negative_transfer}) show that increasing the number of demonstrations lowers DITTO’s MCQ variance and mitigates negative transfer. In practice, choose by task and model: use DITTO to maximize MCQ gains, use ICL for more stable completion, and increase demo counts to stabilize DITTO.

%This asymmetry suggests that iterative optimization benefits discriminative tasks in multilingual models, while in-context demonstration better enhances generative capabilities, especially in Arabic-centric models. The gap between methods narrows in food-based sampling, as demonstrated in Table~\ref{tab:gemma_silma_results}, indicating that domain-specific examples may reduce method-dependent variance. Our targeted experiments showcased in Appendix~\ref{sec:appendix_negative_transfer}, further show that increasing the number of demonstrations reduces the variance of MCQ in DITTO, highlighting practical mitigation strategies. These findings emphasize the need to select alignment methods and demonstration strategies according to both model architecture and task type to amplify gains and limit negative transfer. DITTO is optimal when the main goal is to achieve maximum potential improvement, whereas ICL provides more reliable performance by ensuring a balanced approach.

\paragraph{Transferability with Fine-Grained Sampling.} When alignment data is restricted to a single domain (food), cross-cultural effects remain strong across methods as demonstrated in Table~\ref{tab:gemma_silma_results}. Training on country-specific food-related examples can yield notable accuracy improvements, with Syria and the UAE showing the highest overall average gains (6.66\% and 6.93\% respectively). The results demonstrate asymmetry in knowledge transfer effectiveness. Lebanon consistently performs well as a source of transfer learning, appearing in the top performers for both Gemma-2 and SILMA completion tasks. Notably, MCQ tasks show higher variability, with Gemma-2's DITTO method achieving remarkably strong improvements (averaging 27.67\% across countries), particularly when trained on Syrian and Lebanese examples (37.38\% and 32.01\%, respectively). For completion tasks, Gemma-2 with ICL yields the strongest average improvement (2.72\%), while SILMA benefits more modestly but consistently across methods. These findings indicate that the selection of fine-grained demonstrations fosters robust cross-cultural adaptation, but the degree of reciprocity in knowledge transfer varies substantially by country, model architecture, and assessment method.

\begin{figure*}[t]
  \centering
  \includegraphics[width=0.95\textwidth]{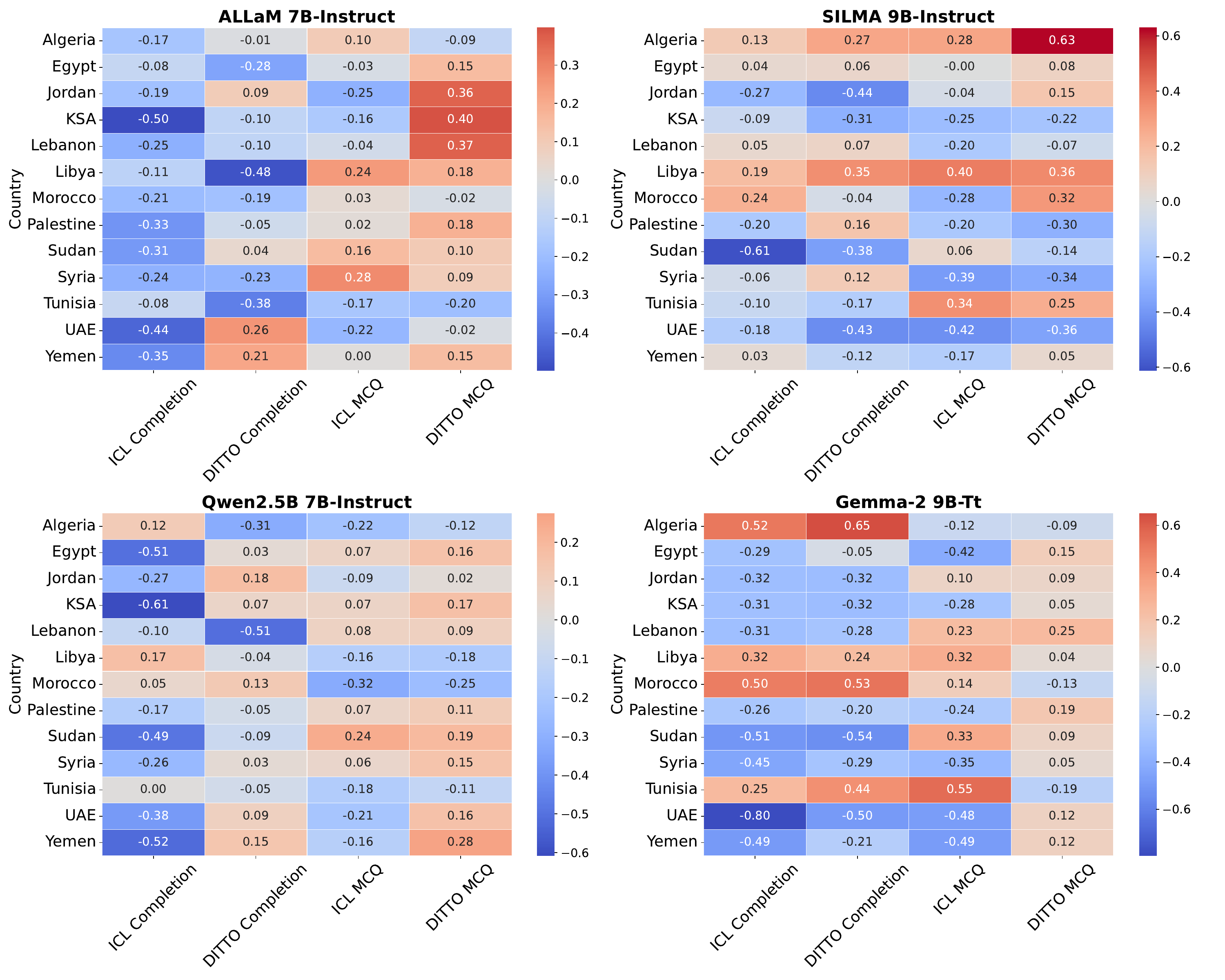}
  \caption{Pearson correlation between geographical distance from the training country and country-level accuracy gain, across four train/eval settings (topic-based sampling).}
  \label{fig:correlations_heatmap}
\end{figure*}

\section{Analysis}
%\subsection{Topic Learnability and Internal Diversity}
\subsection{Topic-wise Transfer} 
Cross-cultural transfer exhibits significant methodological and domain-specific variation, with DITTO achieving superior performance relative to ICL (5.3\% vs.\ 2.3\% average improvement; Figure~\ref{fig:topics}). Cosine similarity analysis across 13 Arab countries reveals uniformly low within-topic similarity scores, providing empirical evidence of highly localized cultural knowledge structures where idioms demonstrate maximum divergence (0.04 average similarity) while agriculture and family relationships exhibit comparatively greater cross-cultural consistency (0.08 and 0.07 respectively; Table~\ref{tab:topic-similarity}). The relationship between cultural diversity and transfer effectiveness reveals unexpected complexity: domains with intuitively high variability do not uniformly yield diminished performance gains, as evidenced by death-related cultural knowledge achieving substantial improvements (+5.7\% DITTO, +2.8\% ICL) and family relationships demonstrating robust transferability (+6.0\% DITTO, +3.1\% ICL), while topics with lowest similarity scores such as idioms (+3.0\% DITTO, +1.6\% ICL) and food practices (+4.2\% DITTO, +1.3\% ICL) produce the most limited gains across both methods. Model architecture fundamentally shapes transfer dynamics: multilingual models (Qwen-2.5, Gemma-2) consistently generate positive improvements across all cultural domains with performance peaks reaching +15.5\% (Qwen-2.5 ICL in family relationships) and +14.4\% (Gemma-2 DITTO in death-related topics), whereas Arabic-centric architectures exhibit pronounced asymmetric responses with ALLaM demonstrating substantial negative ICL sensitivity (ranging from $-$9.5\% in wedding topics to +4.5\% in agriculture) while maintaining consistently positive DITTO performance across all domains, and SILMA achieving modest but stable cross-method improvements, thereby establishing DITTO's superiority for multilingual frameworks and ICL's preferential applicability to Arabic-specialized models when avoiding highly contextual domains.

\subsection{Impact of Geographical Distance on Cross-Cultural Transfer}
Considering that Arab culture is often perceived as homogeneous or grouped geographically into regions (e.g. Gulf, Levant, North Africa, Nile Valley), we used the geographical distances between the capitals of each country shown in Table~\ref{table:distances} in Appendix~\ref{appendix:geo_distance_matrix} and the accuracy improvements per country over the baseline to calculate the Pearson correlation between distance and accuracy improvement for each country to measure the impact of geographical distance on cross-cultural transfer. The average correlation coefficients across all countries and training methods are shown in Table \ref{tab:correlation_means} (Appendix \ref{sec:appendix_correlations}) and a more detailed breakdown of the results for all the models is shown in Figure~\ref{fig:correlations_heatmap}.

% The Pearson correlation results for the Qwen2.5-7B model are shown in Figure \ref{fig:geographic_distance1}, 

% Qwen2.5B 7B-Instruct
% Shows predominantly negative correlations for ICL Completion
% Yemen (-0.52) and Sudan (-0.49) have strong negative correlations for ICL Completion
% Lebanon shows mixed results with negative correlations for completions but positive for MCQs

The results reveal high variation in how the four models perform across the 13 countries using different evaluation methods. The data shows that performance varies not only by country but also by testing approach, with ICL Completion generally producing the most varied results and DITTO MCQ typically showing more positive correlation, as shown in Table \ref{tab:correlation_means}. Notable patterns include the UAE consistently showing negative correlation across most models, while Morocco tends toward positive correlation, particularly with Gemma-2. The Gemma-2 model exhibits the most extreme correlation values, with correlation coefficients ranging from $-$0.8 to 0.65. These disparities likely reflect differences in cultural contexts, and potentially imbalanced training data representation from these regions, highlighting the challenges in developing language models that perform consistently across diverse Arabic-speaking populations.

\subsection{Cultural Similarity}
\label{section:cultural_sim}
We modeled cultural similarity using the 12 broad topics (e.g., food, weddings, holiday activities) defined in the Arab Culture dataset. For each country, we obtain a single country-level embedding that represents all country's cultural data, then using cosine similarity between these embeddings, we quantify cultural similarity between the countries as represented in Table~\ref{table:cultural-similarity} alongside details in the Appendix~\ref{sec:cultural-similarity}.
We then calculated the Pearson correlation between cosine similarity and accuracy improvement for each country for the different models. The detailed results are shown in Figure~\ref{fig:correlations_heatmap_cosine_sim} in Appendix \ref{sec:appendix_correlations_cosine_sim}.
The consistently high similarities (ranging between 0.72--0.89, averaging 0.85) suggest high cultural overlap and help explain transferability between Arab countries. 
The multilingual models exhibit positive average Pearson correlation on MCQ tasks, as shown in Table~\ref{tab:correlation_means_cosine}. This means that higher cosine similarity scores generally correlated with higher accuracy improvements and thus, higher transferability. However, the Arabic models displayed negative Pearson correlation, which might be explained by the chosen sentence embedding model, detailed in Appendix \ref{sec:cultural-similarity}, being multilingual, therefore embeddings used to calculate cosine similarity could be more aligned to the multilingual LLMs' embeddings than the Arabic-centric LLMs.

% This may be because the chosen word embedding model, \texttt{paraphrase-multilingual-MiniLM-L12-v2}, is inherently multilingual and tends to encode Arab cultural samples as more semantically similar to each other, reflecting the generalized patterns seen in multilingual LLMs. As a result, it may not fully capture the finer distinctions between different Arab countries that an Arabic-centric model could identify, thus leading to higher measured similarities and Pearson correlations for multilingual models, but comparatively lower scores for Arabic models that are more attuned to subtle cultural differences.

% \begin{figure}[!ht]
%   \centering
%   \includegraphics[width=\columnwidth]{analysis_graphs/qwen_mcq_completion_topic_corr_cosinesim.pdf}
%   \caption{Pearson Correlation Coefficient between Cosine Similarity and Evaluation Accuracy Improvement for four different train/eval methods (Qwen2.5B 7B-Instruct base model).}
%   \label{fig:similarity_corr}
% \end{figure}

\begin{table}[t]
\centering
\renewcommand{\arraystretch}{1.4}  % increase vertical space

\Large
\resizebox{\columnwidth}{!}{%
\begin{tabular}{l|cc|cc}
\hline
\textbf{Model} 
& \multicolumn{2}{c|}{\textbf{DITTO}} 
& \multicolumn{2}{c}{\textbf{ICL}} \\
\cline{2-5}
& \textbf{Completion} & \textbf{MCQ} 
& \textbf{Completion} & \textbf{MCQ} \\
\hline
ALLaM-7B-Inst & $-$0.191 & $-$0.066 & $-$0.014 & $-$0.188 \\
Qwen-2.5-7B-Inst & \x0.026 & \x0.283 & \x0.184 & \x0.236 \\
SILMA-9B-Inst & $-$0.084 & $-$0.272 & \x0.171 & $-$0.052 \\
Gemma-2-9B-It & $-$0.206 & \x0.326 & \x0.062 & \x0.070 \\
\hline
\end{tabular}
}
\caption{Mean Pearson correlation across countries between cosine similarity and accuracy improvement across models for Completion and MCQ tasks.}
\label{tab:correlation_means_cosine}
\end{table}

% As shown in Figure \ref{fig:geographic_distance1}, aligning on some cultural demonstrations from countries, such as Yemen, Sudan, and KSA, resulted in a positive correlation between distance and accuracy improvement, while other countries, such as Morocco, Libya, and Tunisia, resulted in a negative correlation. In both cases, the correlation was weak, with a maximum correlation coefficient of 0.369 for Yemen (DITTO topic-based training), and a minimum of -0.0054 for Palestine (few-shot topic-based training). The corresponding accuracy improvement vs. distance graphs are shown in Figures \ref{fig:uae_corr} and \ref{fig:yemen_corr}.

% \subsection{Training Beyond Arab Culture}

% \begin{figure*}[t]
%     \centering
%     \includegraphics[width=1\linewidth]{analysis_graphs/region_improvements.png} 
%     \caption{Topic Performance Across Arab Regions. The chart shows performance scores for the top cultural topics across four Arab regions, demonstrating both consistent patterns and subtle regional variations.}
%     \label{fig:region}
% \end{figure*}

\definecolor{allamcolor}{HTML}{00A65D}
\definecolor{qwencolor}{HTML}{8F1ED6}

\subsection{Cross-Cultural Transfer Beyond Arab Culture}
\label{beyond_arab_culture}
Expanding upon exploring the effect of \emph{cross-cultural transfer}, we examined the use of cultural demonstrations beyond the Arab world. We curated 12 demonstrations representing each of the two Indonesian contexts (ID, Aceh \& Papua) curated from the \texttt{IndoCulture} dataset \citep{koto2024} and additional handcrafted demonstrations representing US cultural contexts to evaluate training on cultures beyond the Arab world. We aligned Qwen-2.5-7B and Allam-7B using ICL and \textsc{DiTTO} on cultural contexts and evaluated on Arab cultural commonsense reasoning to measure \emph{cross-cultural transfer} effect. The performance of Indonesian and US demonstrations compared to Arab counterparts is demonstrated in Table~\ref{tab:model-comparison}. 

For Qwen-2.5-7B, MCQ accuracy jumps from 52\% to 69--71\% with ICL and 67--70\% with \textsc{DiTTO}, averaging 69\% and matching the average performance obtained by training on Arab contexts. Completion increases modestly, ID averages 1\% lower than average Arab demonstrations while US scores above in similar magnitude. ALLaM-7B showed similar trends, with improvement in MCQ achieved only using \textsc{DiTTO} with non-Arab contexts. However, in completion, non-Arab contexts exceeded Arab demonstrations using \textsc{DiTTO} whereas ICL performed better with in-culture contexts and delivered the highest completion gain (+4.3\%). These results demonstrate that minimal out-of-culture examples can rival in-culture alignment for MCQ reasoning, though completion generation still benefits the most from culturally proximate demonstrations, underscoring that cultural similarity is helpful but \emph{not} a prerequisite for valuable transfer. Further investigation of out-of-culture demonstrations can be found in Appendix~\ref{sec:indo}.

\begin{table}[t]
\centering
\resizebox{0.84\linewidth}{!}{%
\scriptsize
\setlength{\tabcolsep}{2.5pt}
\begin{tabular}{lrrrrrr}
\toprule
& \multicolumn{3}{c}{\textbf{MCQ Scores (\%)}} & \multicolumn{3}{c}{\textbf{Completion Scores (\%)}} \\
\cmidrule(lr){2-4} \cmidrule(lr){5-7}
\textbf{Context} & \textbf{Base} & \textbf{ICL} & \textbf{DiTTO} & \textbf{Base} & \textbf{ICL} & \textbf{DiTTO} \\
\midrule
\multicolumn{7}{c}{\textcolor{qwencolor}{\textbf{Qwen-2.5-7B-Instruct}}} \\
\midrule
Arab LB & \multirow{5}{*}{51.65} & 63.65 & 66.76 & \multirow{5}{*}{32.89} & 34.34 & 31.98 \\
Papua (ID) & & 71.13 & 67.17 & & 34.49 & 32.14\\
Arab AVG & & 69.30 & 69.21 & & 35.77 & 33.16 \\
Aceh (ID) & & 69.09 & 67.17 & & 34.15 & 32.14 \\
US & & 68.71 & 69.94 & & 35.03 & 34.46 \\
Arab UB & & 71.57 & 70.63 & & 36.88 & 34.53\\
\midrule
\multicolumn{7}{c}{\textcolor{allamcolor}{\textbf{ALLaM-7B-Instruct-preview}}} \\
\midrule
Arab LB & \multirow{5}{*}{69.90} & 44.58 & 57.08 & \multirow{5}{*}{36.35} & 38.99 & 31.51 \\
Papua (ID) & & 71.22 & 71.88 & & 37.76 & 36.44\\
Arab AVG & & 55.02 & 70.04 & & 39.88 & 35.06\\
Aceh (ID) & & 65.63 & 72.60 & & 38.14 & 37.29\\
US & & 65.63 & 73.28 & & 38.71 & 38.91 \\
Arab UB & & 65.06 & 73.61 & & 40.68 & 37.13 \\
\bottomrule
\multicolumn{7}{p{0.8\linewidth}}{\small \vspace{0.05cm}Note: Base scores are constant across contexts. LB \& UB = Lower \& Upper bound. ID contexts (Aceh, Papua) are beyond arab cultures (ID = Indonesian), while Arab contexts represent culturally proximate testing.} \\
\end{tabular}
}
\caption{Performance comparison between Qwen-2.5-7B and ALLaM-7B Models trained on Indonesian and US contexts and evaluated on Arab culture.}
\label{tab:model-comparison}
\end{table}

\subsection{Cultural Representation in Model Latent Space}

\begin{figure}[t]
\centering
\includegraphics[width=\columnwidth]{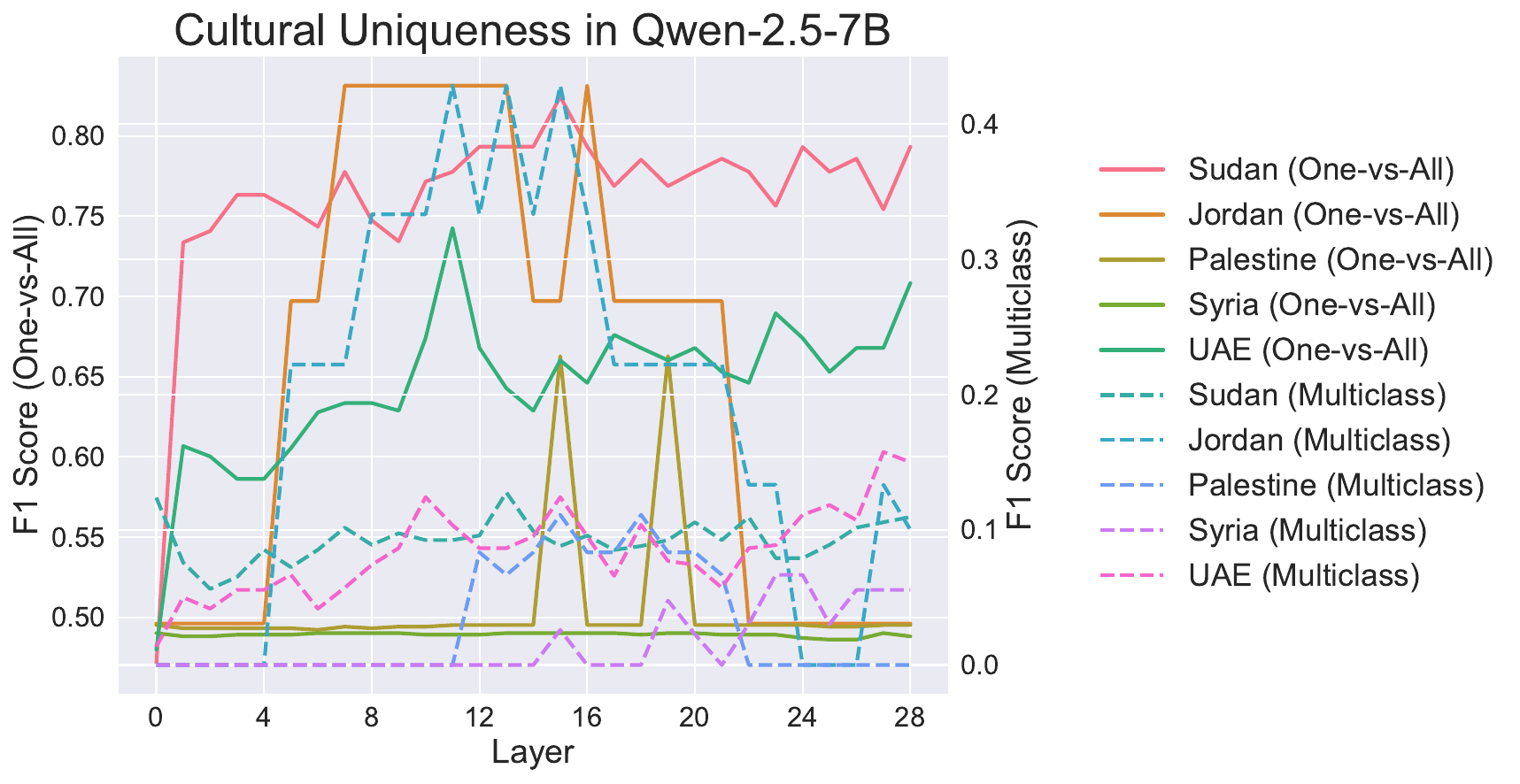}
\caption{F1 scores across model layers for different countries using one-vs-all and multiclass classifiers.}
\label{fig:probing_results}
\end{figure}

To understand how different Arab cultures are internally represented within the model, we conducted a probing analysis across all layers of the Qwen model, using both one-vs-all and multiclass linear classifiers to assess the linear separability of cultural knowledge. The results are illustrated in Figure~\ref{fig:probing_results}. 

In this part, our goal is to assess whether different cultural representations are distinguishable in the model’s latent space. To this end, we adopt layer-wise probing, which enables a direct evaluation of the linear separability of cultural representations across layers. This approach provides clear insights into how cultural knowledge is internally encoded. Details of the probing experiments can be found in Appendix~\ref{probing_details}.

% As shown in Figure~\ref{fig:probing_results}, the representation of cultural knowledge varies considerably across countries. In the base model, countries like Sudan and Jordan exhibit higher linear separability, particularly in the upper layers, achieving one-vs-all F1 scores above 0.75 and 0.8, respectively. This suggests that the model encodes more distinct, separable features for these countries, possibly due to stronger representation or more unique linguistic markers in the pretraining data.

% In contrast, Palestine and Syria show much lower separability, with one-vs-all F1 scores hovering close to random chance (around 0.5) across all layers. This indicates that the model does not learn strongly distinguishable representations for these countries and may generalize them under a broader pan-Arab cultural representation.

% The multiclass probing results further reinforce this pattern, showing low F1 scores (<0.15) for all countries, reflecting the overall difficulty of distinguishing multiple cultures simultaneously in the latent space. Nonetheless, Sudan and Jordan maintain relatively higher multiclass F1 scores compared to other countries, aligning with their stronger one-vs-all performance.

% Overall, these results suggest that the Qwen model selectively encodes cultural uniqueness for some countries, while treating others as less distinct entities in its latent representation space. This selective encoding likely reflects differences in training data distribution, linguistic features, or cultural salience within the pretraining corpus.

\begin{figure}[t]
\centering
\includegraphics[width=\columnwidth]{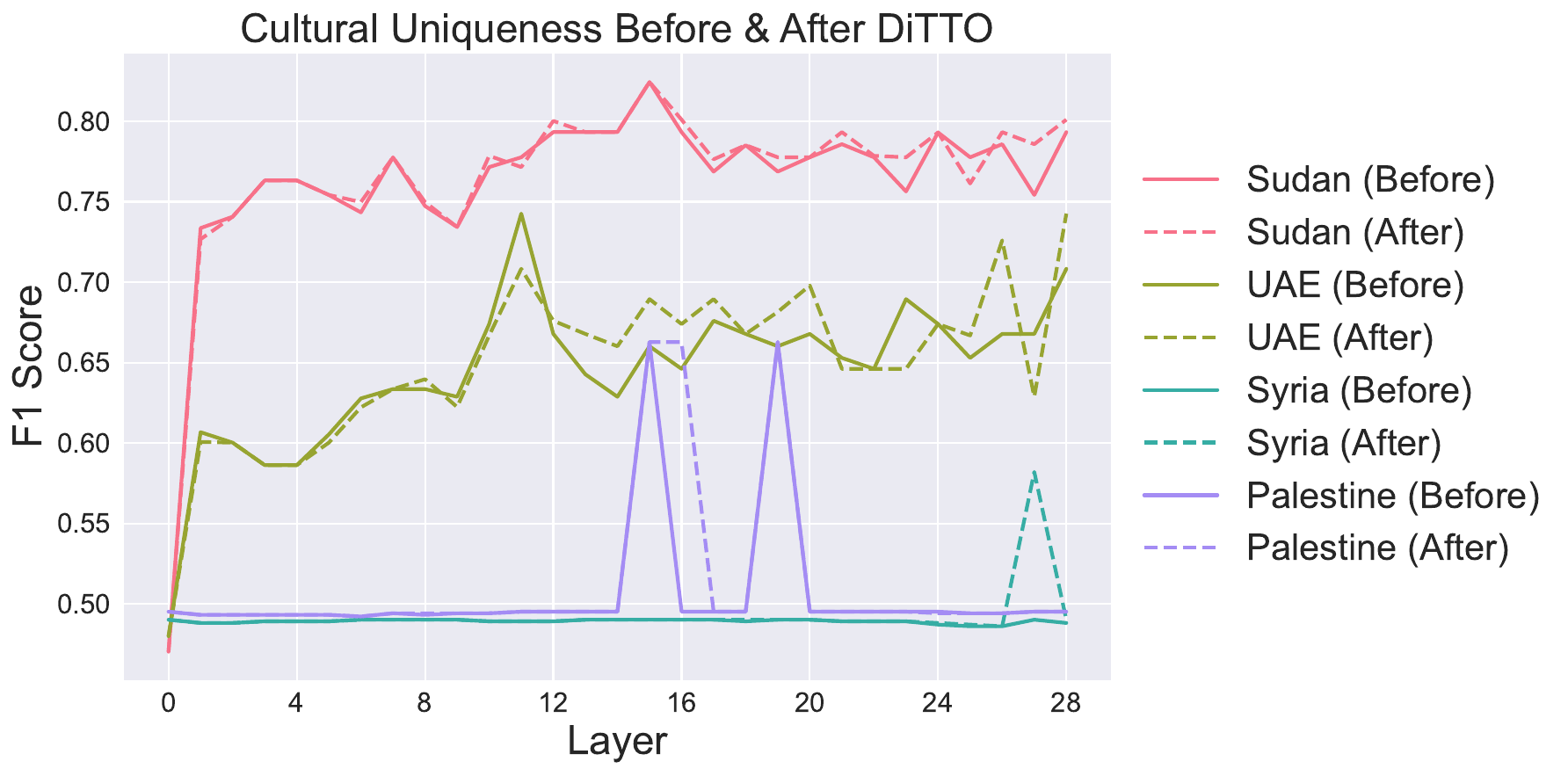}
\caption{F1 scores across model layers for Sudan, UAE, Syria, and Palestine before and after UAE-specific alignment on Qwen-2.5-7B.}
\label{fig:uae_alignment}
\end{figure}

Our probing analysis shown in Figures \ref{fig:probing_results} and \ref{fig:uae_alignment} reveals that Qwen-2.5-7B encodes Arab cultures with varying distinctness, showing high linear separability for Sudan and Jordan but much lower for Palestine and Syria. Multiclass probing confirms the difficulty of jointly distinguishing multiple cultures, though Sudan and Jordan remain relatively more separable. 
The results of other models can be found in Appendix~\ref{cultural_repre}, and they show similar distinctness. In other models, the separability for Jordan, Sudan, and the UAE is relatively high, which may originate from their adopting a lapped pre-training corpus having more knowledge of these countries. 
After UAE-specific alignment, only the UAE showed improved cultural encoding, while other countries remained largely unchanged, yet reasoning performance improved across all countries. This suggests that targeted cultural alignment can enhance specific representations while indirectly benefiting generalization, offering a viable path toward culturally adaptive NLP systems.

As to why the inference effects vary greatly between countries, one factor is the imbalance in the LLM’s pretraining corpus, where countries with larger populations and greater digital presence are more likely to be represented extensively in pretraining data~\citep{dunn}.
The layer-wise probing results indicate that even before alignment, each country already exhibits a degree of cultural representation in the model’s internal layers as demonstrated in Figure~\ref{fig:probing_results}. The alignment process then shifts or reinforces these representations, which we observe as changes in cultural separability across layers showcased in Figure~\ref{fig:uae_alignment}. These findings suggest that prior representation strength, driven by pretraining exposure, affects a country’s ability to serve as an effective source.

% To further investigate how cultural alignment affects the model’s internal representations, we compared the cultural separability before and after UAE-specific alignment. Figure~\ref{fig:uae_alignment} shows the one-vs-all F1 scores across model layers for selected countries before and after the tuning.

% As shown in Figure~\ref{fig:uae_alignment}, UAE is the only country that exhibits a clear improvement in linear separability after alignment, with noticeable increases in F1 scores across middle and upper layers. This indicates that the model has learned more unique, country-specific knowledge about UAE culture through the alignment process.

% In contrast, the separability of other countries such as Sudan, Syria, and Palestine remained largely unchanged, suggesting that the alignment process primarily affected the representation of UAE culture without substantially altering the encoding of other cultures. This selective improvement implies that targeted cultural alignment can enhance the cultural uniqueness of underrepresented countries while preserving the broader cultural generalization learned by the base model.

% Notably, although only UAE showed an increase in separability, the model’s overall reasoning performance across countries also improved. This suggests that enriching the model’s knowledge of UAE culture indirectly support broader pan-Arab commonsense reasoning.

% These findings highlight the potential of targeted cultural alignment techniques to enrich the model’s representation of specific cultural contexts without sacrificing generalization across similar cultures. This mechanism offers a promising pathway for developing culturally adaptive natural language processing (NLP) systems that balance specificity and generalizability.

\section{Conclusion}
Our study demonstrates that LLMs can effectively achieve cross-cultural adaptation using lightweight alignment methods such as ICL and DITTO, producing consistent gains in Arab countries—even with limited and culturally specific data.The results of our experiments indicate strong cross-cultural transfer, where training on one country’s dataset can significantly improve accuracy in other countries, with gains often exceeding 15–20\% across multilingual models. Some country pairs show modest gains even at large distances, while others see minimal improvement despite close proximity, suggesting that cultural proximity is not strictly tied to geographic location. In contrast, we observed an overall positive correlation between cultural similarity and accuracy improvement across multilingual models, suggesting greater transferability between countries that are more culturally similar. Probing analyses further show that targeted alignment enhances cultural encoding (e.g., for the UAE) without harming overall performance, highlighting the feasibility and benefits of culturally adaptive NLP in multilingual settings. 

In general, our results highlight that lightweight alignment methods can effectively align on the cultural commonsense reasoning task by incorporating region-specific cultural demonstrations. Whether through ICL or DITTO, LLMs can learn robust cultural representations that transfer to new countries. Therefore, this work reinforces the notion that cross-cultural adaptation is feasible and beneficial in multilingual NLP settings, particularly in the Arab world.

\section{Limitations}
\label{sec:Limitations}
While our findings illuminate promising insights into pathways of cross-cultural transfer, several critical limitations constrain the scope of our conclusions.
\paragraph{Task Diversity.} Our primary focus was on the evaluation of cultural multiple choice questions and a completion task. The realm of open-ended tasks (e.g., dialogue, narrative generation) introduces additional layers of complexity for cross-cultural alignment, underscoring the necessity for a deeper investigation into how cultural knowledge extrapolates across open-ended text generation.

\paragraph{Country Coverage} While there are 22 countries that are members of the Arab League, the data set we use only represents 13 of them, which although more representative than other datasets, still does not completely represent the Arab world. This further underscores the point we bring up in the introduction about the discrepancies in data availability by country, and emphasizes the importance of investigating cross-cultural transfer in low-resource settings.

% \paragraph{Model and Method Generality.} \mena{remove this section from limitations since we have 4 models}We focused on Qwen2.5 7B-Instruct, coupled with two alignment methodologies (ICL, DITTO). While these strategies notably enhanced cultural performance, experimenting with different LLM architectures or alternative preference optimization approaches might yield different results. Testing on a wider range of model sizes and families, and experimenting with various alignment methods could strengthen generalizability claims. 

\paragraph{Fine-Grained Cultural Nuances.} Our analysis highlights performance variations even within topic categories, such as family relationships and idioms. In practice, cultural norms can be more nuanced and context-dependent than captured by any small demonstration set. A larger set of demonstrations and supervised fine-tuning may be required to mastering the intricacy of cultural knowledge that required memorization. 

Despite these constraints, our work demonstrates that meticulously chosen examples, irrespective of being derived from broad topics or targeted domains, can improve performance in varied cultural settings. These findings pave the way for future work that refines cross-cultural alignment strategies and investigates the interplay between linguistic diversity, cultural distance and similarity in multilingual NLP.

% \clearpage
\bibliography{custom}
\bibliographystyle{acl_natbib}

%\clearpage

\appendix
\section{Technical Implementation Details}

\subsection{DITTO Configuration}

We followed the standard DITTO implementation provided in the original paper ~\cite{shaikh2024showdonttellaligning}, with these default hyperparameters now explicitly listed for transparency: \textbf{LoRA} (rank 32, $\alpha$=64); \textbf{SFT} (batch 4, LR $3 \times 10^{-5}$); \textbf{DPO} (batch 24 via $8\times3$ rescale, LR $1 \times 10^{-6}$, $\beta$=0.05, 40 grad steps); \textbf{DITTO-specific} (10 negatives, resample every 10 steps at temp 1.0; comparison sampling: 0.7 expert/0.2 replay/0.1 inter-model). This clarification does not affect our methodological setup or conclusions, as our reported improvements rely on the adaptation approach itself, not optimization variations.

\begin{table*}[t]
\Large
\renewcommand{\arraystretch}{1.2}
\centering
\resizebox{\textwidth}{!}{
\begin{tabular}{lccccccccccccc}
\toprule
\rowcolor{gray!20}
\textbf{From/To} & \textbf{Morocco} & \textbf{Algeria} & \textbf{Tunisia} & \textbf{Libya} & \textbf{Egypt} & \textbf{Sudan} & \textbf{Palestine} & \textbf{Jordan} & \textbf{Syria} & \textbf{Lebanon} & \textbf{KSA} & \textbf{UAE} & \textbf{Yemen} \\
\midrule
Morocco   & 0     & 948   & 1,569 & 1,859 & 3,596 & 4,435 & 3,913 & 3,968 & 3,958 & 3,876 & 5,234 & 5,946 & 5,493 \\
Algeria   & 948   & 0     & 630   & 1,016 & 2,706 & 3,755 & 2,996 & 3,048 & 3,027 & 2,944 & 4,340 & 5,032 & 4,695 \\
Tunisia   & 1,569 & 630   & 0     & 518   & 2,090 & 3,245 & 2,368 & 2,419 & 2,397 & 2,314 & 3,717 & 4,403 & 4,117 \\
Libya     & 1,859 & 1,016 & 518   & 0     & 1,739 & 2,753 & 2,077 & 2,135 & 2,148 & 2,071 & 3,377 & 4,098 & 3,680 \\
Egypt     & 3,596 & 2,706 & 2,090 & 1,739 & 0     & 1,596 & 432   & 494   & 613   & 485   & 1,639 & 2,363 & 2,104 \\
Sudan     & 4,435 & 3,755 & 3,245 & 2,753 & 1,596 & 0     & 1,794 & 1,821 & 1,997 & 2,027 & 1,738 & 2,426 & 1,201 \\
Palestine & 3,913 & 2,996 & 2,368 & 2,077 & 432   & 1,794 & 0     & 63    & 213   & 234   & 1,369 & 2,036 & 2,039 \\
Jordan    & 3,968 & 3,048 & 2,419 & 2,135 & 494   & 1,821 & 63    & 0     & 177   & 219   & 1,328 & 1,984 & 2,027 \\
Syria     & 3,958 & 3,027 & 2,397 & 2,148 & 613   & 1,997 & 213   & 177   & 0     & 86    & 1,408 & 2,019 & 2,170 \\
Lebanon   & 3,876 & 2,944 & 2,314 & 2,071 & 485   & 2,027 & 234   & 219   & 86    & 0     & 1,494 & 2,107 & 2,240 \\
KSA       & 5,234 & 4,340 & 3,717 & 3,377 & 1,638 & 1,738 & 1,369 & 1,328 & 1,408 & 1,494 & 0     & 773   & 1,070 \\
UAE       & 5,946 & 5,032 & 4,403 & 4,098 & 2,363 & 2,426 & 2,036 & 1,984 & 2,019 & 2,107 & 773   & 0     & 1,467 \\
Yemen     & 5,493 & 4,695 & 4,117 & 3,680 & 2,104 & 1,201 & 2,039 & 2,027 & 2,170 & 2,240 & 1,070 & 1,467 & 0     \\
\bottomrule
\end{tabular}
}
\caption{Geographical distances (km) between Arab country capitals used for correlation analysis with cross-cultural transfer performance.}
\label{table:distances}  % Changed from table:distances2
\end{table*}

\begin{table*}[t]
\Large
\renewcommand{\arraystretch}{1.2}
\centering
\resizebox{\textwidth}{!}{
\begin{tabular}{lcccccccccccccc}
\toprule
\rowcolor{gray!20}
 & \textbf{Morocco} & \textbf{Algeria} & \textbf{Tunisia} & \textbf{Libya} & \textbf{Egypt} & \textbf{Sudan} & \textbf{Palestine} & \textbf{Jordan} & \textbf{Syria} & \textbf{Lebanon} & \textbf{KSA} & \textbf{UAE} & \textbf{Yemen} & \textbf{Avg.*} \\
\midrule
Morocco & 1.00 & 0.89 & 0.84 & 0.92 & 0.87 & 0.95 & 0.92 & 0.76 & 0.83 & 0.81 & 0.85 & 0.84 & 0.79 & 0.86 \\ 
Algeria & 0.89 & 1.00 & 0.77 & 0.91 & 0.92 & 0.92 & 0.94 & 0.91 & 0.94 & 0.71 & 0.95 & 0.91 & 0.94 & 0.89 \\ 
Tunisia & 0.84 & 0.77 & 1.00 & 0.83 & 0.76 & 0.84 & 0.82 & 0.70 & 0.74 & 0.85 & 0.71 & 0.76 & 0.68 & 0.77 \\ 
Libya & 0.92 & 0.91 & 0.83 & 1.00 & 0.85 & 0.93 & 0.95 & 0.85 & 0.88 & 0.73 & 0.88 & 0.86 & 0.86 & 0.87 \\ 
Egypt & 0.87 & 0.92 & 0.76 & 0.85 & 1.00 & 0.89 & 0.90 & 0.89 & 0.89 & 0.72 & 0.93 & 0.93 & 0.91 & 0.87 \\ 
Sudan & 0.95 & 0.92 & 0.84 & 0.93 & 0.89 & 1.00 & 0.93 & 0.79 & 0.86 & 0.79 & 0.87 & 0.86 & 0.85 & 0.87 \\ 
Palestine & 0.92 & 0.94 & 0.82 & 0.95 & 0.90 & 0.93 & 1.00 & 0.88 & 0.89 & 0.77 & 0.92 & 0.91 & 0.88 & 0.89 \\ 
Jordan & 0.76 & 0.91 & 0.70 & 0.85 & 0.89 & 0.79 & 0.88 & 1.00 & 0.92 & 0.62 & 0.94 & 0.90 & 0.94 & 0.84 \\ 
Syria & 0.83 & 0.94 & 0.74 & 0.88 & 0.89 & 0.86 & 0.89 & 0.92 & 1.00 & 0.64 & 0.92 & 0.87 & 0.93 & 0.86 \\ 
Lebanon & 0.81 & 0.71 & 0.85 & 0.73 & 0.72 & 0.79 & 0.77 & 0.62 & 0.64 & 1.00 & 0.66 & 0.73 & 0.61 & 0.72 \\ 
KSA & 0.85 & 0.95 & 0.71 & 0.88 & 0.93 & 0.87 & 0.92 & 0.94 & 0.92 & 0.66 & 1.00 & 0.92 & 0.96 & 0.88 \\ 
UAE & 0.84 & 0.91 & 0.76 & 0.86 & 0.93 & 0.86 & 0.91 & 0.90 & 0.87 & 0.73 & 0.92 & 1.00 & 0.91 & 0.87 \\ 
Yemen & 0.79 & 0.94 & 0.68 & 0.86 & 0.91 & 0.85 & 0.88 & 0.94 & 0.93 & 0.61 & 0.96 & 0.91 & 1.00 & 0.85 \\
\bottomrule
\end{tabular}
}
\caption{Cosine similarity matrix between countries based on cultural embeddings. Higher values indicate greater cultural similarity. *Diagonal values (1.0) excluded from average calculation.}
\label{table:cultural-similarity}  % Fixed: removed space in label
\end{table*}

\subsection{Geographical Distance Matrix}
\label{appendix:geo_distance_matrix}

Table~\ref{table:distances} shows the approximate distances (in kilometers) between the capitals of the 13 Arab countries used in our correlation analysis (Section 5.3). These distances were calculated using the Haversine formula based on geographical coordinates of each capital city.

\subsection{Cultural Similarity}
\label{sec:cultural-similarity}Table~\ref{table:cultural-similarity} presents the cosine similarity matrix between countries based on cultural embeddings computed using \texttt{paraphrase-multilingual-MiniLM-L12-v2}. These similarities were calculated by averaging topic-level embeddings across the 12 cultural domains for each country. Specifically, for each country, we first obtained the sentence-level embeddings for each cultural sample and found the average embedding per topic. We then averaged across all topics to get a single average embedding per country. Then calculated the cosine similarity for each country with the other countries to represent cultural similarity between countries and average across them as presented in the country columns and last column of Table \ref{table:cultural-similarity}, respectively.

The high similarity values (0.72--0.89, averaging 0.85) indicate substantial cultural overlap across Arab countries, supporting the feasibility of cross-cultural transfer. Notably, Lebanon shows the lowest average similarity (0.72), while Palestine and Algeria show the highest (0.89), though these differences do not strongly predict transfer effectiveness as discussed in Section~\ref{section:cultural_sim}.

\begin{table}[t]
\centering
\small
\begin{tabular}{lcccc}
\toprule
\textbf{Country} & \textbf{Agriculture} & \textbf{Family} & \textbf{Food} & \textbf{Idioms} \\
\midrule
Algeria    & 0.09 & 0.08 & 0.05 & 0.05 \\
Egypt      & 0.09 & 0.07 & 0.07 & 0.04 \\
Jordan     & 0.13 & 0.06 & 0.07 & 0.03 \\
KSA        & 0.10 & 0.05 & 0.08 & 0.04 \\
Lebanon    & 0.07 & 0.08 & 0.06 & 0.04 \\
Libya      & 0.08 & 0.09 & 0.08 & 0.04 \\
Morocco    & 0.09 & 0.05 & 0.05 & 0.03 \\
Palestine  & 0.09 & 0.04 & 0.05 & 0.03 \\
Sudan      & 0.07 & 0.10 & 0.05 & 0.06 \\
Syria      & 0.06 & 0.05 & 0.08 & 0.03 \\
Tunisia    & 0.06 & 0.08 & 0.05 & 0.04 \\
UAE        & 0.07 & 0.09 & 0.05 & 0.05 \\
Yemen      & 0.07 & 0.08 & 0.06 & 0.04 \\
\midrule
\textbf{Average} & \textbf{0.08} & \textbf{0.07} & \textbf{0.06} & \textbf{0.04} \\
\bottomrule
\end{tabular}
\caption{Within-topic cosine similarity scores across 13 Arab countries for selected cultural domains. Lower scores indicate greater cross-cultural diversity within the topic.}
\label{tab:topic-similarity}
\end{table}

\section{Topic-Level Cultural Similarity Analysis}
\label{app:topic-similarity}

Referenced in the main discussion (Section 5), we hypothesized that topics like idioms and food are inherently diverse and context-dependent, making them harder to transfer across cultures compared to more structured domains like family relationships or agriculture, which we validated through a comprehensive cosine similarity analysis examining within-topic cultural consistency across all 13 Arab countries in our dataset. For each cultural topic, we computed pairwise cosine similarities between training examples from different countries within the same topical domain by embedding each cultural statement using a multilingual sentence transformer and calculating the average cosine similarity between all country pairs for each topic, where lower similarity scores indicate greater cross-cultural diversity within that topic while higher scores suggest more consistent cultural patterns across the Arab region. Our quantitative analysis, presented in Table~\ref{tab:topic-similarity}, provides strong empirical support for our theoretical framework, demonstrating a clear hierarchy of cultural consistency where idioms exhibit the lowest average similarity (0.04), followed by food practices (0.06), while family relationships (0.07) and agriculture (0.08) demonstrate higher internal consistency across countries, with this pattern holding remarkably stable across all 13 countries and idioms showing the most restricted similarity range (0.03-0.06) while agriculture displays the greatest variability (0.06-0.13) yet maintains the highest average. These findings provide quantitative evidence that topics involving richer cultural nuances and linguistic specificity (idioms, food) require more extensive memorization and contextual grounding, making them less amenable to lightweight alignment approaches, while more structured domains with clearer cross-cultural regularities (family relationships, agriculture) demonstrate greater transferability potential, aligning with our observed performance patterns in the main results.

\section{Detailed Topic-Level Performance Analysis}
\label{app:topic-performance}

This section provides comprehensive performance breakdowns for both DITTO and ICL alignment methods across all 12 cultural topics and 4 language models, supporting the topic-wise transfer analysis presented in Section 5 and corresponding to the radar chart visualizations in Figure~\ref{fig:topics}, where performance improvements were calculated as percentage point differences between aligned model accuracy and baseline model accuracy for each cultural topic using country-specific demonstrations from the \texttt{ArabCulture} dataset. The detailed results reveal several critical patterns: DITTO demonstrates superior average performance (+5.3\%) compared to ICL (+2.3\%) while exhibiting lower variance and fewer negative transfers, with multilingual models consistently outperforming Arabic-centric models across both alignment methods where Qwen-2.5 and Gemma-2 show complementary strengths, particularly with Gemma-2 achieving the highest individual improvements in death-related topics (+14.4\%) and family relationships (+13.6\%) under DITTO, while Qwen-2.5 demonstrates exceptional ICL performance reaching +15.5\% in family relationships, whereas Arabic-centric models show pronounced variability with ALLaM exhibiting consistent negative transfers across most ICL topics (ranging from -2.1\% to -9.5\%) while maintaining positive DITTO performance, and SILMA achieving modest but stable improvements across both methods. Topic-specific effects are pronounced across both alignment approaches, with family relationships and parenting showing the highest transferability, death-related topics demonstrating substantial improvements despite cultural variation, while idioms and food practices exhibit the most limited improvements, and model architecture significantly influences alignment method effectiveness, suggesting that Arabic-centric models may require different optimization strategies for cultural alignment tasks compared to multilingual architectures that demonstrate robust cross-topic adaptability.

\section{Correlation between Geographical Distance and Accuracy Improvement}
\label{sec:appendix_correlations}
\begin{figure}[t]
  \centering
  \includegraphics[width=\columnwidth]{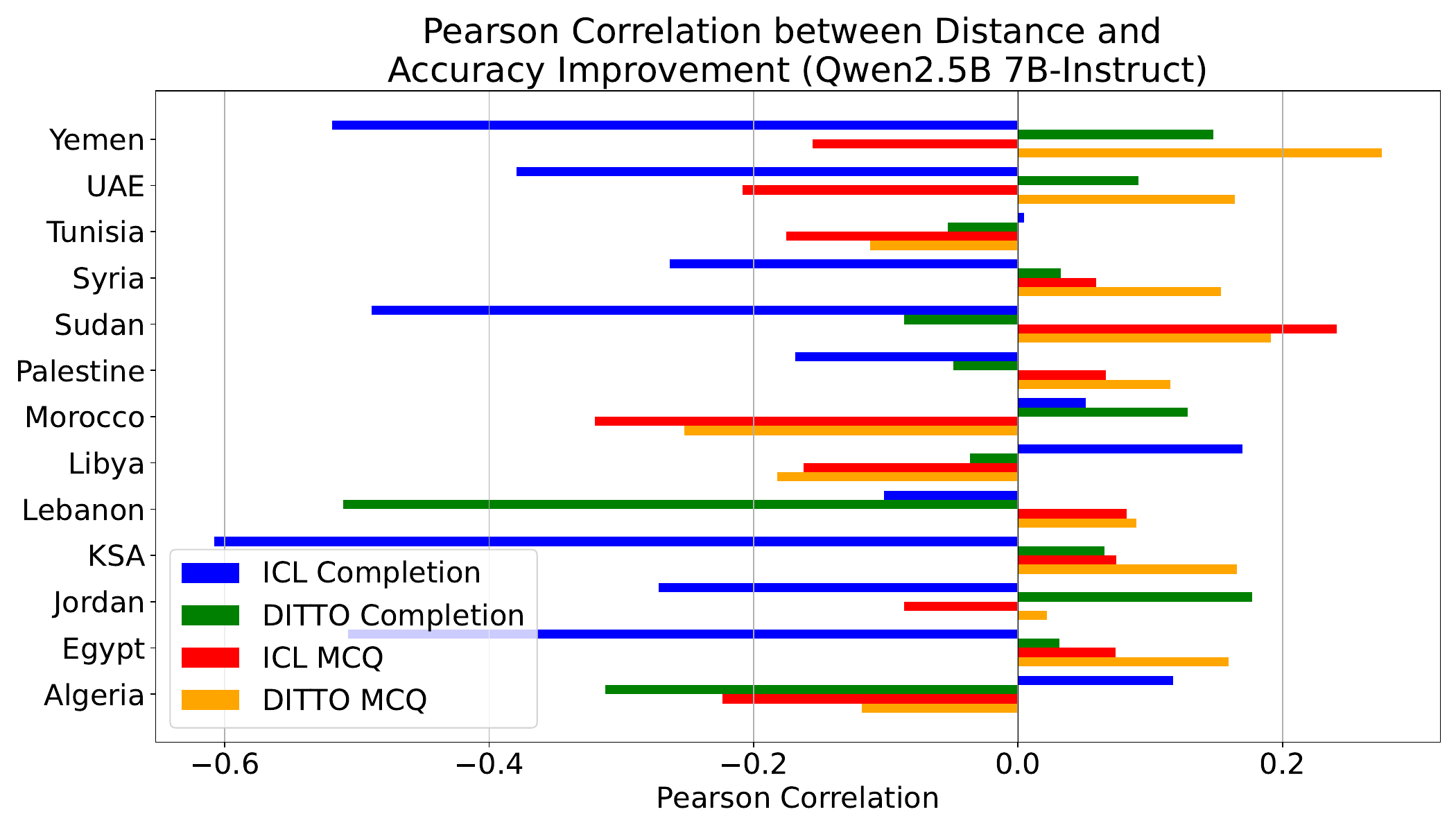}
  \caption{Pearson correlation coefficient between distance from training country and evaluation accuracy improvement for four different train/eval methods (Qwen-2.5-7B-Instruct base model).}
  \label{fig:geographic_distance1}
\end{figure}

\begin{figure}[t]
  \centering
  \includegraphics[width=\columnwidth]{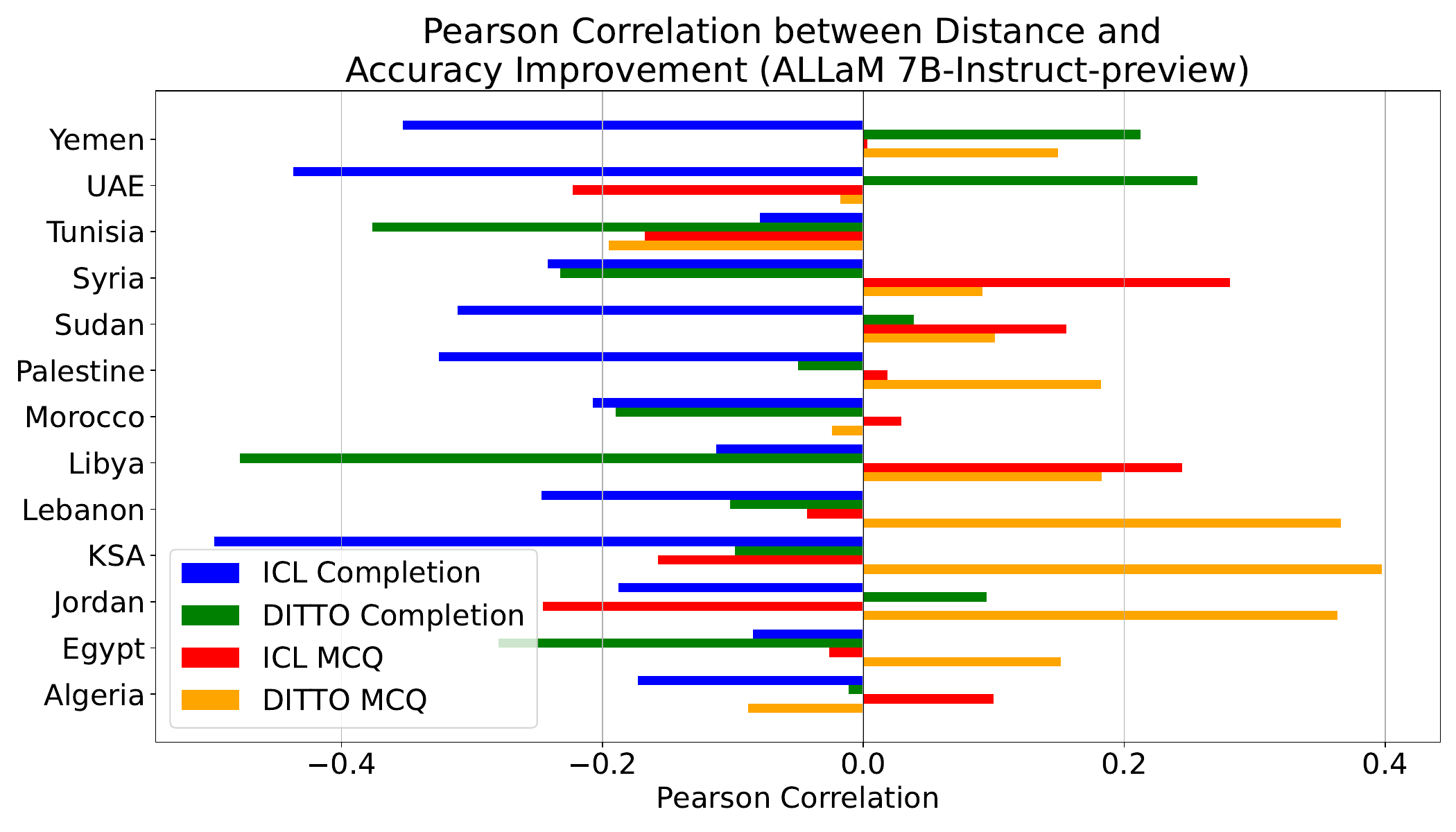}
  \caption{Pearson correlation coefficient between distance from training country and evaluation accuracy improvement for four different train/eval methods (ALLaM-7B-Instruct-preview base model).}
  \label{fig:geographic_distance2}
\end{figure}

\begin{figure}[t]
  \centering
  \includegraphics[width=\columnwidth]{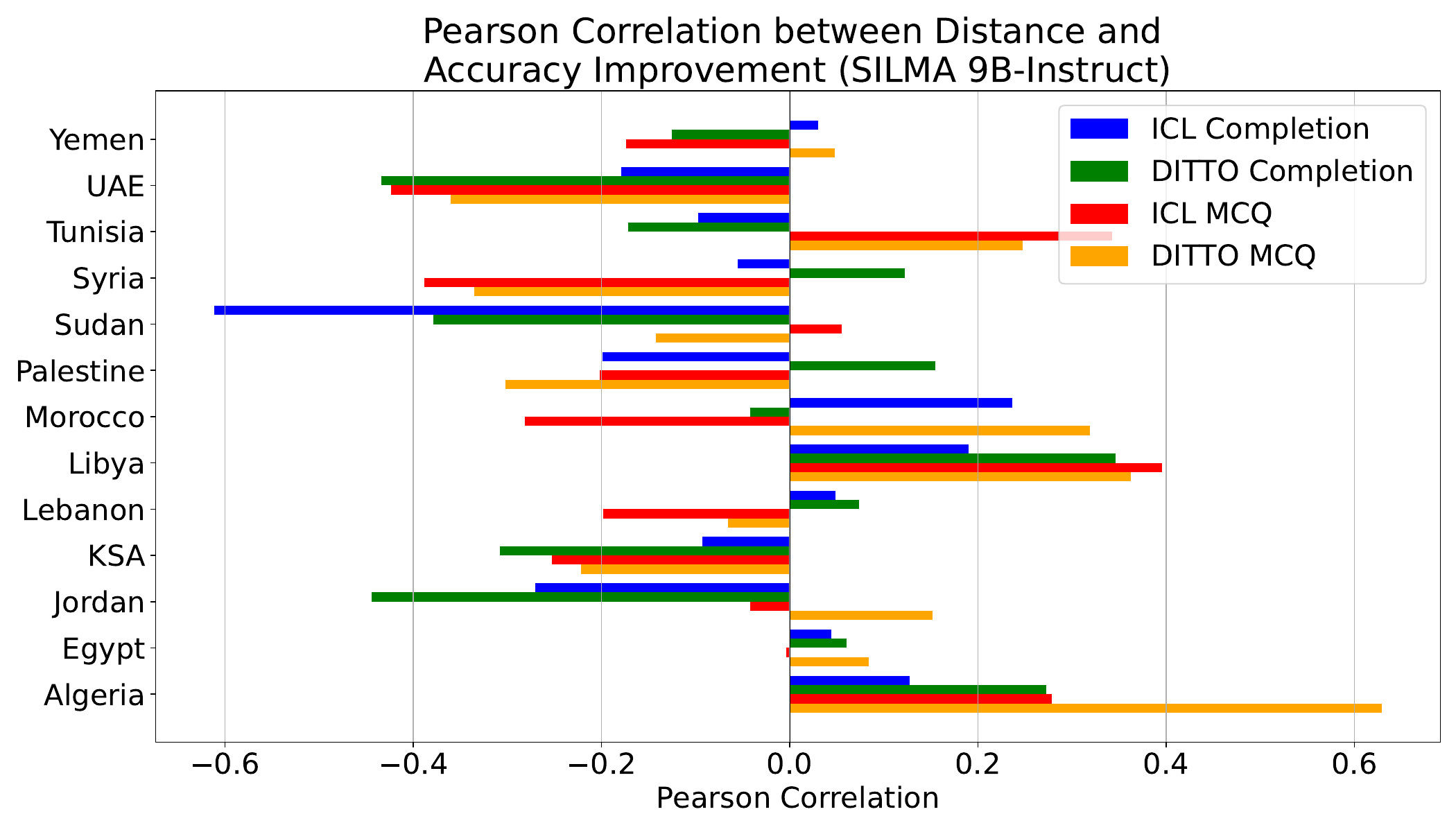}
  \caption{Pearson correlation coefficient between distance from training country and evaluation accuracy improvement for four different train/eval methods (SiLMA-9B-Instruct base model).}
  \label{fig:geographic_distance3}
\end{figure}

\begin{figure}[t]
  \centering
  \includegraphics[width=\columnwidth]{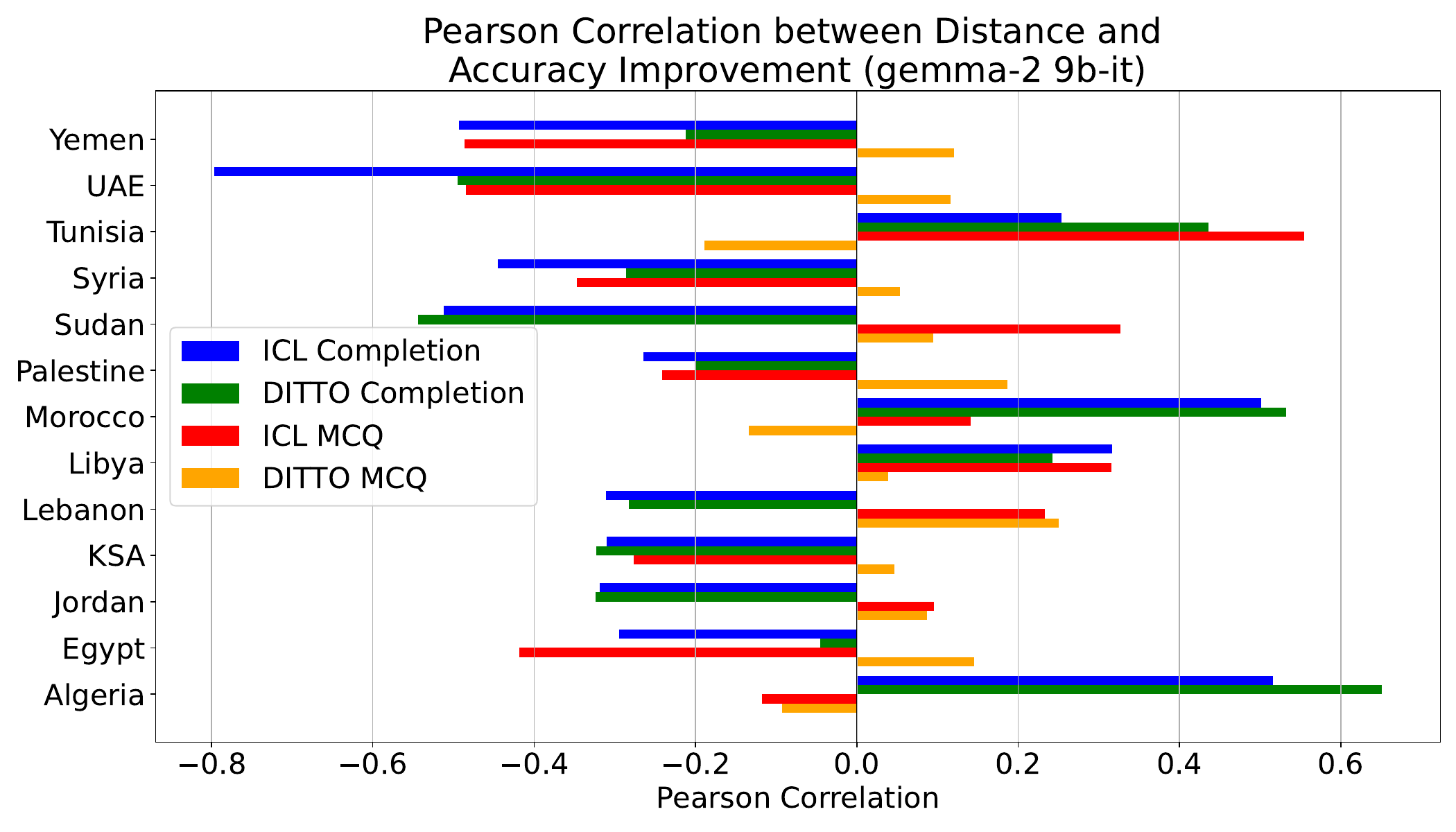}
  \caption{Pearson correlation coefficient between distance from training country and evaluation accuracy improvement for four different train/eval methods (Gemma-2-9B-It base model).}
  \label{fig:geographic_distance4}
\end{figure}

\begin{figure}[t]
  \centering
  \includegraphics[width=0.9\columnwidth]{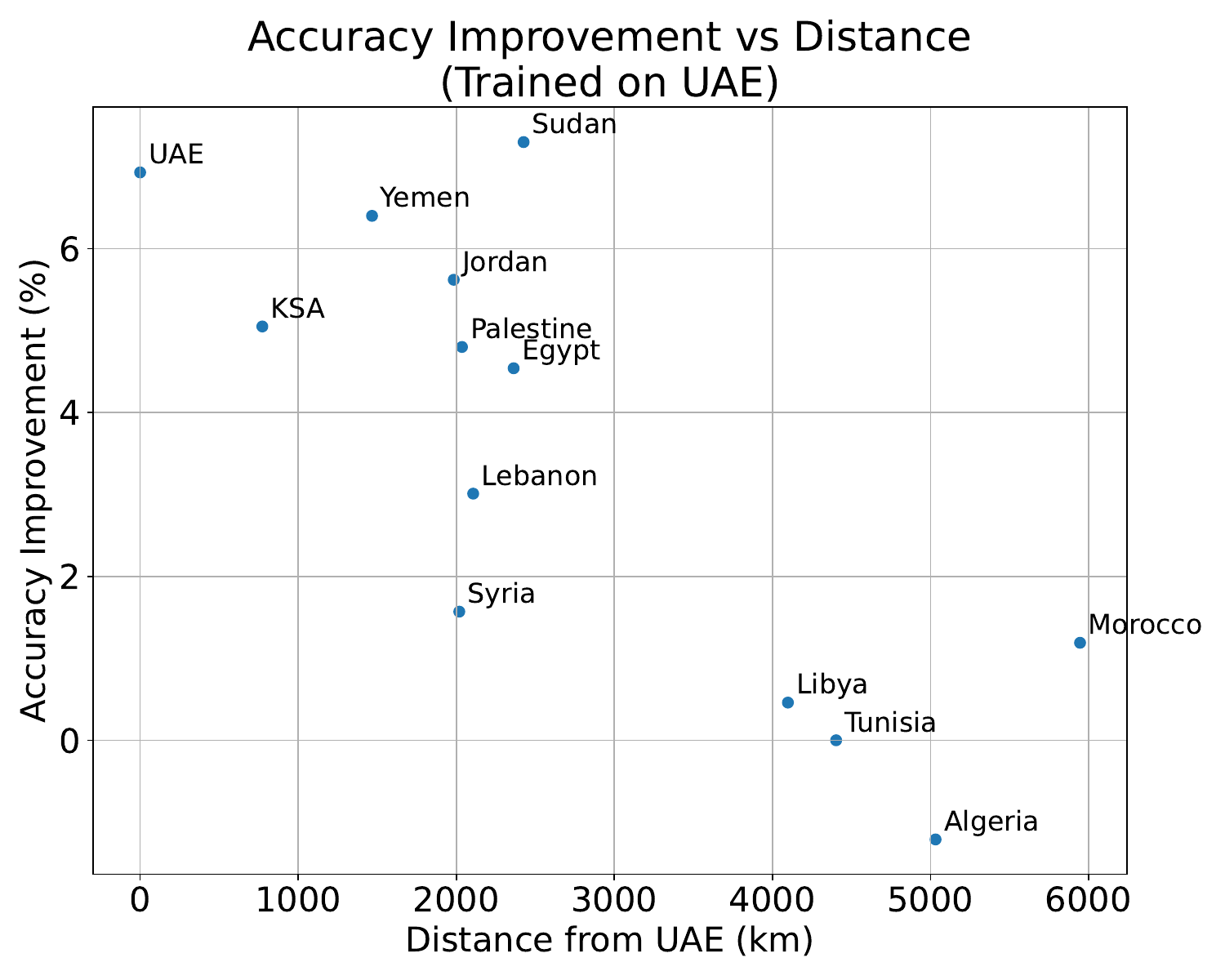}
  \caption{Evaluation accuracy improvement vs.\ distance for ICL topic-based training on samples from the UAE with Completion evaluation (Gemma-2-9B-It). Pearson correlation = $-$0.797.}
  \label{fig:uae_corr}
\end{figure}

\begin{figure}[t]
  \centering
  \includegraphics[width=0.9\columnwidth]{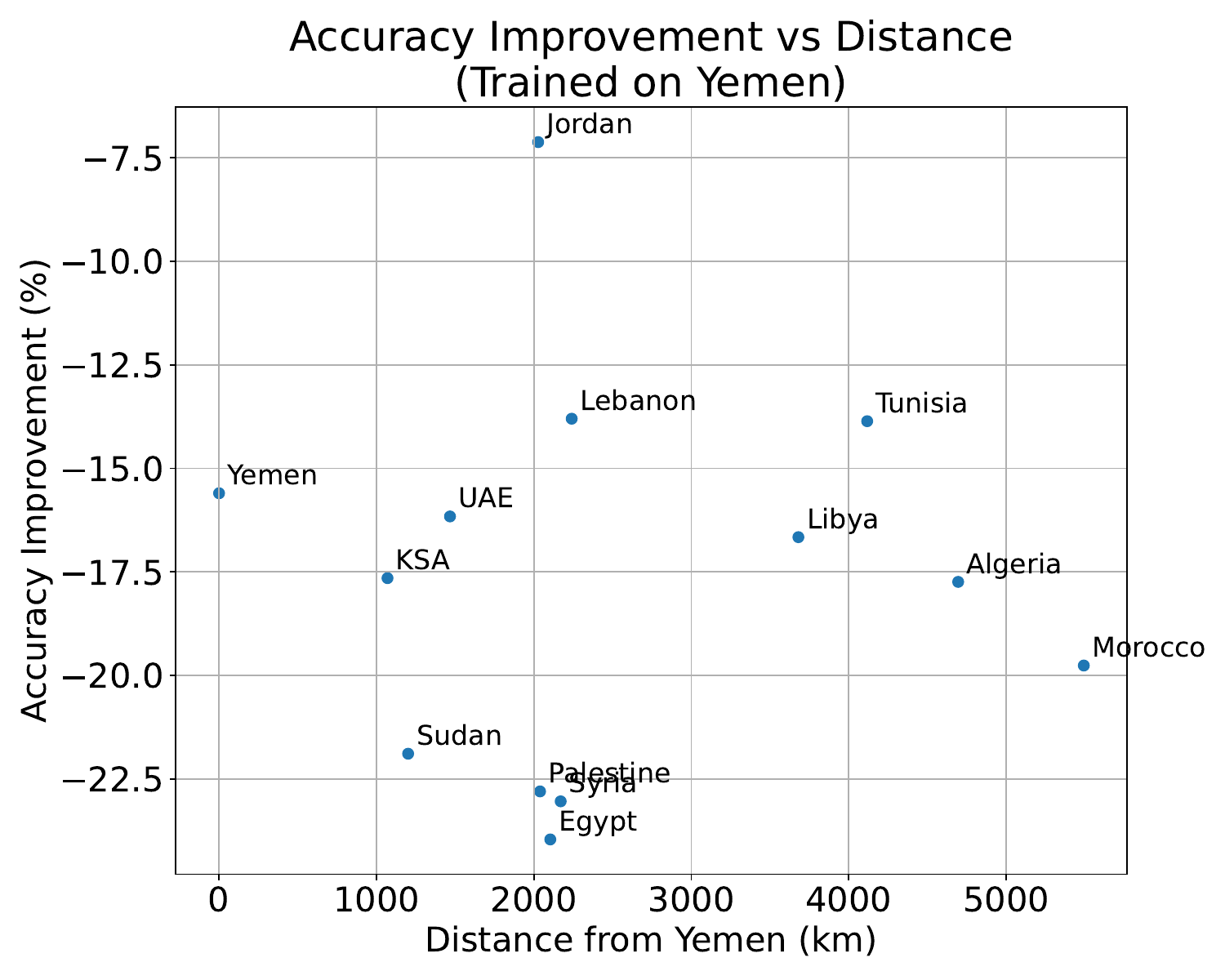}
  \caption{Evaluation accuracy improvement vs.\ Distance for ICL topic-based training on samples from Yemen with MCQ Evaluation (ALLaM-7B-Instruct-preview). Pearson correlation = 0.003.} 
  \label{fig:yemen_corr}
\end{figure}

Table \ref{tab:correlation_means} and Table \ref{tab:correlation_medians} show the overall means and  median correlation scores across models and settings. Figures \ref{fig:geographic_distance1}, \ref{fig:geographic_distance2},  \ref{fig:geographic_distance3}, and \ref{fig:geographic_distance4} show the correlation scores for the 4 models (Qwen-2.5-7B-Instruct, ALLaM-7B-Instruct, SILMA-9B-Instruct, and Gemma-2-9B-It), which are also displayed in heatmaps in Figure \ref{fig:correlations_heatmap}. 
% Additionally, Figure \ref{fig:geographic_distance5} shows the correlation scores averaged across the Arabic models ALLaM 7B-Instruct and SILMA 9B-Instruct, while Figure \ref{fig:geographic_distance6} shows the correlation scores averaged across the multilingual models Qwen2.5 7B-Instruct and Gemma-2 9B-It. 
To demonstrate what the correlation looks like, Figure \ref{fig:uae_corr} shows the accuracy improvement vs.\ distance graph for the strongest correlation, while Figure \ref{fig:yemen_corr} is for the weakest correlation.

%\begin{figure*}[t]
%  \centering
%  \includegraphics[width=\textwidth]{analysis_graphs/correlation_heatmaps.pdf}
%  \caption{Pearson Correlation Coefficient between Distance from Training Country and Evaluation Accuracy Improvement for four different train/eval methods with topic-based sampling.}
%  \label{fig:correlations_heatmap}
%\end{figure*}

\begin{table}[t]
\centering
\renewcommand{\arraystretch}{1.4}  % increase vertical space

\Large
\resizebox{\columnwidth}{!}{%
\begin{tabular}{l|cc|cc}
\hline
\textbf{Model} 
& \multicolumn{2}{c|}{\textbf{DITTO}} 
& \multicolumn{2}{c}{\textbf{ICL}} \\
\cline{2-5}
& \textbf{Completion} & \textbf{MCQ} 
& \textbf{Completion} & \textbf{MCQ} \\
\hline
ALLaM-7B-Inst & $-$0.094 &  0.128 & $-$0.251 & $-$0.002 \\
Qwen-2.5-7B-Inst  & $-$0.029 &  0.052 & $-$0.228 & $-$0.056 \\
SILMA-9B-Inst & $-$0.067 &  0.032 & $-$0.064 & $-$0.069 \\
Gemma-2-9B-it & $-$0.065 &  0.056 & $-$0.166 & $-$0.054 \\
\hline
\end{tabular}
}
\caption{Mean Pearson correlation across countries between distance and accuracy improvement across models for Completion and MCQ tasks.}
\label{tab:correlation_means}
\end{table}

\begin{table}[t]
\renewcommand{\arraystretch}{1.3}  % increase vertical space

\centering
\small
\resizebox{\columnwidth}{!}{%
\begin{tabular}{l|cc|cc}
\hline
\textbf{Model} 
& \multicolumn{2}{c|}{\textbf{DITTO}} 
& \multicolumn{2}{c}{\textbf{ICL}} \\
\cline{2-5}
& \textbf{Completion} & \textbf{MCQ} 
& \textbf{Completion} & \textbf{MCQ} \\
\hline
ALLaM-7B-Inst & $-$0.099 &  \x0.149 & $-$0.242 &  \x0.003 \\
Qwen-2.5-7B-Inst  &  \x0.031 &  \x0.115 & $-$0.264 & $-$0.086 \\
SILMA-9B-Inst & $-$0.042 &  \x0.048 & $-$0.055 & $-$0.174 \\
Gemma-2-9B-It & $-$0.212 &  \x0.087 & $-$0.310 & $-$0.118 \\
\hline
\end{tabular}
}
\caption{Median Pearson correlation between distance and accuracy improvement across models for Completion and MCQ tasks.}
\label{tab:correlation_medians}
\end{table}

\section{Correlation between Cultural Similarity and Accuracy Improvement}
\label{sec:appendix_correlations_cosine_sim}
\begin{figure*}[t]
  \centering
  \includegraphics[width=\textwidth]{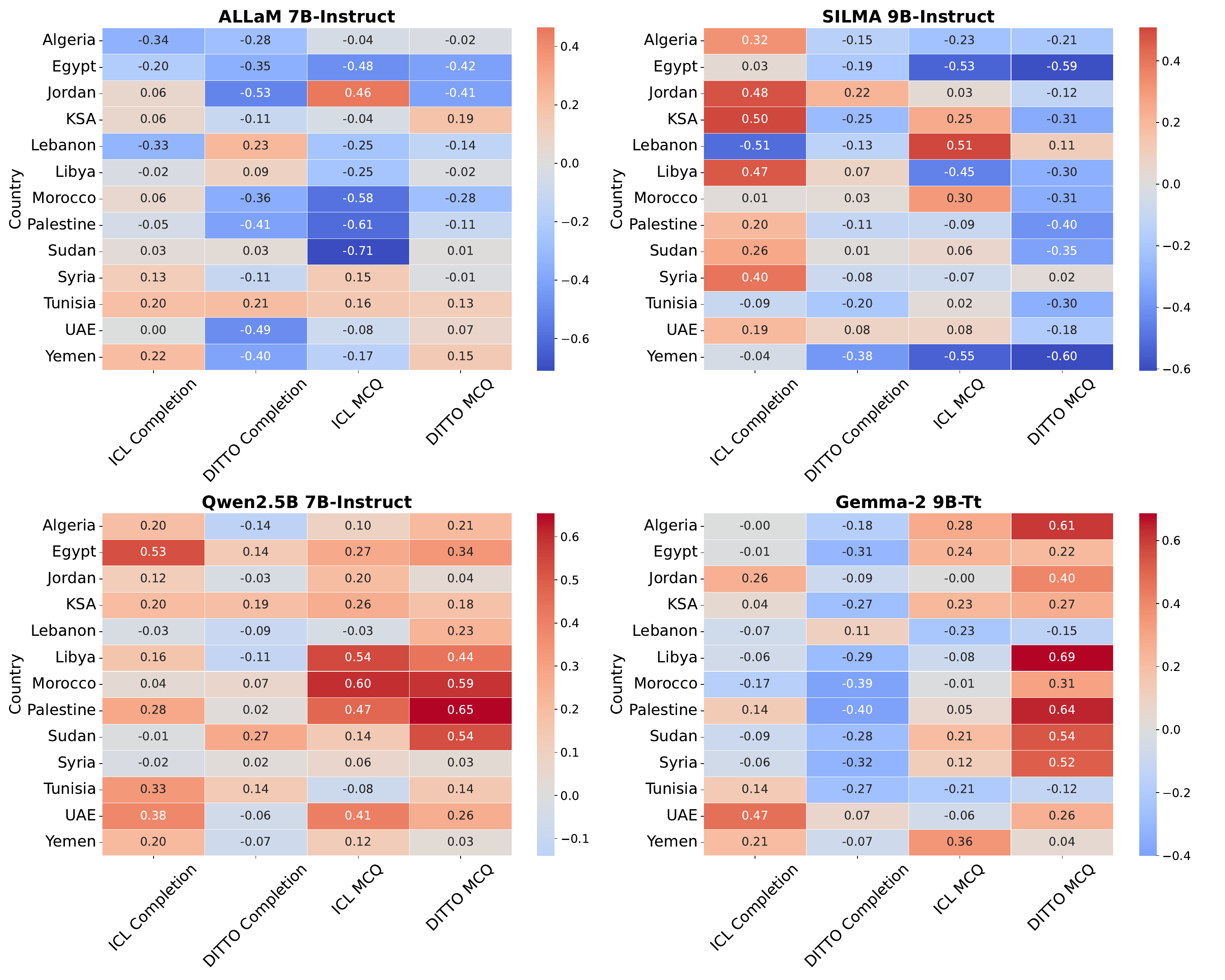}
  \caption{Pearson correlation between cosine-based cultural similarity to the training country and country-level accuracy gain across four train/eval settings (topic-based sampling).}
  \label{fig:correlations_heatmap_cosine_sim}
\end{figure*}

Table \ref{tab:correlation_medians_cosine} shows the overall median correlation scores across models and settings to supplement the means in Table \ref{tab:correlation_means_cosine}. Figure \ref{fig:correlations_heatmap_cosine_sim} shows the correlation scores for all the models across all the countries.

\begin{table}[H]
\renewcommand{\arraystretch}{1.3}  % increase vertical space

\centering
\small
\resizebox{\columnwidth}{!}{%
\begin{tabular}{l|cc|cc}
\hline
\textbf{Model} 
& \multicolumn{2}{c|}{\textbf{DITTO}} 
& \multicolumn{2}{c}{\textbf{ICL}} \\
\cline{2-5}
& \textbf{Completion} & \textbf{MCQ} 
& \textbf{Completion} & \textbf{MCQ} \\
\hline
ALLaM-7B-Inst   & -0.2825 & -0.0152 &  0.0262 & -0.1653 \\
Qwen-2.5-7B-Inst &  0.0158 &  0.2304 &  0.1951 &  0.1963 \\
SILMA-9B-Inst   & -0.1067 & -0.3007 &  0.1964 &  0.0191 \\
Gemma-2-9B-It   & -0.2711 &  0.3135 & -0.0042 &  0.0547 \\
\hline
\end{tabular}
}
\caption{Median Pearson correlation between distance and accuracy improvement across models for Completion and MCQ tasks.}
\label{tab:correlation_medians_cosine}
\end{table}

\section{Negative Transfer Effects}
\label{sec:appendix_negative_transfer}
Due to the nature of ditto iterative alignment, especially with small demonstration sets, can result in sensitivity to intermediate model outputs and the initial demonstrations selected, causing variable transfer effects. To investigate this, we conducted an additional targeted experiment increasing demonstrations from 12 (topic-based) to 100 (randomly sampled), summarized in Table~\ref{tab:variance_comparison}.

We find that increasing demonstrations reduces MCQ transfer variability (1.13→0.48) and maintains the same low variability in completion tasks (0.62→0.66), confirming our hypothesis that larger demonstration sets improve robustness. However, our primary motivation remains data-efficient alignment and potential mitigation strategies (e.g., careful curation or slightly increasing samples) can help control variability and negative transfer. Detailed scores breakdown are listed in Table~\ref{tab:qwen_sample_variance}.

\begin{table}[t]
\centering
\small
\begin{tabular}{lcc}
\toprule
\textbf{Setup} & \textbf{MCQ} & \textbf{Completion} \\
\midrule
12 topic-based demos & 1.13 & 0.62 \\
100 random demos     & 0.48 & 0.66 \\
\bottomrule
\end{tabular}
\caption{Variance comparison between topic-based and random demonstration setups for MCQ and Completion tasks using Ditto on Qwen-2.5-7B-Instruct.}
\label{tab:variance_comparison}
\end{table}

\begin{table*}[h]
\centering
\small
\resizebox{\linewidth}{!}{
\begin{tabular}{lcccc}
\toprule
\textbf{Trained on (country)} & \textbf{MCQ (Topic, n=12)} & \textbf{Completion (Topic)} & \textbf{MCQ (100 Random)} & \textbf{Completion (100 Random)} \\
\midrule
\textbf{Base (No training)} & 51.65 & 32.89 & 51.42 & 32.36 \\
Algeria   & 68.39 & 33.08 & 71.13 & 30.52 \\
Egypt     & 69.21 & 32.80 & 69.89 & 31.39 \\
Jordan    & 70.56 & 32.58 & 70.76 & 32.58 \\
KSA       & 69.49 & 32.83 & 71.45 & 32.95 \\
Lebanon   & 70.03 & 33.80 & 71.04 & 30.75 \\
Libya     & 66.76 & 32.77 & 70.62 & 32.31 \\
Morocco   & 68.90 & 34.53 & 70.67 & 32.08 \\
Palestine & 69.62 & 32.86 & 70.99 & 31.81 \\
Sudan     & 70.63 & 33.96 & 70.49 & 33.27 \\
Syria     & 68.65 & 33.87 & 71.40 & 31.62 \\
Tunisia   & 68.83 & 32.08 & 71.36 & 31.39 \\
UAE       & 68.49 & 33.96 & 68.93 & 31.81 \\
Yemen     & 70.22 & 31.98 & 70.44 & 31.16 \\
\midrule
\textbf{Average}  & \textbf{69.21} & \textbf{33.16} & \textbf{70.70} & \textbf{31.82} \\
\textbf{Variance} & \textbf{1.13}  & \textbf{0.62}  & \textbf{0.48} & \textbf{0.66} \\
\bottomrule
\end{tabular}
}
\caption{Effect of sample count on performance variability on Qwen-2.5-7B-Instruct using DITTO, comparing topic-based (12 samples) vs random (100 samples). Note: Performance scores are listed for variance calculation. Comparing scores between sampling methods is not equivalent due to differing unseen evaluation sets.}
\label{tab:qwen_sample_variance}
\end{table*}

\section{Cultural Representation in Models}\label{cultural_repre}
\label{sec:appendix_rep}
% \begin{figure}[H]
% \centering
% \includegraphics[width=0.5\textwidth]{figures/report_qwen.pdf}
% \caption{F1 scores across model layers for different countries using one-vs-all and multiclass classifiers.}
% \label{fig:probing_results1}
% \end{figure}

\begin{figure}[H]
\centering
\includegraphics[width=\columnwidth]{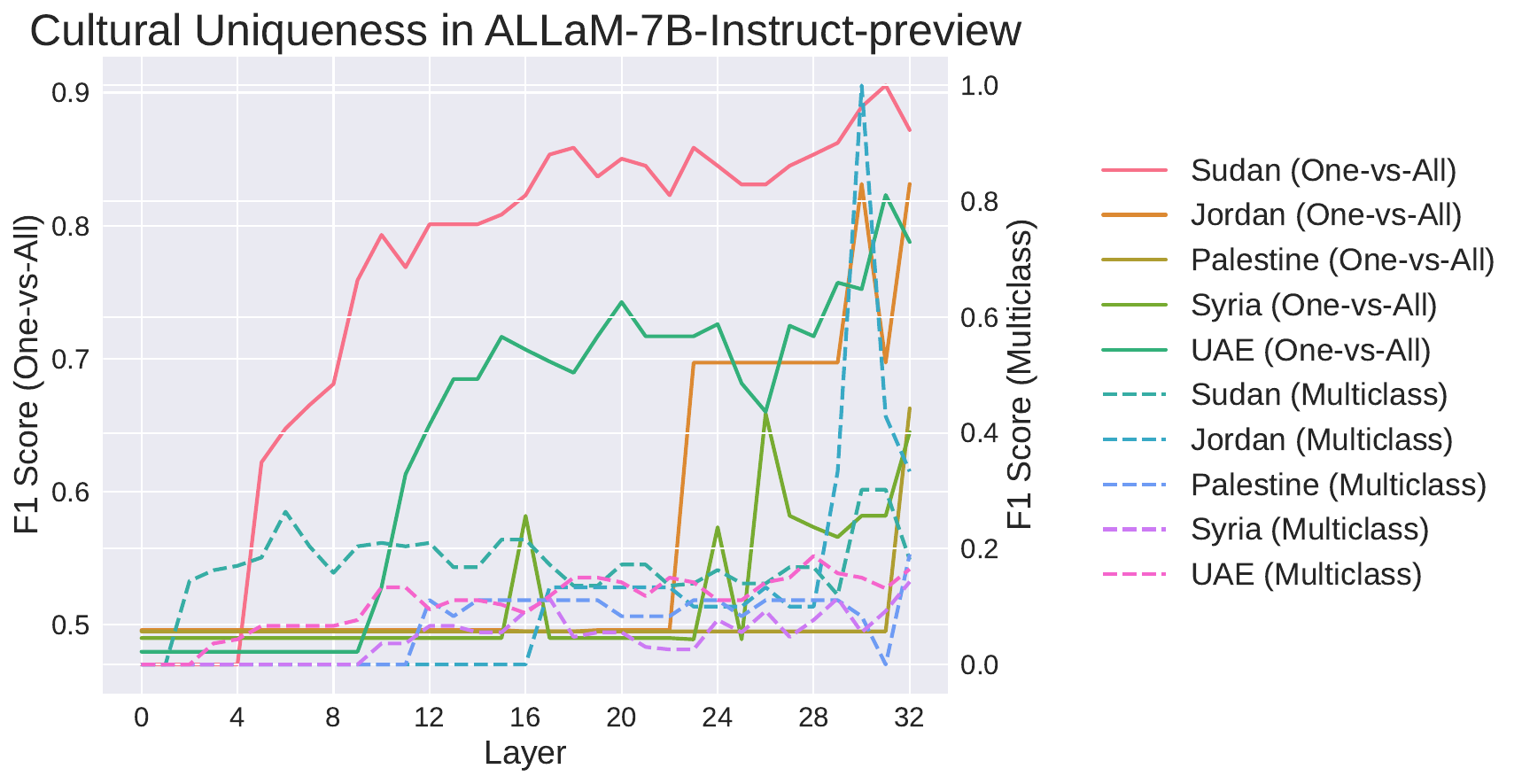}
\caption{F1 scores across model layers for different countries using one-vs-all and multiclass classifiers.}
\label{fig:probing_results2}
\end{figure}
\begin{figure}[H]
\centering
\includegraphics[width=\columnwidth]{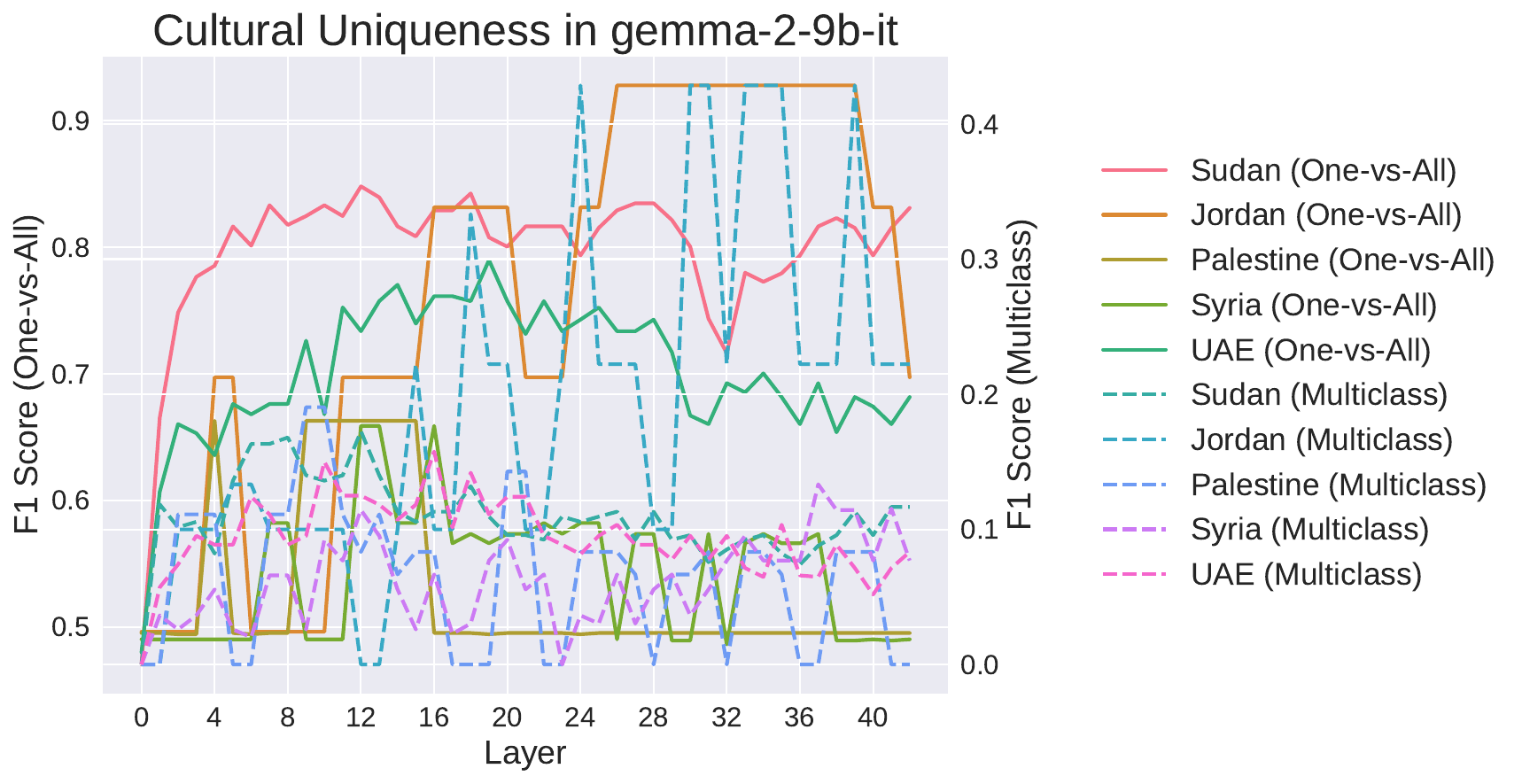}
\caption{F1 scores across model layers for different countries using one-vs-all and multiclass classifiers.}
\label{fig:probing_results3}
\end{figure}
\begin{figure}[H]
\centering
\includegraphics[width=\columnwidth]{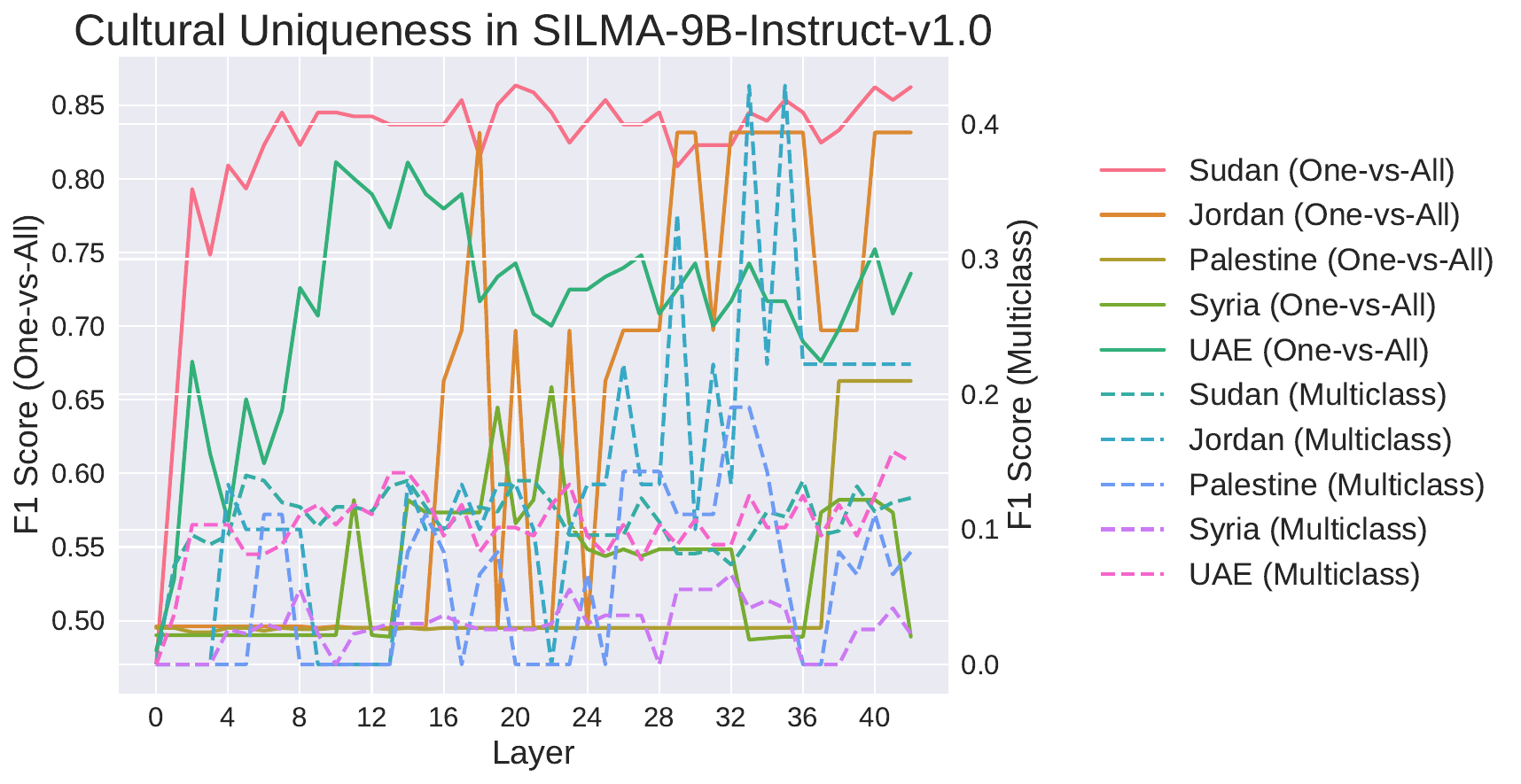}
\caption{F1 scores across model layers for different countries using one-vs-all and multiclass classifiers.}
\label{fig:probing_results4}
\end{figure}

\section{Probing Experiment Details}\label{probing_details}

In our probing experiments, the input to the model is a task-specific prompt constructed for each example, including the cultural statement and candidate responses. 

For each layer, we extract the hidden state of the final token in the prompt, which corresponds to the last token of the full input sequence. This token’s hidden state serves as the representation of the model’s internal encoding for that input. 
We will include this experimental detail in the camera-ready version of the paper.

To better understand this, we provide a example that illustrates the probing process step by step, from input text to cultural classification: 

This raw input is formatted into the following prompt (as used in our experiments): 

\begin{tcolorbox}[colback=gray!10, left=0mm, right=0mm, top=1mm, bottom=1mm] 
\small 

You are tasked with selecting the most culturally appropriate option based on the context provided below. 

Location: None   

Statement: When there is a success or a happy occasion, the mother is keen to cook a dessert as a good omen.

Consider the cultural nuances of the specified location and choose the most suitable response from the options provided. 

Options:   

A: The mother prepares date cookies.  

B: The mother prepares beverages.

C: The mother kneads khfaf.

\end{tcolorbox}

Typically, the pipeline is as follows:

1. LLM Encoding: We feed this prompt into a large language model, which produces high-dimensional hidden representations (tensors) for each token across all layers.

2. Representation Extraction: From each layer, we extract the hidden state of the final token in the prompt, resulting in a fixed-size vector (typically of shape d × 1).

3. Probing Classifier: This vector is passed into a logistic classifier trained to map the internal representation to a cultural class label. In this example, the correct label is Algeria.

\section{Cross-Cultural Transfer Beyond Arab cultures}
\label{sec:indo}
\paragraph{Data \& Training.} For Indonesian contexts, we selected Aceh \& Papua provinces based on geographical location/distance, since Aceh represents the westernmost province and Papua the westernmost. Data are curated from the Indoculture dataset \citep{koto2024} with 12 demonstrations selected representing one per topic. For US culture, we curated 12 human-written cultural samples (one per topic) to represent the US cultural context. These samples were used to fine-tune the model, which was then evaluated on Arab cultural tasks.
\paragraph{US Culture.} As represented in Section~\ref{beyond_arab_culture}, US contexts achieve the highest observed performance gains compared to other non-arab contexts. This shows that training with US cultural samples is transferred positively to Arab cultural tasks across both models. For Qwen-2.5-7B, the US achieved a MCQ accuracy of 68.71\% with ICL and 69.94\% with DiTTO, comparable to the average and lower bound of the Arab contexts. Similarly, for ALLaM-7B, the US reached 65.63\% (ICL) and 73.28\% (DiTTO), closely matching the Arab Upper Bound. The US context outperformed other non-Arab cultures, such as Indonesian contexts, indicating that cultural transfer can be observed in western contexts. For completion tasks, the improvements were more modest. Qwen-2.5-7B, the US achieved 35.03\% with ICL, similar to the Arab Lower Bound, while DiTTO (34.46\%) was able to match the Arab Upper Bound. In ALLaM-7B, the US achieved 38.71\% (ICL) and 38.91\% (DiTTO), which are slightly below the Arab Upper Bound. This suggests that while US cultural samples transfer well in MCQ tasks, their impact on completion tasks is more limited. 

To understand why US cultural contexts transfer effectively, we assessed the curated samples, where 7 of 12 samples are non-conflicting with Arab culture values or country-specific to the US, suggesting that compatibility and relevance of certain US cultural elements with Arab cultural values may contribute to the effective transfer across these contexts.

\begin{figure}[H]
  \centering
  \includegraphics[width=\columnwidth]{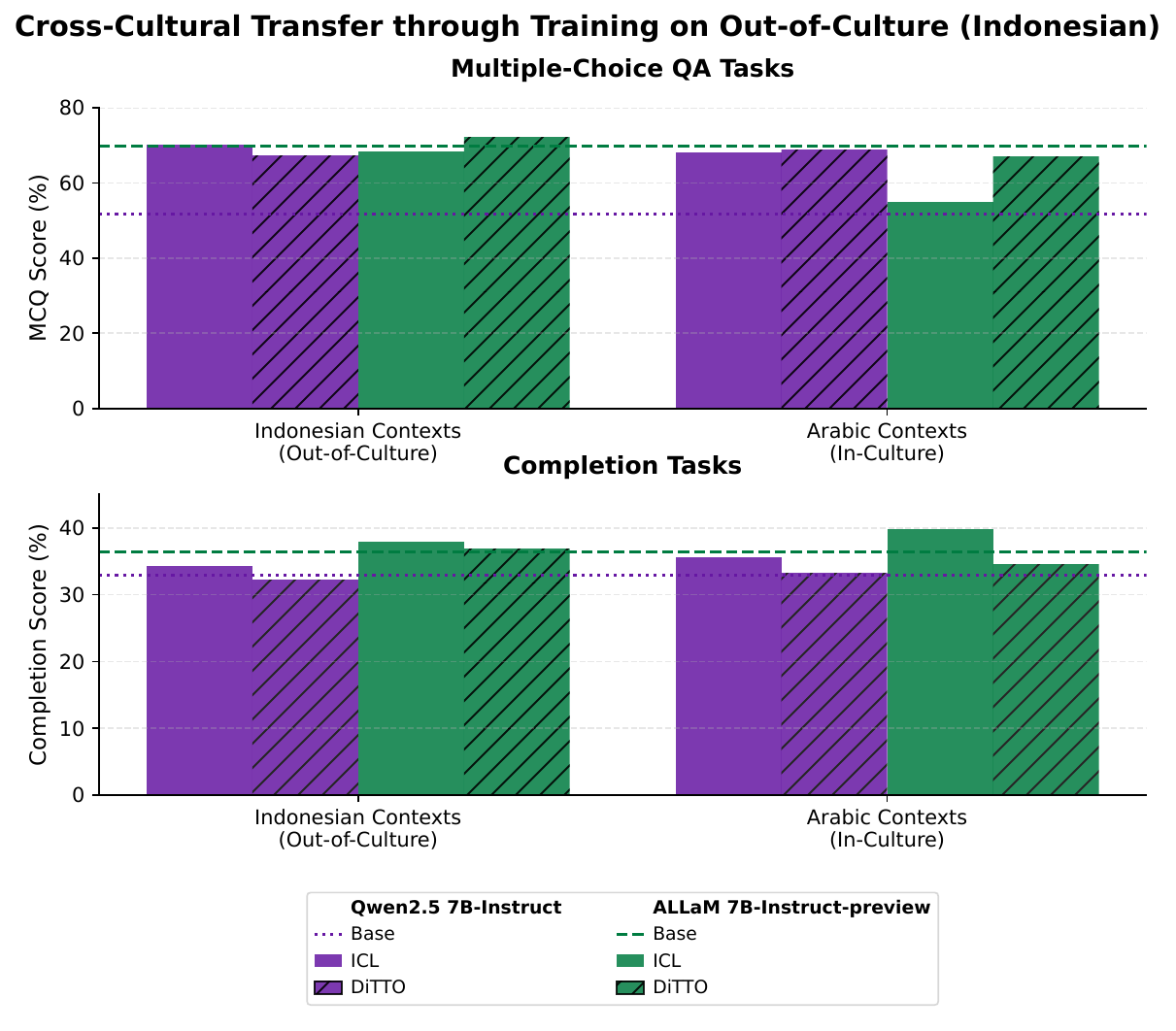}
  \caption{Performance comparison between Qwen-2.5-7B and ALLaM-7B Models trained on Indonesian contexts and evaluated on Arab culture.}
  \label{fig:indo_culture}
\end{figure}

\newpage
\section{Detailed Results}
\label{sec:results}

\begin{table}[h]
\centering
\scriptsize
\setlength{\tabcolsep}{4.5pt}
\begin{tabular}{lcccc}
\toprule
\textbf{Topic} & \textbf{ALLaM} & \textbf{Qwen} & \textbf{Gemma} & \textbf{SILMA} \\
\midrule
Agriculture & $+$4.2 & $+$3.4 & $+$12.9 & $+$1.8 \\
Art & $+$1.4 & $+$3.0 & $+$11.6 & $+$1.8 \\
Daily Activities & $+$0.5 & $+$10.7 & $+$10.9 & $+$0.2 \\
Death & $-$1.4 & $+$10.2 & $+$14.4 & $-$0.5 \\
Family Relations & $-$3.4 & $+$13.5 & $+$13.6 & $+$0.3 \\
Food & $-$2.2 & $+$7.7 & $+$10.5 & $+$0.8 \\
Habits & $-$1.6 & $+$10.0 & $+$13.7 & $+$1.1 \\
Holiday Activities & $-$1.8 & $+$9.6 & $+$13.7 & $+$1.4 \\
Idioms & $+$1.7 & $+$5.3 & $+$4.4 & $+$0.5 \\
Parenting & $+$4.5 & $+$9.4 & $+$9.7 & $+$0.1 \\
Traditional Games & $+$1.4 & $+$9.7 & $+$10.9 & $+$0.3 \\
Wedding & $-$1.9 & $+$10.1 & $+$11.4 & $+$1.1 \\
\midrule
\textbf{Average} & \textbf{$+$0.1} & \textbf{$+$9.4} & \textbf{$+$11.9} & \textbf{$+$0.7} \\
\bottomrule
\end{tabular}
\caption{DITTO performance improvements (\%).}
\label{tab:ditto-detailed}
\end{table}

\begin{table}[h]
\centering
\scriptsize
\setlength{\tabcolsep}{4.5pt}
\begin{tabular}{lcccc}
\toprule
\textbf{Topic} & \textbf{ALLaM} & \textbf{Qwen} & \textbf{Gemma} & \textbf{SILMA} \\
\midrule
Agriculture & $-$2.2 & $+$3.0 & $+$3.4 & $-$1.4 \\
Art & $-$3.8 & $+$4.3 & $+$1.6 & $-$1.2 \\
Daily Activities & $-$5.5 & $+$12.2 & $+$3.1 & $+$1.1 \\
Death & $-$6.1 & $+$12.2 & $+$4.5 & $+$0.8 \\
Family Relations & $-$5.9 & $+$15.5 & $+$3.8 & $-$1.1 \\
Food & $-$6.2 & $+$7.6 & $+$2.8 & $+$1.2 \\
Habits & $-$5.7 & $+$11.4 & $+$5.1 & $+$0.5 \\
Holiday Activities & $-$6.1 & $+$12.6 & $+$5.3 & $+$2.6 \\
Idioms & $+$0.4 & $+$4.8 & $-$0.5 & $+$1.7 \\
Parenting & $-$2.1 & $+$14.1 & $+$3.2 & $+$2.0 \\
Traditional Games & $-$6.3 & $+$10.6 & $+$2.7 & $-$0.7 \\
Wedding & $-$9.5 & $+$11.3 & $+$2.5 & $+$0.6 \\
\midrule
\textbf{Average} & \textbf{$-$4.9} & \textbf{$+$10.0} & \textbf{$+$3.2} & \textbf{$+$0.4} \\
\bottomrule
\end{tabular}
\caption{ICL performance improvements (\%).}
\label{tab:icl-detailed}
\end{table}

%\input{tables/main_results}

% Table for Qwen2.5 7B-Inst model
\begin{table*}[t]
\centering
\scriptsize  % Use smaller font size (smaller than \small)
\setlength{\tabcolsep}{2.2pt}
\renewcommand{\arraystretch}{1.8}  % increase vertical space

\label{tab:cross-test-qwen}
\resizebox{\textwidth}{!}{
\begin{tabular}{c|l|*{13}{c}|*{13}{c}}
\toprule
\multirow{2}{*}{\textbf{Method}} & \multirow{2}{*}{\begin{tabular}[c]{@{}l@{}}\textbf{Trained}\\\textbf{On}\end{tabular}} & \multicolumn{13}{c|}{\textbf{$\Delta$ MCQ vs. Base}} & \multicolumn{13}{c}{\textbf{$\Delta$ Completion vs. Base}} \\
\cmidrule(lr){3-15} \cmidrule(lr){16-28}
& & \rotatebox{90}{\tiny Alg} & \rotatebox{90}{\tiny Egy} & \rotatebox{90}{\tiny Jor} & \rotatebox{90}{\tiny KSA} & \rotatebox{90}{\tiny Leb} & \rotatebox{90}{\tiny Lib} & \rotatebox{90}{\tiny Mor} & \rotatebox{90}{\tiny Pal} & \rotatebox{90}{\tiny Sud} & \rotatebox{90}{\tiny Syr} & \rotatebox{90}{\tiny Tun} & \rotatebox{90}{\tiny UAE} & \rotatebox{90}{\tiny Yem} & \rotatebox{90}{\tiny Alg} & \rotatebox{90}{\tiny Egy} & \rotatebox{90}{\tiny Jor} & \rotatebox{90}{\tiny KSA} & \rotatebox{90}{\tiny Leb} & \rotatebox{90}{\tiny Lib} & \rotatebox{90}{\tiny Mor} & \rotatebox{90}{\tiny Pal} & \rotatebox{90}{\tiny Sud} & \rotatebox{90}{\tiny Syr} & \rotatebox{90}{\tiny Tun} & \rotatebox{90}{\tiny UAE} & \rotatebox{90}{\tiny Yem} \\
\midrule
\multirow{13}{*}{\begin{tabular}[c]{@{}c@{}}\rotatebox{90}{\small\includegraphics[height=1.5ex]{images/qwen-icon.png}} \\\rotatebox{90}{\textbf{ICL}}\end{tabular}} 
& Algeria & 18.1 & 20.7 & 15.7 & 19.3 & 17.2 & 28.2 & 21.7 & 24.8 & 15.9 & 16.4 & 13.9 & 25.0 & 4.4 & 0.4 & 6.2 & 7.9 & 2.1 & -0.4 & 1.9 & 3.2 & 3.2 & 8.6 & 0.8 & 1.3 & 1.9 & -0.8 \\
& Egypt & 22.2 & 22.3 & 16.5 & 22.3 & 15.1 & 29.6 & 24.5 & 27.2 & 15.0 & 17.2 & 13.4 & 25.4 & 5.6 & -0.4 & 8.7 & 6.4 & 2.5 & -1.7 & -1.4 & 0.8 & 4.0 & 7.3 & 2.4 & 1.3 & 0.8 & 0.0 \\
& Jordan & 15.3 & 18.2 & 17.2 & 15.1 & 13.4 & 23.6 & 17.8 & 22.8 & 15.0 & 16.8 & 13.0 & 24.6 & 9.2 & -2.8 & 3.7 & 13.9 & 0.0 & 0.0 & 1.9 & 3.5 & 0.4 & 12.9 & 3.5 & 0.4 & 0.8 & -2.8 \\
& KSA & 18.6 & 22.3 & 18.0 & 19.3 & 15.1 & 27.8 & 22.9 & 27.2 & 17.2 & 19.5 & 13.9 & 27.3 & 10.0 & -1.6 & 6.2 & 6.4 & 3.4 & 1.7 & 0.5 & 1.2 & 3.6 & 7.7 & 3.1 & 2.1 & 4.2 & 3.2 \\
& Lebanon & 21.4 & 18.6 & 16.1 & 19.8 & 15.5 & 29.2 & 21.0 & 23.2 & 12.9 & 18.4 & 12.2 & 26.2 & 8.8 & -1.2 & 7.0 & 12.4 & -0.8 & 1.7 & 1.9 & 4.7 & 1.6 & 13.7 & -0.8 & 0.8 & 2.7 & 1.6 \\
& Libya & 15.3 & 19.8 & 15.7 & 16.8 & 11.6 & 26.9 & 15.8 & 22.4 & 12.9 & 16.4 & 11.3 & 24.6 & 8.0 & -1.2 & 6.6 & 9.0 & 0.4 & -0.4 & 2.3 & 1.2 & 5.6 & 8.2 & 2.0 & 2.1 & 5.0 & 1.2 \\
& Morocco & 19.8 & 16.9 & 13.9 & 20.2 & 15.5 & 28.7 & 24.5 & 24.0 & 15.9 & 18.0 & 15.5 & 26.2 & 6.0 & -0.8 & 6.2 & 10.9 & 0.0 & 1.7 & 1.4 & 3.2 & 4.4 & 10.3 & 2.7 & 2.9 & -0.4 & 0.4 \\
& Palestine & 19.3 & 17.4 & 16.9 & 19.8 & 16.8 & 30.1 & 20.2 & 21.6 & 13.7 & 16.0 & 11.8 & 25.8 & 6.0 & -2.8 & 2.5 & 4.5 & 1.3 & -0.4 & 0.0 & 4.0 & 3.6 & 6.4 & -0.4 & 0.0 & 0.0 & 0.0 \\
& Sudan & 16.9 & 15.3 & 14.2 & 19.3 & 16.4 & 25.0 & 16.2 & 19.6 & 10.7 & 16.4 & 8.4 & 21.5 & 10.4 & 0.0 & 5.4 & 13.5 & 2.1 & 2.2 & 0.5 & 0.8 & 2.8 & 11.2 & 1.2 & 2.1 & 1.5 & -2.0 \\
& Syria & 18.9 & 20.2 & 16.5 & 16.0 & 17.2 & 32.9 & 22.1 & 22.8 & 16.3 & 18.8 & 11.8 & 27.7 & 7.2 & -1.6 & 6.2 & 9.7 & 4.2 & 1.3 & 1.9 & 3.2 & 3.2 & 10.3 & 1.2 & 2.9 & 2.3 & -2.0 \\
& Tunisia & 20.6 & 17.8 & 16.1 & 19.8 & 13.8 & 30.1 & 21.4 & 22.0 & 14.2 & 18.8 & 10.1 & 23.8 & 8.0 & 0.8 & 6.2 & 1.1 & 0.8 & 1.7 & 0.5 & 0.0 & 4.0 & 6.0 & 1.2 & 2.9 & 2.7 & -1.6 \\
& UAE & 13.7 & 14.0 & 15.7 & 19.8 & 13.4 & 25.5 & 16.6 & 20.4 & 13.7 & 16.8 & 13.4 & 24.6 & 11.2 & 0.8 & 7.9 & 7.9 & 4.2 & 1.3 & 1.4 & 2.8 & 3.6 & 11.6 & 2.7 & 0.4 & 6.2 & 0.8 \\
& Yemen & 6.4 & 12.4 & 12.7 & 16.8 & 8.6 & 18.0 & 9.5 & 16.4 & 15.4 & 11.7 & 8.4 & 17.3 & 2.8 & -1.2 & 3.7 & 5.6 & 0.4 & 1.7 & 2.8 & 1.2 & 1.2 & 4.7 & 1.2 & 0.4 & 1.9 & 4.0 \\
\midrule
\multirow{13}{*}{\begin{tabular}[c]{@{}c@{}}\rotatebox{90}{\small\includegraphics[height=1.5ex]{images/qwen-icon.png}} \\\rotatebox{90}{\textbf{Ditto}}\end{tabular}}
& Algeria & 18.6 & 16.5 & 14.6 & 14.7 & 16.4 & 21.3 & 21.7 & 22.8 & 15.0 & 13.3 & 9.7 & 23.5 & 9.6 & 0.8 & 2.9 & 3.0 & 0.0 & 1.3 & 1.9 & -2.8 & -0.4 & 2.6 & -0.8 & 1.3 & -5.0 & -1.6 \\
& Egypt & 23.0 & 18.6 & 16.9 & 16.0 & 15.1 & 24.5 & 22.9 & 21.6 & 15.4 & 15.6 & 7.6 & 23.1 & 8.0 & -1.2 & 1.2 & 1.5 & 0.0 & -0.4 & -0.9 & 1.2 & 1.6 & 2.6 & -5.5 & -0.4 & 1.2 & -2.0 \\
& Jordan & 19.4 & 20.7 & 15.4 & 18.9 & 19.0 & 25.5 & 24.5 & 24.0 & 17.2 & 19.5 & 8.8 & 23.5 & 10.0 & -1.6 & 6.2 & -0.7 & -0.4 & -2.6 & 1.4 & 2.4 & -1.6 & 2.2 & -3.5 & -1.3 & -2.3 & -1.6 \\
& KSA & 18.9 & 19.4 & 16.1 & 16.8 & 18.1 & 23.6 & 23.3 & 24.8 & 15.4 & 16.4 & 8.4 & 21.2 & 9.6 & 1.2 & 3.3 & 2.6 & 0.8 & -1.3 & 0.0 & 0.0 & -3.6 & 4.3 & -4.7 & -1.3 & -2.7 & 0.8 \\
& Lebanon & 21.0 & 17.4 & 16.5 & 16.4 & 18.5 & 23.2 & 22.5 & 24.4 & 16.3 & 16.4 & 12.6 & 24.2 & 9.6 & 1.6 & 3.7 & 3.0 & 2.1 & 0.9 & -0.5 & -3.6 & 2.0 & 2.6 & -0.8 & 1.7 & -0.8 & 0.0 \\
& Libya & 17.3 & 13.2 & 15.0 & 15.6 & 13.8 & 19.9 & 19.0 & 18.8 & 14.6 & 16.4 & 9.7 & 20.8 & 2.4 & -1.2 & 3.3 & 0.0 & 2.1 & -0.9 & 0.0 & -5.5 & 1.2 & 0.9 & -2.7 & 2.5 & -1.5 & 0.8 \\
& Morocco & 20.6 & 16.1 & 15.7 & 17.6 & 13.8 & 24.5 & 23.7 & 21.2 & 15.4 & 14.5 & 8.4 & 23.8 & 8.8 & 0.8 & 3.3 & 6.0 & 1.3 & 1.3 & -0.9 & 2.0 & -0.4 & 9.0 & -2.3 & 0.0 & -1.2 & 2.4 \\
& Palestine & 18.6 & 21.9 & 16.1 & 17.6 & 15.1 & 25.9 & 22.9 & 22.8 & 18.0 & 13.7 & 10.1 & 21.9 & 9.6 & -0.4 & 6.2 & 0.0 & 2.5 & -1.3 & 0.5 & -1.6 & -1.6 & 2.6 & -3.9 & 2.1 & -1.5 & -3.2 \\
& Sudan & 19.4 & 22.7 & 16.5 & 19.3 & 16.0 & 24.5 & 24.1 & 22.8 & 18.4 & 18.4 & 9.7 & 23.1 & 12.0 & 2.4 & 2.5 & 6.0 & 1.3 & -0.4 & 0.0 & 3.6 & -2.0 & 6.9 & -2.0 & 1.7 & -3.8 & -2.0 \\
& Syria & 16.9 & 17.8 & 15.4 & 16.4 & 16.4 & 22.7 & 23.3 & 21.2 & 16.3 & 14.1 & 8.8 & 22.3 & 9.6 & 2.8 & 4.1 & 3.8 & 0.0 & -1.3 & 1.9 & -1.2 & 1.2 & 5.2 & -3.1 & 2.1 & -1.2 & -1.2 \\
& Tunisia & 16.9 & 17.4 & 15.7 & 15.1 & 17.7 & 23.6 & 21.0 & 22.0 & 17.2 & 13.7 & 11.8 & 21.9 & 10.0 & 2.0 & 2.1 & -5.2 & 1.7 & -0.4 & 0.0 & -2.0 & -1.2 & 1.3 & -5.1 & -0.8 & -1.5 & -0.4 \\
& UAE & 18.1 & 15.7 & 16.9 & 17.6 & 16.0 & 24.5 & 22.1 & 20.8 & 18.4 & 12.1 & 8.8 & 21.2 & 7.2 & 1.6 & 4.1 & 3.4 & 0.8 & -0.4 & 0.5 & -1.2 & 2.0 & 6.9 & -1.2 & 1.3 & -3.5 & 0.0 \\
& Yemen & 18.6 & 19.8 & 15.7 & 17.6 & 16.8 & 23.6 & 23.3 & 24.8 & 19.3 & 16.0 & 11.8 & 23.5 & 10.8 & 0.8 & 2.9 & -3.0 & 0.4 & -1.7 & -2.3 & 0.4 & -0.4 & 1.7 & -4.7 & 0.0 & -4.6 & -0.8 \\
\bottomrule
\end{tabular}
}
\caption{Cross-country evaluation results for Qwen2.5 7B-Instruct. Models are evaluated on different countries (columns) after being trained on specific countries (rows). Values represent score difference from the base model.}
\end{table*}

\begin{table*}[t]
\centering
\Large
\scriptsize  % Use smaller font size (smaller than \small)
\setlength{\tabcolsep}{1.9pt}  
\renewcommand{\arraystretch}{1.8}  % increase vertical space
% Reduce column spacing even further
\label{tab:cross-test-gemma}
\resizebox{\textwidth}{!}{
\begin{tabular}{c|l|*{13}{c}|*{13}{c}}
\toprule
\multirow{2}{*}{\textbf{Method}} & \multirow{2}{*}{\begin{tabular}[c]{@{}l@{}}\textbf{Trained}\\\textbf{On}\end{tabular}} & \multicolumn{13}{c|}{\textbf{$\Delta$ MCQ vs. Base}} & \multicolumn{13}{c}{\textbf{$\Delta$ Completion vs. Base}} \\
\cmidrule(lr){3-15} \cmidrule(lr){16-28}
& & \rotatebox{90}{\tiny Alg} & \rotatebox{90}{\tiny Egy} & \rotatebox{90}{\tiny Jor} & \rotatebox{90}{\tiny KSA} & \rotatebox{90}{\tiny Leb} & \rotatebox{90}{\tiny Lib} & \rotatebox{90}{\tiny Mor} & \rotatebox{90}{\tiny Pal} & \rotatebox{90}{\tiny Sud} & \rotatebox{90}{\tiny Syr} & \rotatebox{90}{\tiny Tun} & \rotatebox{90}{\tiny UAE} & \rotatebox{90}{\tiny Yem} & \rotatebox{90}{\tiny Alg} & \rotatebox{90}{\tiny Egy} & \rotatebox{90}{\tiny Jor} & \rotatebox{90}{\tiny KSA} & \rotatebox{90}{\tiny Leb} & \rotatebox{90}{\tiny Lib} & \rotatebox{90}{\tiny Mor} & \rotatebox{90}{\tiny Pal} & \rotatebox{90}{\tiny Sud} & \rotatebox{90}{\tiny Syr} & \rotatebox{90}{\tiny Tun} & \rotatebox{90}{\tiny UAE} & \rotatebox{90}{\tiny Yem} \\
\midrule
\multirow{13}{*}{\begin{tabular}[c]{@{}c@{}}\rotatebox{90}{\small\includegraphics[height=1.5ex]{images/gemma-icon.png}} \\\rotatebox{90}{\textbf{ICL}}\end{tabular}} 
& Algeria & 6.5 & 4.5 & 0.7 & 3.8 & 3.5 & 2.3 & 5.9 & 8.4 & 1.7 & 10.6 & 2.1 & 5.8 & 0.4 & -0.4 & 2.9 & 9.4 & 8.0 & 3.5 & -0.5 & 2.8 & 5.2 & 8.6 & 2.0 & 2.9 & 3.1 & 2.4 \\
& Egypt & -0.4 & 0.8 & 0.7 & 1.3 & 0.0 & 0.0 & 0.4 & 1.2 & 0.9 & 1.6 & 0.8 & 0.8 & 0.0 & -2.8 & 3.7 & 3.8 & 3.8 & 1.7 & -1.4 & 2.4 & 3.6 & 5.6 & -1.6 & 3.4 & -0.4 & 0.8 \\
& Jordan & 14.5 & 6.6 & 1.1 & 12.6 & 6.9 & 13.0 & 12.6 & 16.4 & 4.7 & 20.7 & 10.9 & 15.0 & 0.4 & -2.4 & 3.3 & 17.6 & 6.3 & 3.0 & -0.9 & 4.7 & 2.8 & 10.7 & 2.7 & 3.4 & 3.9 & 6.0 \\
& KSA & 7.3 & 5.8 & 0.4 & 8.0 & 4.8 & 3.7 & 2.4 & 10.4 & 4.3 & 16.8 & 3.8 & 6.5 & 2.8 & -2.4 & 1.2 & 8.6 & 4.2 & 3.5 & 0.9 & 5.9 & 4.8 & 6.9 & 0.0 & 1.7 & 3.1 & 5.6 \\
& Lebanon & 0.0 & 0.0 & 0.0 & 0.0 & 0.0 & 0.0 & 0.0 & 0.0 & 0.0 & -0.4 & 0.4 & -0.4 & 2.0 & -3.6 & 4.5 & 8.2 & 9.7 & 4.7 & 0.0 & 4.0 & 0.8 & 6.4 & 2.7 & 2.5 & 3.1 & 2.4 \\
& Libya & 0.0 & 0.0 & 0.0 & 0.0 & 0.0 & 0.0 & 0.0 & 0.0 & 0.0 & -0.4 & 0.0 & -0.4 & 2.8 & -4.0 & 2.1 & 7.5 & 4.2 & 0.0 & -0.9 & 3.6 & 5.2 & 3.9 & 0.4 & 4.2 & 1.9 & 2.4 \\
& Morocco & 6.9 & 6.6 & 0.7 & 8.4 & 4.8 & 4.6 & 5.5 & 10.4 & 2.1 & 17.6 & 6.3 & 10.0 & 4.4 & -3.2 & 5.0 & 10.1 & 8.0 & 1.7 & 0.0 & 3.6 & 3.6 & 8.6 & 1.2 & 2.5 & 3.5 & 4.0 \\
& Palestine & 0.0 & 0.8 & 0.4 & 1.3 & 0.0 & 0.0 & 0.4 & 0.0 & 0.9 & 1.6 & 0.4 & 0.4 & -1.2 & -2.0 & 1.2 & 9.7 & 7.2 & -0.9 & 0.5 & 0.4 & 4.4 & 10.3 & 1.6 & 3.8 & 3.9 & 4.0 \\
& Sudan & 16.1 & 12.0 & 3.0 & 16.0 & 6.0 & 12.5 & 14.6 & 28.4 & 3.0 & 24.6 & 10.9 & 21.5 & 0.0 & -2.8 & 0.8 & 10.9 & 4.2 & 4.3 & -2.3 & 4.7 & 0.8 & 11.6 & 1.2 & 2.5 & 1.9 & 4.0 \\
& Syria & 0.0 & 0.8 & 0.0 & 0.4 & 0.9 & 0.0 & 0.4 & 0.0 & 0.9 & 1.6 & 0.0 & 0.4 & 0.8 & -6.5 & 2.1 & 6.0 & 4.6 & 3.9 & -1.9 & 4.4 & 5.6 & 7.3 & 3.9 & 0.0 & 1.2 & 2.8 \\
& Tunisia & 0.0 & 0.8 & 0.0 & 0.4 & 0.4 & 0.0 & 0.4 & 1.2 & 1.3 & 1.6 & 0.4 & 0.8 & 1.6 & -4.0 & 3.7 & 5.6 & 4.6 & 1.7 & -1.4 & 2.4 & 5.2 & 3.4 & 0.4 & 5.0 & 1.9 & 0.8 \\
& UAE & 0.8 & 0.4 & 0.4 & 2.5 & 3.0 & 0.5 & 0.8 & 2.8 & 0.9 & 4.7 & 1.3 & 2.3 & 2.0 & -1.2 & 4.5 & 5.6 & 5.0 & 3.0 & 0.5 & 1.2 & 4.8 & 7.3 & 1.6 & 0.0 & 6.9 & 6.4 \\
& Yemen & -0.4 & 0.8 & 0.0 & 0.8 & 0.0 & 0.0 & 0.4 & 0.0 & 0.4 & 0.8 & 0.0 & -0.4 & 1.2 & -0.8 & 2.9 & 4.5 & 5.5 & 0.4 & 0.9 & 1.6 & 6.4 & 3.9 & -1.2 & 3.8 & 1.5 & 5.2 \\
\midrule
\multirow{13}{*}{\begin{tabular}[c]{@{}c@{}}\rotatebox{90}{\small\includegraphics[height=1.5ex]{images/gemma-icon.png}} \\\rotatebox{90}{\textbf{Ditto}}\end{tabular}} 
& Algeria & 25.4 & 16.1 & 33.0 & 26.9 & 10.4 & 32.4 & 32.8 & 28.8 & 25.8 & 29.3 & 18.5 & 26.2 & 21.6 & -4.0 & -1.2 & -1.9 & 1.7 & 2.2 & -1.9 & -4.4 & 1.6 & 4.7 & -0.8 & -0.4 & -1.2 & 3.2 \\
& Egypt & 24.6 & 18.6 & 43.8 & 26.5 & 15.5 & 35.2 & 41.5 & 31.2 & 39.5 & 30.1 & 15.1 & 30.0 & 15.2 & -4.8 & 1.2 & -6.0 & 0.0 & 3.5 & -3.7 & -3.2 & -5.2 & 1.7 & -4.7 & 1.7 & -1.5 & 1.2 \\
& Jordan & 31.9 & 23.1 & 49.4 & 32.8 & 20.3 & 40.7 & 44.7 & 38.4 & 41.2 & 35.2 & 21.8 & 33.5 & 26.4 & -5.2 & 2.1 & 5.6 & 2.1 & 1.7 & -0.9 & 0.0 & 0.0 & 7.7 & 0.0 & 2.9 & 1.5 & 1.6 \\
& KSA & 18.2 & 20.2 & 44.6 & 23.9 & 13.4 & 32.4 & 37.6 & 25.6 & 40.8 & 30.9 & 18.9 & 31.9 & 16.8 & -1.6 & 7.4 & 6.4 & 2.1 & 6.5 & 1.9 & 0.8 & 0.0 & 9.5 & 1.6 & 3.8 & 0.4 & 4.8 \\
& Lebanon & 7.3 & 1.2 & 2.2 & 3.8 & 1.7 & 15.3 & 15.0 & 11.2 & 9.0 & 15.2 & 0.0 & 6.9 & 4.8 & -5.6 & 3.3 & 4.5 & 3.0 & 4.3 & 0.9 & -1.2 & -4.4 & 7.7 & 0.8 & 2.1 & 0.4 & 1.6 \\
& Libya & 32.3 & 21.1 & 49.8 & 32.4 & 16.0 & 40.7 & 41.1 & 40.0 & 43.3 & 36.3 & 20.2 & 30.4 & 30.8 & -2.4 & 4.1 & 0.8 & 5.5 & 3.0 & -1.9 & -4.7 & -0.8 & 6.9 & -3.5 & 0.8 & -1.9 & 0.0 \\
& Morocco & 8.5 & 9.9 & 30.3 & 8.4 & 4.3 & 14.8 & 23.7 & 14.4 & 26.2 & 12.1 & 6.3 & 11.5 & 6.0 & -6.5 & -2.1 & 4.1 & 3.0 & 1.3 & -1.9 & -2.0 & -5.2 & 4.3 & -3.1 & 0.0 & 0.4 & 2.4 \\
& Palestine & 22.6 & 18.6 & 30.3 & 23.5 & 13.4 & 34.7 & 33.2 & 26.8 & 29.2 & 30.1 & 17.2 & 23.5 & 20.8 & -2.8 & 2.9 & 0.8 & 0.0 & 0.4 & 2.3 & -4.0 & -2.4 & 1.7 & -0.8 & 3.4 & -0.8 & 4.0 \\
& Sudan & 14.1 & 10.7 & 21.3 & 13.0 & 5.2 & 17.1 & 23.3 & 18.0 & 20.6 & 17.6 & 10.1 & 14.2 & 10.0 & -4.4 & 0.4 & 3.0 & 2.5 & 2.2 & -0.9 & 0.4 & 3.2 & 3.0 & -1.2 & 2.9 & 2.7 & 4.8 \\
& Syria & 29.0 & 24.8 & 42.7 & 32.4 & 18.1 & 40.3 & 40.7 & 37.2 & 36.9 & 35.9 & 19.3 & 32.7 & 22.4 & -5.6 & 3.7 & 4.1 & 0.9 & 3.9 & 0.9 & -2.0 & -4.8 & 6.9 & -2.3 & -0.4 & -1.9 & -0.4 \\
& Tunisia & 21.0 & 14.9 & 23.6 & 18.9 & 12.1 & 28.7 & 27.7 & 22.4 & 21.9 & 25.8 & 14.7 & 20.0 & 15.6 & -1.2 & 2.5 & 5.6 & 3.8 & 2.6 & -1.4 & -2.0 & -1.2 & 7.3 & -3.1 & 0.4 & -1.5 & 6.0 \\
& UAE & 24.2 & 17.8 & 43.8 & 24.8 & 17.2 & 37.0 & 37.9 & 30.4 & 34.3 & 28.9 & 17.2 & 30.8 & 20.0 & -2.4 & 5.0 & 6.4 & 3.8 & 2.6 & 1.9 & 0.0 & 1.6 & 4.3 & 0.8 & 1.3 & 0.4 & 5.6 \\
& Yemen & 2.8 & 2.9 & 12.7 & 2.5 & 0.4 & 6.9 & 12.2 & 4.0 & 18.0 & 4.3 & 2.5 & 3.5 & 1.2 & 1.2 & 2.5 & 3.4 & 0.0 & 2.2 & -2.3 & -2.4 & -5.2 & 3.9 & -7.4 & 0.0 & -1.9 & 2.0 \\
\bottomrule
\end{tabular}
}
\caption{Cross-country evaluation results for Gemma-2 9B-it. Models are evaluated on different countries (columns) after being trained on specific countries (rows). Values represent score difference from the base model.}
\end{table*}

% Table for ALLaM 7b-Instruct-preview model
\begin{table*}[t]
\centering
\scriptsize  % Use smaller font size (smaller than \small)
\setlength{\tabcolsep}{1.8pt}  % Reduce column spacing even further
\renewcommand{\arraystretch}{1.8}  % increase vertical space

\label{tab:cross-test-allam}
\resizebox{\textwidth}{!}{
\begin{tabular}{c|l|*{13}{c}|*{13}{c}}
\toprule
\multirow{2}{*}{\textbf{Method}} & \multirow{2}{*}{\begin{tabular}[c]{@{}l@{}}\textbf{Trained}\\\textbf{On}\end{tabular}} & \multicolumn{13}{c|}{\textbf{$\Delta$ MCQ vs. Base}} & \multicolumn{13}{c}{\textbf{$\Delta$ Completion vs. Base}} \\
\cmidrule(lr){3-15} \cmidrule(lr){16-28}
& & \rotatebox{90}{\tiny Alg} & \rotatebox{90}{\tiny Egy} & \rotatebox{90}{\tiny Jor} & \rotatebox{90}{\tiny KSA} & \rotatebox{90}{\tiny Leb} & \rotatebox{90}{\tiny Lib} & \rotatebox{90}{\tiny Mor} & \rotatebox{90}{\tiny Pal} & \rotatebox{90}{\tiny Sud} & \rotatebox{90}{\tiny Syr} & \rotatebox{90}{\tiny Tun} & \rotatebox{90}{\tiny UAE} & \rotatebox{90}{\tiny Yem} & \rotatebox{90}{\tiny Alg} & \rotatebox{90}{\tiny Egy} & \rotatebox{90}{\tiny Jor} & \rotatebox{90}{\tiny KSA} & \rotatebox{90}{\tiny Leb} & \rotatebox{90}{\tiny Lib} & \rotatebox{90}{\tiny Mor} & \rotatebox{90}{\tiny Pal} & \rotatebox{90}{\tiny Sud} & \rotatebox{90}{\tiny Syr} & \rotatebox{90}{\tiny Tun} & \rotatebox{90}{\tiny UAE} & \rotatebox{90}{\tiny Yem} \\
\midrule
\multirow{13}{*}{\begin{tabular}[c]{@{}c@{}}\rotatebox{90}{\small\includegraphics[height=1.5ex]{images/allam-icon.jpg}} \\\rotatebox{90}{\textbf{ICL}}\end{tabular}} 
& Algeria & -9.7 & -15.7 & 0.0 & -11.8 & -11.2 & -14.4 & -11.9 & -14.4 & -9.0 & -10.2 & -9.2 & -9.6 & -10.8 & -0.4 & 2.1 & 11.2 & 4.2 & 3.0 & 3.7 & 5.1 & 3.6 & 3.4 & 4.7 & 8.8 & 0.8 & 1.2 \\
& Egypt & -25.4 & -26.0 & -17.2 & -26.9 & -20.7 & -23.6 & -28.9 & -34.8 & -22.3 & -31.3 & -17.6 & -29.2 & -24.4 & -1.6 & 0.4 & 10.5 & 5.0 & 1.3 & 2.3 & 3.6 & 4.0 & 4.7 & 2.0 & 7.6 & 2.7 & 4.0 \\
& Jordan & -3.2 & -8.3 & -0.4 & -5.5 & -5.2 & -6.9 & -7.9 & -4.0 & -3.9 & -5.9 & -6.7 & -2.7 & -3.2 & 0.0 & 0.8 & 13.5 & 3.4 & 2.6 & 1.4 & 5.1 & 3.2 & 9.0 & 2.3 & 5.0 & 1.2 & 2.0 \\
& KSA & -16.9 & -19.4 & -6.7 & -18.1 & -16.4 & -18.1 & -19.4 & -23.2 & -16.7 & -17.6 & -13.4 & -13.5 & -13.2 & -4.0 & 2.1 & 9.4 & 5.5 & 2.2 & 1.4 & 2.4 & 3.6 & 4.7 & 5.1 & 3.8 & 1.5 & 2.8 \\
& Lebanon & -14.5 & -17.4 & -2.3 & -16.4 & -15.5 & -14.8 & -15.8 & -21.6 & -12.0 & -17.2 & -15.5 & -16.5 & -15.2 & -0.8 & 1.2 & 10.9 & 5.0 & 1.3 & 5.6 & 2.8 & 4.0 & 4.7 & 4.3 & 5.9 & 3.5 & 3.2 \\
& Libya & -14.5 & -16.5 & -6.0 & -13.5 & -13.4 & -13.4 & -16.6 & -15.6 & -13.3 & -13.7 & -13.0 & -11.5 & -11.6 & -2.8 & 0.8 & 9.0 & 3.8 & 2.6 & 4.2 & 2.0 & 2.8 & 3.4 & 3.9 & 3.8 & -0.8 & 1.6 \\
& Morocco & -9.3 & -13.6 & -6.4 & -13.9 & -13.4 & -13.0 & -15.0 & -14.0 & -13.3 & -12.5 & -8.0 & -8.9 & -12.0 & 0.0 & 1.7 & 11.2 & 3.4 & 2.2 & 3.7 & 8.7 & 3.2 & 5.2 & 2.3 & 4.6 & 2.3 & 2.0 \\
& Palestine & -19.8 & -21.9 & -12.7 & -18.1 & -19.0 & -22.7 & -20.6 & -24.8 & -22.3 & -21.5 & -14.3 & -21.2 & -18.8 & 0.0 & 0.8 & 9.7 & 4.2 & 0.9 & 4.2 & 1.6 & 4.0 & 0.4 & 2.3 & 5.0 & -0.4 & 3.2 \\
& Sudan & -21.8 & -25.6 & -21.7 & -23.5 & -18.5 & -21.3 & -27.7 & -27.2 & -26.6 & -23.4 & -17.2 & -21.5 & -20.8 & -2.4 & 0.4 & 9.4 & 2.1 & 1.3 & 1.9 & 4.8 & 3.6 & 8.6 & 2.0 & 5.9 & 1.9 & 2.0 \\
& Syria & -2.4 & -13.2 & -1.1 & -8.4 & -6.9 & -5.1 & -5.5 & -8.8 & -6.4 & -5.1 & -5.5 & -4.2 & -4.8 & -1.2 & 0.8 & 10.1 & 5.0 & 0.9 & 4.2 & 5.1 & 6.0 & 5.6 & 6.6 & 7.1 & 1.9 & 2.0 \\
& Tunisia & -19.4 & -22.7 & -7.1 & -23.1 & -15.1 & -17.1 & -24.5 & -26.4 & -15.9 & -22.7 & -11.3 & -20.0 & -16.4 & -0.4 & 0.0 & 8.6 & 2.9 & 3.0 & 5.6 & 5.9 & 5.2 & 4.7 & 6.3 & 6.7 & 3.5 & 4.0 \\
& UAE & -12.1 & -16.5 & -6.0 & -11.8 & -10.8 & -11.6 & -14.2 & -18.0 & -10.7 & -11.7 & -10.9 & -9.6 & -10.4 & -2.0 & 0.0 & 8.2 & 3.8 & 1.3 & 4.2 & -0.8 & 5.2 & 6.9 & 5.5 & 5.0 & 3.5 & 2.8 \\
& Yemen & -17.7 & -24.0 & -7.1 & -17.7 & -13.8 & -16.7 & -19.8 & -22.8 & -21.9 & -23.0 & -13.9 & -16.2 & -15.6 & 0.0 & -0.8 & 6.0 & 2.5 & 0.9 & 5.6 & 0.4 & 2.8 & 3.9 & 4.3 & 4.2 & 1.5 & 6.0 \\
\midrule
\multirow{13}{*}{\begin{tabular}[c]{@{}c@{}}\rotatebox{90}{\small\includegraphics[height=1.5ex]{images/allam-icon.jpg}} \\\rotatebox{90}{\textbf{Ditto}}\end{tabular}}
& Algeria & -2.8 & -5.0 & 3.0 & 4.6 & -1.7 & 9.3 & 3.2 & 0.0 & 0.4 & 4.3 & 3.8 & 3.5 & -1.6 & -5.2 & 0.8 & -5.2 & -2.5 & -0.9 & 0.0 & 0.4 & 1.6 & 2.2 & 3.1 & 3.8 & -2.7 & 1.2 \\
& Egypt & -4.8 & -6.2 & 0.0 & 0.8 & -2.2 & -0.9 & -1.6 & 2.0 & 0.9 & -4.3 & 4.2 & -0.4 & -1.6 & -6.1 & 1.7 & -9.0 & -6.3 & 0.4 & 0.0 & -4.4 & -2.4 & -4.7 & 0.4 & 2.1 & -5.4 & -1.2 \\
& Jordan & -10.5 & -13.2 & -20.6 & -13.9 & -9.9 & -8.3 & -13.8 & -18.0 & -16.3 & -10.2 & -7.1 & -13.9 & -9.6 & -6.5 & -2.1 & -12.4 & -7.6 & -3.0 & -0.9 & -4.4 & -6.8 & -6.4 & 1.2 & 1.7 & -6.9 & -7.6 \\
& KSA & -1.2 & -4.5 & -0.8 & -0.4 & -7.3 & 0.0 & -0.4 & -4.0 & -3.0 & -0.8 & 2.1 & -1.9 & -2.8 & -4.0 & 4.1 & -0.4 & 1.3 & -0.4 & 3.2 & 0.4 & 1.6 & 0.0 & 1.6 & 3.4 & -1.9 & 2.0 \\
& Lebanon & 2.8 & 0.0 & 3.0 & 6.7 & 1.7 & 8.8 & 5.1 & 3.6 & 1.3 & 3.5 & 5.1 & 3.8 & 3.2 & -0.4 & 2.9 & 0.0 & -8.0 & 0.4 & 2.8 & -1.2 & 2.8 & 0.9 & 0.4 & 5.5 & -3.5 & 2.4 \\
& Libya & -0.4 & -6.6 & 2.2 & -0.4 & -0.4 & -0.5 & -1.2 & -0.4 & 0.0 & 0.8 & 0.8 & 0.0 & 2.0 & -4.4 & 1.2 & -7.5 & -6.7 & -3.0 & 0.9 & -1.2 & -0.8 & -4.7 & 1.6 & 2.9 & -3.5 & -1.2 \\
& Morocco & 2.0 & -2.5 & 5.6 & 2.9 & -0.9 & 7.0 & 2.4 & 2.8 & -0.4 & 6.6 & 5.5 & 4.2 & 5.2 & -3.6 & 3.3 & 4.9 & -2.5 & 2.2 & -0.9 & 0.0 & -0.8 & -2.1 & 1.6 & 2.9 & -5.4 & -0.8 \\
& Palestine & 2.8 & -5.0 & 1.9 & 0.4 & 0.9 & 5.1 & -2.4 & 0.8 & -2.6 & -0.4 & 4.2 & 1.5 & 3.6 & -6.5 & -0.8 & -10.5 & -7.6 & -0.4 & 0.0 & -5.9 & -4.8 & -6.0 & 2.0 & 2.9 & -6.9 & 2.4 \\
& Sudan & -2.4 & -2.9 & 2.6 & 2.5 & -2.2 & 5.1 & 3.2 & 0.0 & 2.1 & 2.7 & 6.3 & 1.9 & 3.6 & -3.2 & 2.1 & -4.9 & -3.4 & -0.9 & -2.8 & -2.0 & 1.2 & -2.6 & 0.8 & 2.5 & -5.0 & -3.2 \\
& Syria & 2.0 & -2.5 & 0.8 & 2.1 & 0.9 & 7.9 & -0.4 & 4.8 & 0.4 & 3.9 & 8.0 & 1.5 & 4.4 & -4.0 & 2.1 & -5.6 & -5.0 & -0.4 & 0.0 & -2.0 & -0.4 & -3.0 & 4.3 & 2.9 & -6.5 & -0.4 \\
& Tunisia & -0.4 & -0.8 & -0.8 & 1.7 & 1.3 & 7.4 & 3.2 & 0.0 & -1.3 & 0.4 & 2.1 & 1.5 & 4.0 & -2.4 & 3.3 & 2.6 & -0.8 & 2.2 & 0.0 & 0.8 & 0.0 & -3.0 & 0.0 & 2.9 & -3.1 & 0.4 \\
& UAE & 0.0 & -2.9 & 5.6 & 5.9 & -1.7 & 8.3 & 4.0 & 4.8 & 1.7 & 5.9 & 5.5 & 2.7 & 6.8 & -5.7 & -0.4 & -8.6 & -3.4 & -1.7 & -0.9 & -1.2 & -1.6 & -3.0 & 0.0 & 5.5 & -3.5 & -1.6 \\
& Yemen & -2.4 & -3.3 & 0.4 & -1.7 & -5.6 & 6.0 & -1.2 & 0.0 & -2.2 & -0.4 & 3.4 & -0.8 & 1.6 & -3.2 & 4.6 & 0.0 & -4.2 & 1.7 & 1.4 & 0.8 & 2.4 & -3.9 & 2.7 & 5.9 & -2.3 & 1.2 \\
\bottomrule
\end{tabular}
}
\caption{Cross-country evaluation results for ALLaM 7B-Instruct-preview. Models are evaluated on different countries (columns) after being trained on specific countries (rows). Values represent score difference from the base model.}
\end{table*}

% Table for Silma 9b-Instruct model
\begin{table*}[t]
\centering
\scriptsize  % Use smaller font size (smaller than \small)
\setlength{\tabcolsep}{1.9pt}  % Reduce column spacing even further
\renewcommand{\arraystretch}{1.8}  % increase vertical space

\label{tab:cross-test-silma}
\resizebox{\textwidth}{!}{
\begin{tabular}{c|l|*{13}{c}|*{13}{c}}
\toprule
\multirow{2}{*}{\textbf{Method}} & \multirow{2}{*}{\begin{tabular}[c]{@{}l@{}}\textbf{Trained}\\\textbf{On}\end{tabular}} & \multicolumn{13}{c|}{\textbf{$\Delta$ MCQ vs. Base}} & \multicolumn{13}{c}{\textbf{$\Delta$ Completion vs. Base}} \\
\cmidrule(lr){3-15} \cmidrule(lr){16-28}
& & \rotatebox{90}{\tiny Alg} & \rotatebox{90}{\tiny Egy} & \rotatebox{90}{\tiny Jor} & \rotatebox{90}{\tiny KSA} & \rotatebox{90}{\tiny Leb} & \rotatebox{90}{\tiny Lib} & \rotatebox{90}{\tiny Mor} & \rotatebox{90}{\tiny Pal} & \rotatebox{90}{\tiny Sud} & \rotatebox{90}{\tiny Syr} & \rotatebox{90}{\tiny Tun} & \rotatebox{90}{\tiny UAE} & \rotatebox{90}{\tiny Yem} & \rotatebox{90}{\tiny Alg} & \rotatebox{90}{\tiny Egy} & \rotatebox{90}{\tiny Jor} & \rotatebox{90}{\tiny KSA} & \rotatebox{90}{\tiny Leb} & \rotatebox{90}{\tiny Lib} & \rotatebox{90}{\tiny Mor} & \rotatebox{90}{\tiny Pal} & \rotatebox{90}{\tiny Sud} & \rotatebox{90}{\tiny Syr} & \rotatebox{90}{\tiny Tun} & \rotatebox{90}{\tiny UAE} & \rotatebox{90}{\tiny Yem} \\
\midrule
\multirow{13}{*}{\begin{tabular}[c]{@{}c@{}}\rotatebox{90}{\small\includegraphics[height=1.5ex]{images/silma-icon.png}} \\\rotatebox{90}{\textbf{ICL}}\end{tabular}} 
& Algeria & -1.2 & -2.5 & -1.5 & 1.7 & 3.5 & -1.9 & 3.6 & 2.8 & -0.9 & 2.0 & -0.4 & 7.3 & -2.8 & 0.4 & 0.0 & 6.0 & 1.7 & -1.7 & 1.9 & 2.4 & 1.2 & 7.3 & 2.0 & 0.8 & -0.4 & 2.0 \\
& Egypt & -6.5 & -4.1 & -3.4 & 0.0 & 6.0 & -3.7 & 2.4 & 2.4 & 0.4 & -2.0 & 0.4 & 2.7 & -3.2 & -2.4 & -1.2 & 1.9 & 1.7 & -2.6 & 1.9 & 0.8 & 0.0 & 1.7 & -0.4 & 0.4 & -3.8 & 2.8 \\
& Jordan & 2.0 & 1.7 & 1.1 & 4.6 & 3.0 & 1.4 & 5.9 & 6.4 & 3.0 & 4.7 & 0.0 & 5.0 & 1.2 & 0.4 & 1.2 & 8.2 & 5.0 & -0.4 & 2.8 & 2.4 & 3.2 & 3.9 & 1.2 & -1.3 & -3.1 & 3.2 \\
& KSA & 1.2 & 2.1 & -2.6 & 4.6 & 0.9 & -0.5 & 2.0 & 4.0 & -2.1 & 0.8 & 0.8 & 4.6 & 2.0 & 0.8 & 0.0 & 5.6 & 4.6 & -1.7 & 2.8 & 2.0 & 0.8 & 2.1 & -1.6 & -0.8 & -2.3 & 4.0 \\
& Lebanon & -6.1 & -4.1 & -2.6 & -1.3 & 3.0 & -8.3 & 0.4 & -0.8 & 0.9 & 0.8 & 0.4 & 2.7 & -5.6 & -3.2 & -1.2 & 1.9 & 3.8 & -2.6 & 1.4 & 2.0 & 0.8 & 2.1 & 2.0 & -0.4 & -0.8 & 2.4 \\
& Libya & -13.7 & -12.0 & -3.7 & -5.5 & -1.3 & -14.8 & -7.9 & -4.4 & -2.1 & -9.8 & -3.0 & -0.8 & -12.4 & 0.0 & 1.2 & 2.6 & 3.4 & -0.9 & 1.9 & 3.2 & 2.8 & 2.1 & -0.8 & -0.4 & -0.8 & 3.2 \\
& Morocco & 4.0 & 2.5 & 0.0 & 2.9 & 5.2 & 0.5 & 6.7 & 6.0 & 1.3 & 7.0 & 4.2 & 4.2 & 0.4 & -1.6 & 2.9 & 6.0 & 4.6 & 0.0 & 2.8 & 2.8 & 1.6 & 3.9 & 0.8 & -0.8 & -0.4 & 2.0 \\
& Palestine & -3.6 & -3.7 & -1.1 & 2.5 & 2.6 & -4.2 & 1.2 & 2.4 & 1.3 & 2.0 & -0.4 & 4.2 & -3.6 & -2.4 & 0.8 & 4.9 & 5.0 & -0.9 & 2.3 & 2.0 & 1.6 & 3.9 & 1.2 & 0.8 & -1.2 & 2.0 \\
& Sudan & -1.6 & 0.0 & -1.1 & 4.6 & 2.2 & -0.9 & 3.2 & 1.6 & 1.7 & 0.0 & 2.1 & 3.8 & -1.2 & -1.6 & 1.2 & 6.4 & 4.6 & 0.4 & 2.3 & 2.0 & 0.8 & 9.0 & 0.4 & -1.3 & -2.3 & 2.4 \\
& Syria & 2.8 & 1.7 & 0.8 & 4.2 & 6.5 & 2.3 & 3.6 & 6.4 & 3.0 & 6.6 & 0.4 & 5.0 & 0.8 & -1.6 & -1.2 & 3.0 & 1.7 & -1.7 & 1.9 & 0.8 & 0.8 & 4.3 & 1.2 & -1.7 & -3.1 & 3.2 \\
& Tunisia & -0.8 & 1.2 & -3.4 & 1.7 & 5.2 & -5.1 & 1.6 & 2.8 & -1.3 & 2.4 & -1.3 & 2.7 & -1.6 & -0.8 & 2.1 & 0.8 & 1.3 & -1.7 & 1.4 & 0.0 & 0.4 & 0.9 & -0.4 & 1.3 & -2.3 & 2.4 \\
& UAE & -1.2 & 1.2 & -1.5 & 5.9 & 3.5 & -1.4 & 4.0 & 4.4 & 1.3 & 4.3 & 2.1 & 5.4 & 1.2 & -0.4 & 2.1 & 3.4 & 2.9 & 0.0 & 2.8 & 2.4 & 2.8 & 5.2 & 2.3 & -0.4 & -0.4 & 4.0 \\
& Yemen & -4.8 & -5.4 & -2.6 & 0.4 & 3.9 & -5.1 & 0.8 & 0.0 & -0.4 & -0.4 & 2.1 & 1.9 & -0.4 & -0.4 & 2.1 & 2.6 & 2.1 & 0.9 & 1.4 & 2.0 & -2.4 & 0.9 & -0.8 & 1.7 & -4.6 & 3.2 \\
\midrule
\multirow{13}{*}{\begin{tabular}[c]{@{}c@{}}\rotatebox{90}{\small\includegraphics[height=1.5ex]{images/silma-icon.png}} \\\rotatebox{90}{\textbf{Ditto}}\end{tabular}}
& Algeria & 0.0 & -1.7 & 0.0 & 4.2 & 5.6 & -1.9 & 1.2 & 2.0 & 1.7 & 5.5 & 0.8 & 6.5 & 3.2 & -2.8 & -5.4 & -3.0 & -2.5 & 3.0 & 0.0 & -0.8 & -2.0 & 1.7 & 3.5 & -3.8 & -2.7 & 3.6 \\
& Egypt & -3.2 & -4.1 & -3.0 & 0.4 & 1.3 & -0.5 & -2.4 & -0.4 & 0.0 & -2.7 & 0.8 & 0.8 & -2.4 & -2.4 & 0.0 & -4.1 & -3.0 & 0.9 & -2.8 & -0.4 & -3.2 & -0.4 & 1.2 & 0.0 & -1.9 & 3.6 \\
& Jordan & 0.4 & -2.1 & -0.4 & 3.4 & 2.2 & 1.4 & 2.8 & 0.4 & 0.9 & 5.1 & 1.7 & 3.1 & 1.6 & -3.2 & -0.4 & 4.9 & 4.6 & 0.9 & 0.0 & 0.0 & 0.8 & 3.9 & 3.1 & 0.4 & -3.1 & 2.8 \\
& KSA & 0.0 & -1.2 & 0.4 & 0.8 & 2.6 & 2.3 & -0.8 & 0.8 & 0.9 & 2.4 & 0.4 & -0.4 & 0.8 & -2.0 & 0.4 & 1.1 & -1.7 & 1.3 & -0.5 & -0.8 & 0.0 & 1.7 & 1.6 & 0.4 & -1.2 & 2.4 \\
& Lebanon & -6.9 & -9.1 & -4.1 & -3.8 & 0.0 & -3.2 & -3.2 & -3.6 & -1.3 & 0.0 & -5.5 & 2.3 & -1.2 & -3.6 & -2.9 & 0.4 & 3.4 & 0.4 & 1.4 & 4.4 & 1.6 & 2.1 & 3.1 & -1.3 & -0.4 & 2.8 \\
& Libya & 0.0 & -6.2 & -0.8 & -0.8 & 3.5 & -3.2 & -1.2 & -0.8 & 1.3 & 4.3 & -2.1 & -0.8 & 2.0 & -2.0 & -2.9 & 2.2 & 2.5 & 2.2 & 1.9 & -2.8 & 1.6 & 4.3 & 4.3 & -2.9 & -1.5 & 4.0 \\
& Morocco & 1.2 & -2.1 & 0.8 & 3.8 & 6.5 & 0.0 & 3.2 & 3.2 & 0.4 & 5.9 & 2.5 & 5.0 & 4.4 & -0.4 & -1.2 & 3.4 & 2.5 & 0.0 & 0.9 & 4.0 & 0.0 & 2.6 & 2.7 & -0.8 & -3.5 & 4.4 \\
& Palestine & -1.2 & -2.5 & 0.4 & 5.0 & 6.5 & 0.5 & 1.6 & 2.4 & 0.0 & 7.0 & 1.7 & 1.2 & 4.0 & -2.0 & -4.6 & -2.6 & -1.7 & 0.9 & 0.9 & 0.4 & 1.6 & 2.6 & 2.7 & 2.5 & 0.8 & 3.6 \\
& Sudan & 2.4 & -1.7 & 1.1 & 5.5 & 6.0 & 1.9 & 0.4 & 4.8 & 0.9 & 6.3 & 1.7 & 2.3 & 4.0 & -0.4 & -3.7 & 6.0 & 3.8 & -1.3 & 1.4 & 0.0 & 4.0 & 4.7 & 2.0 & 2.9 & -0.8 & 3.2 \\
& Syria & -1.6 & -1.2 & -0.8 & 2.9 & 4.7 & 0.5 & 0.8 & 2.4 & 1.7 & 6.6 & -0.4 & 5.0 & 4.4 & -3.2 & -6.6 & -3.0 & -0.4 & 2.6 & 1.4 & 1.2 & -1.6 & 2.1 & 2.3 & -2.1 & -1.9 & 6.0 \\
& Tunisia & 2.8 & -2.5 & 0.8 & 2.1 & 2.6 & 2.3 & 0.0 & 3.6 & -0.9 & 7.0 & 1.7 & 3.5 & 6.8 & 2.0 & -1.7 & 1.9 & 0.8 & 0.9 & 0.9 & -0.8 & -0.8 & 1.3 & 0.8 & 0.0 & -4.2 & 2.8 \\
& UAE & 1.2 & -2.1 & 1.1 & 3.4 & 5.6 & 1.9 & 0.4 & 1.2 & 0.9 & 7.4 & 0.0 & 1.5 & 6.0 & -1.6 & -1.2 & 3.0 & 5.0 & 0.9 & 0.0 & 0.4 & 3.6 & 1.7 & 4.7 & -0.4 & -1.5 & 7.2 \\
& Yemen & -1.2 & -2.9 & -1.5 & -0.8 & 6.9 & 3.2 & -3.2 & -0.8 & -2.6 & 2.4 & 4.6 & 3.5 & -1.2 & -2.0 & -1.2 & -3.0 & 0.4 & 2.6 & -1.4 & 0.8 & -1.6 & 1.3 & 0.0 & 2.9 & -4.2 & 4.4 \\
\bottomrule
\end{tabular}
}
\caption{Cross-country evaluation results for SILMA 9B-Instruct. Models are evaluated on different countries (columns) after being trained on specific countries (rows). Values represent score difference from the base model.}
\end{table*}

\end{document}